\documentclass[letterpaper,times]{IONconf}

\usepackage{times}
\usepackage{epsfig}
\usepackage[T1]{fontenc}
\usepackage{graphicx}
\usepackage{booktabs}
\usepackage[misc]{ifsym}

\usepackage{paralist}
\usepackage{eurosym}
\usepackage{flushend}
\usepackage{amsmath}
\usepackage{amssymb}
\usepackage{url}
\usepackage[utf8]{inputenc}
\usepackage{amsfonts}
\usepackage{soul}
\usepackage[justification=centering]{caption}
\usepackage[list=true]{subcaption}
\usepackage{tabu}
\usepackage{wrapfig}
\usepackage{arydshln}
\usepackage[normalem]{ulem}
\usepackage{multirow}
\usepackage{siunitx}
\usepackage{natbib}
\usepackage{hyperref}
\usepackage{makecell}
\usepackage{adjustbox}
\usepackage{array}
\usepackage{diagbox}

\def\rot{\rotatebox}

\title{Evaluation of (Un-)Supervised Machine Learning Methods for GNSS Interference Classification with Real-World Data Discrepancies}

\author{Lucas Heublein\textsuperscript{1}, Nisha L. Raichur\textsuperscript{1}, Tobias Feigl\textsuperscript{1}, Tobias Brieger\textsuperscript{1}, Fin Heuer\textsuperscript{2}, Lennart Asbach\textsuperscript{2}, Alexander Rügamer\textsuperscript{1},\\ 
\vspace{0.1cm}
Felix Ott\textsuperscript{1}\\
\textsuperscript{1}\textit{Fraunhofer Institute for Integrated Circuits IIS, Nürnberg, Germany}\\
\vspace{0.1cm}
\textsuperscript{2}\textit{German Aerospace Center (DLR), Braunschweig, Germany}\\
{\tt\small \{lucas.heublein,nisha.lakshmana.raichur,tobias.feigl,tobias.brieger,alexander.ruegamer,}\\
\tt\small {felix.ott\}@iis.fraunhofer.de}, {\tt\small \{fin.heuer,lennart.asbach\}@dlr.de}
}

\begin{document}

\maketitle

\section*{biography}

\biography{Lucas Heublein}{earned his M.Sc.~in Integrated Life Science from Friedrich-Alexander University (FAU) Erlangen-Nürnberg and completed his Master of Science in Computer Science in 2024. He began his association with the Hybrid Positioning \& Information Fusion group at the Fraunhofer Institute for Integrated Circuits IIS in 2020 as a student assistant and currently serves as a research assistant in the Self-Learning Systems group.}

\biography{Nisha L. Raichur}{received her M.Sc.~degree in Cognitive Systems from Technical University Ulm, Germany in 2020.
Since October 2020 she works as a research assistant in the Hybrid Positioning \& Information Fusion group of the Fraunhofer Institute for Integrated Circuits IIS, Nuremberg, Germany. Her research focus is on multimodal learning to improve digital signal processing and information fusion.}

\biography{Tobias Feigl}{received his Ph.D.~degree in Computer Science from the Friedrich-Alexander University (FAU) Erlangen-Nürnberg in 2021 and his Masters degree from the University of Applied Sciences Erlangen-Nuremberg, Germany, in 2017. In 2017 he joined the Fraunhofer Institute for Integrated Circuits IIS Nuremberg, Germany, where he worked as a student since 2009, with a strong focus on AI and information fusion. In parallel, since 2017 he is a lecturer at the Computer Science department at FAU, where he gives courses on machine and deep learning and supervises related qualification work.}

\biography{Tobias Brieger}{received his M.Sc.~degree in Computer Science from Friedrich-Alexander University (FAU) Erlangen-Nürnberg, Germany in 2023. Since February 2023, he works as a research assistant at the Fraunhofer Institute for Integrated Circuits IIS Nürnberg, Germany, in the Specialized SatNav Receivers group. His research focus is on multimodal learning and post-quantum cryptography for GNSS systems.}

\biography{Fin Heuer}{M.Sc., completed his Master of Science degree in Computer Science at the Technical University in Braunschweig in 2022. He has joined the DLR (German Aerospace Center) Institute of Transportation Systems in 2015. His work primarily revolves around railway automation and safety as well as verification and validation processes within the automotive domain, applying rigorous standards to ensure the reliability and safety of automotive systems.}

\biography{Lennart Asbach}{received his Dipl.-Ing.~degree in Mechatronics from the Technical University Dresden. In 2009, he started as a Scientist at the DLR (German Aerospace Center) in Braunschweig, Germany. After his Team Lead position, Lennart Asbach is the Head of Department for transport research, automotive, and railway in Brunswick, Germany.}

\biography{Alexander Rügamer}{received his Dr.-Ing.~degree from the Friedrich-Alexander University (FAU) Erlangen-Nürnberg in 2024 and his Dipl.-Ing.~(FH) degree in Electrical Engineering from the University of Applied Sciences Würzburg-Schweinfurt in 2007. Since then, he has been working at the Fraunhofer Institute for Integrated Circuits IIS in the field of GNSS receiver development. In February 2012, he was promoted to Senior Engineer. From 2013 to 2023, he led a research group on secure GNSS receivers. Since 2024, he has been the head of the department of satellite-based localization. His main research interests focus on GNSS multi-band reception, integrated circuits, and immunity to interference.}

\biography{Felix Ott}{received his M.Sc.~degree in Computational Engineering from Friedrich-Alexander University (FAU) Erlangen-Nürnberg in 2019. He subsequently joined the Self-Learning Systems group within the Machine Intelligence department at the Fraunhofer Institute for Integrated Circuits IIS. In 2023, he was awarded his Ph.D.~from Ludwig-Maximilians University (LMU) Munich, where he conducted research in the Probabilistic Machine and Deep Learning group. He is currently serving as a postdoctoral researcher and project leader. His research focuses on representation learning, domain adaptation, and few-shot learning for GNSS-based interference monitoring.}

\section*{Abstract}

The accuracy and reliability of vehicle localization on roads are crucial for applications such as self-driving cars, toll systems, and digital tachographs. To achieve accurate positioning, vehicles typically use global navigation satellite system (GNSS) receivers to validate their absolute positions. However, GNSS-based positioning can be compromised by interference signals, necessitating the identification, classification, determination of purpose, and localization of such interference to mitigate or eliminate it. Recent approaches based on machine learning (ML) have shown superior performance in monitoring interference. However, their feasibility in real-world applications and environments has yet to be assessed. Effective implementation of ML techniques requires training datasets that incorporate realistic interference signals, including real-world noise and potential multipath effects that may occur between transmitter, receiver, and satellite in the operational area. Additionally, these datasets require reference labels. Creating such datasets is often challenging due to legal restrictions, as causing interference to GNSS sources is strictly prohibited. Consequently, the performance of ML-based methods in practical applications remains unclear. To address this gap, we describe a series of large-scale measurement campaigns conducted in real-world settings at two highway locations in Germany and the Seetal Alps in Austria, and in large-scale controlled indoor environments. We evaluate the latest supervised ML-based methods to report on their performance in real-world settings and present the applicability of pseudo-labeling for unsupervised learning. We demonstrate the challenges of combining datasets due to data discrepancies and evaluate outlier detection, domain adaptation, and data augmentation techniques to present the models' capabilities to adapt to changes in the datasets. \\
\textbf{Datasets and source code:} \url{https://gitlab.cc-asp.fraunhofer.de/darcy_gnss}
\section{Introduction}
\label{label_introduction}

\begin{figure}[!t]
    \centering
	\begin{minipage}[t]{0.61\linewidth}
        \centering
    	\includegraphics[width=1.0\linewidth]{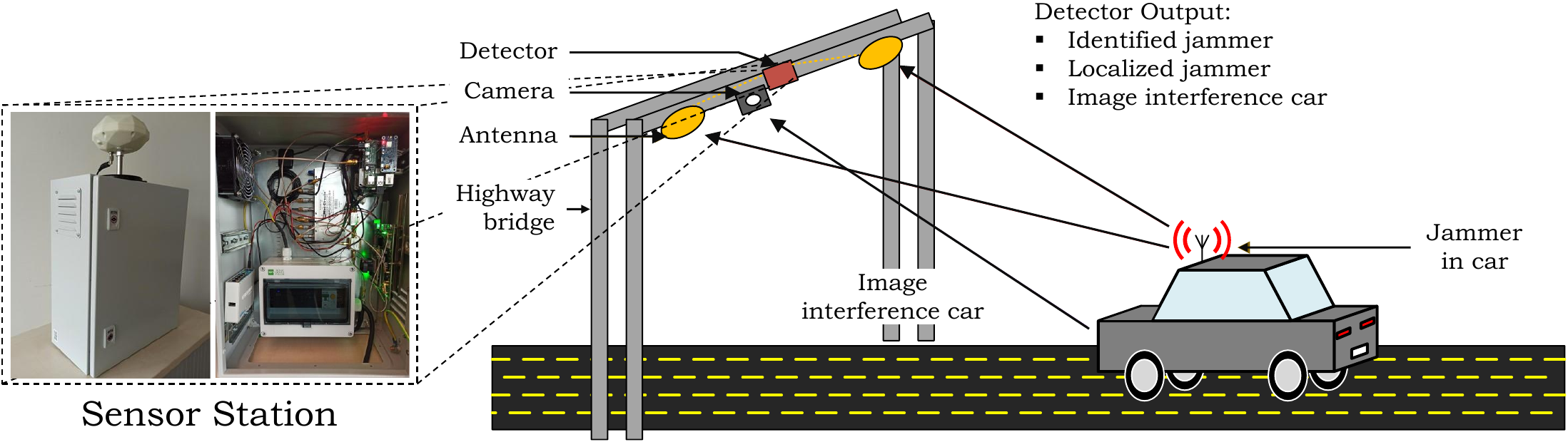}
    	\subcaption{Our GNSS sensor station, equipped with an antenna and camera, is mounted along a German highway to function as a jammer speed trap.}
    	\label{figure_intro_jammer_trap1}
    \end{minipage}
    \hfill
	\begin{minipage}[t]{0.16\linewidth}
        \centering
    	\includegraphics[width=1.0\linewidth]{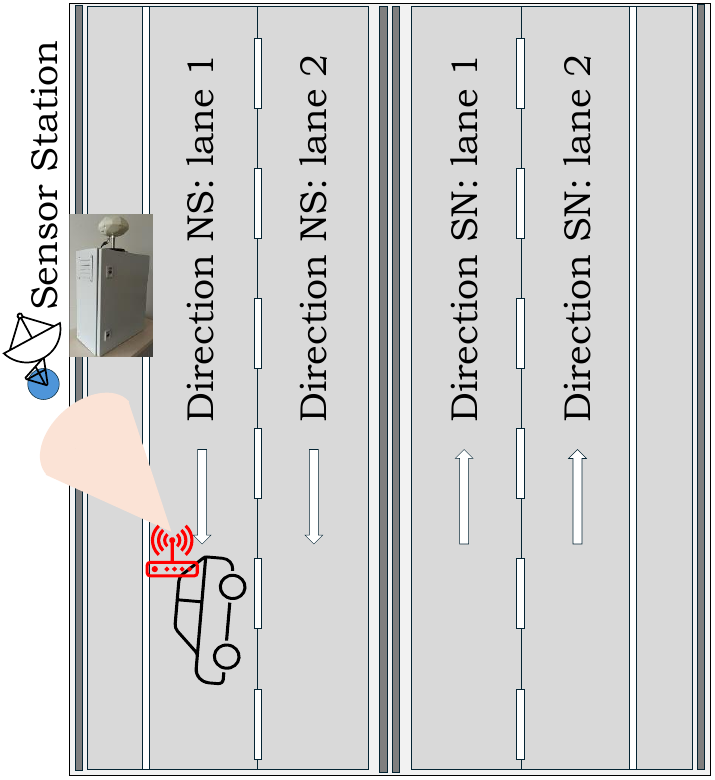}
    	\subcaption{Visualization of highway lanes.}
    	\label{figure_intro_jammer_trap2}
    \end{minipage}
    \hfill
	\begin{minipage}[t]{0.22\linewidth}
        \centering
    	\includegraphics[trim=30 0 110 0, clip, width=1.0\linewidth]{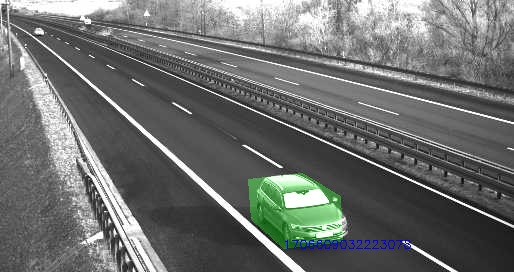}
    	\subcaption{Exemplary image of a detected jammer in a vehicle.}
    	\label{figure_intro_jammer_trap3}
    \end{minipage}
    \caption{Visualization of our jammer speed trap includes a GNSS receiver sensor station mounted along a highway to detect and classify interferences from vehicles, complemented by a camera.}
    \label{figure_intro_jammer_trap}
\end{figure}

First, we introduce the general concepts of GNSS interference detection, classification, and characterization. Next, we explain the motivation behind deploying a jammer speed trap along a highway and detail the challenges, particularly the real-world data discrepancies between different datasets. Finally, we summarize our contributions and provide an outlook of the manuscript.

\setlength{\intextsep}{6pt}
\setlength{\columnsep}{12pt}
\begin{wrapfigure}{R}{4.7cm}
    \begin{minipage}[b]{1.0\linewidth}
        \includegraphics[trim=0 0 0 10, clip, width=1.0\linewidth]{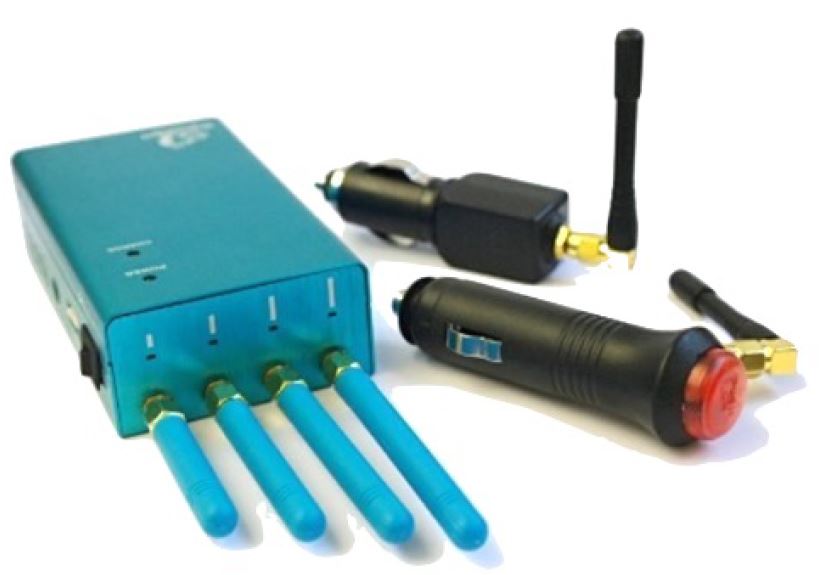}
        \caption{Mobile interference devices.}
        \label{figure_jamming_devices}
    \end{minipage}
\end{wrapfigure}
\paragraph{GNSS Interference Classification.} The accuracy of GNSS receivers' localization is significantly compromised by interference signals emanating from jamming devices~\citep{crespillo_ruiz,yuan_shen}, see Figure~\ref{figure_jamming_devices}. This issue has significantly intensified in recent years, as noted in the latest report by \cite{ainonline} due to the increased prevalence of cost-effective and easily accessible jamming devices~\citep{merwe_franco,mehr_minetto_dovis,miguel_chen_lo}. Consequently, it is essential to mitigate interference signals or eliminate the transmitter (the source of interference), specifically the jammer. To effectively eliminate an interference source, it is crucial to detect, classify, and localize it~\citep{raichur_ion_gnss,jdidi_brieger,ott_heublein_icl,raichur_heublein}. Accurate classification of the waveform of an interference signal, which serves as a unique fingerprint of a jammer, facilitates the determination of its purpose and, consequently, simplifies its localization and mitigation. Given the unpredictable emergence of novel jammer types, the objective is to develop resilient ML models. Currently, numerous techniques exist to achieve this, including classic methods~\citep{ferre_fuente,swinney_woods,gross_humphreys} and ML-based approaches~\citep{jdidi_brieger,raichur_ion_gnss,raichur_heublein,brieger_ion_gnss,merwe_franco,ott_heublein_icl}. To our knowledge, there is no publicly accessible reference dataset that facilitates the research, development, evaluation, and validation of effective and reliable techniques supporting practical GNSS interference management in a real-world environment. The evaluation of ML-based methods in the literature has been limited to synthetic data or controlled laboratory environments.

\paragraph{Applications \& Jammer Speed Trap.} Applications of GNSS interference classification encompass a wide range of fields, including air traffic control~\citep{ainonline,bartl_berglez,zuo_shi_jin,nasser_berz_gomez,nicola_falco}, crowdsourcing~\citep{joerger,strizic_akos_lo}, particularly from smartphones~\citep{raichur_ion_gnss}, maritime applications~\citep{gpspatron,marcos_konovaltsev}, and car rental control~\citep{bauernfeind_kraus}. Another significant application is the control and collection of tolls from trucks on highways. Most modern toll collection systems use electronic toll collection methods, which can be implemented through several technologies: On-board units are electronic devices installed in trucks that communicate with toll collection systems using global positioning system (GPS), dedicated short-range communication (DSRC), or global system for mobile communications (GSM) to track and report vehicle location and mileage. Satellite-based tolling systems use GPS technology to track the distance traveled by trucks on toll roads~\citep{european_gnss_agency}. This data is transmitted to a central system that calculates the toll based on the vehicle's route. However, GNSS can be interfered with by jamming devices, potentially preventing the systems from detecting the affected truck. The goal is to detect and classify the interference and localize the jammer in the affected truck. Figure~\ref{figure_intro_jammer_trap} presents the process of detecting, classifying, and localizing a jamming device in a vehicle. First, we built a sensor station (see Figure~\ref{figure_intro_jammer_trap1}) equipped with a GNSS receiver and an antenna, mounted along a German highway, to record GNSS snapshots. Next, the interference in the snapshots is detected and classified. Furthermore, from these snapshots, the traveling direction of the vehicle and the highway lane can be predicted~\citep{kocher_hansen}, refer to Figure~\ref{figure_intro_jammer_trap2}. With this information, the affected vehicle can be identified, as shown by the green segmentation in Figure~\ref{figure_intro_jammer_trap3}, which depicts a vehicle with various integrated handheld devices (see Figure~\ref{figure_jamming_devices}).

\begin{figure*}[!t]
    \centering
    \includegraphics[width=0.93\linewidth]{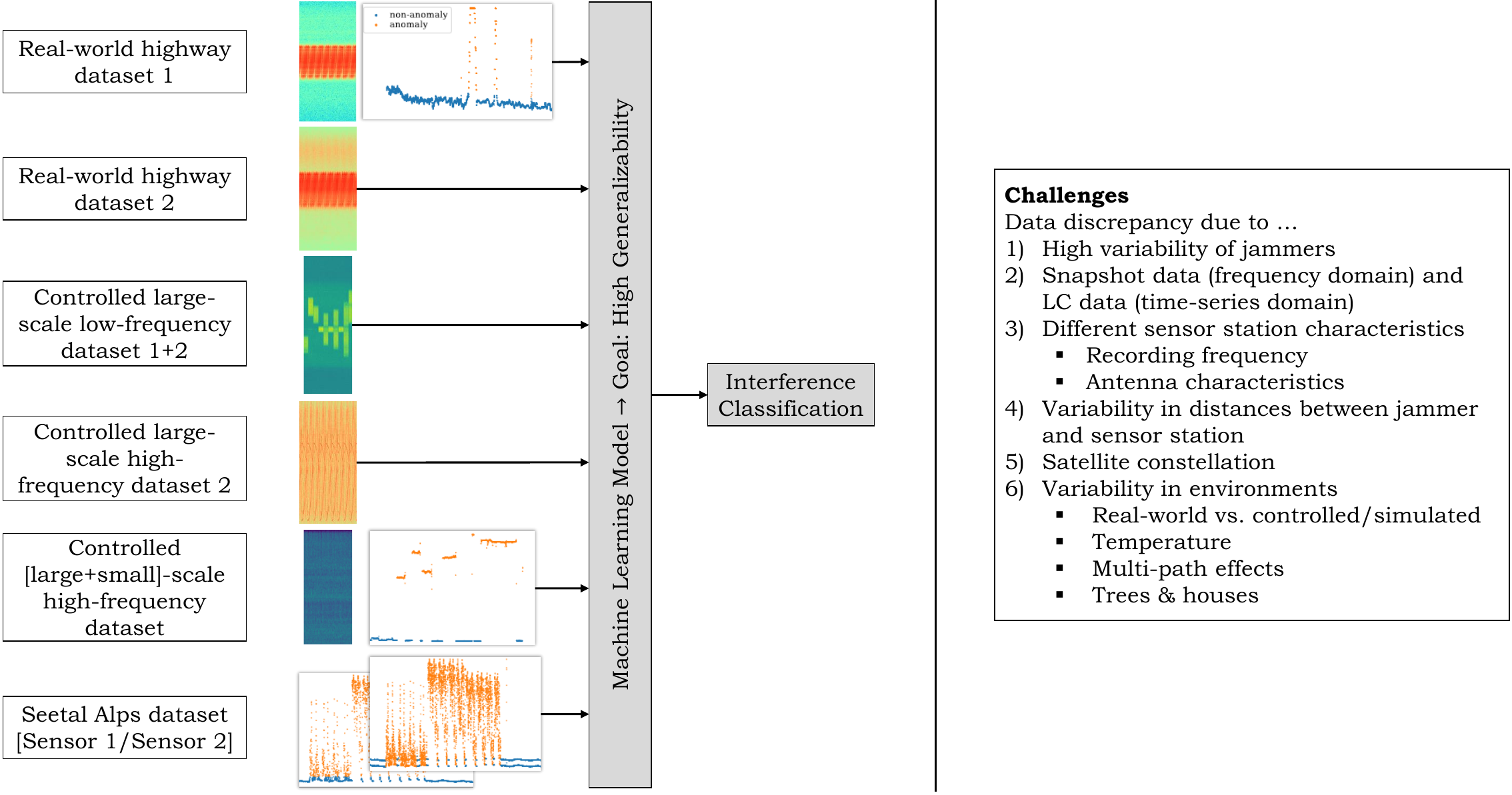}
    \caption{Overview of our GNSS interference classification pipeline addressing real-world data discrepancies. The high variability in data sources presents significant challenges in the interference classification task.}
    \label{figure_overview_data_discrepancy}
\end{figure*}

\setlength{\intextsep}{6pt}
\setlength{\columnsep}{12pt}
\begin{wrapfigure}{R}{4.6cm}
    \begin{minipage}[b]{1.0\linewidth}
        \includegraphics[trim=0 0 0 0, clip, width=1.0\linewidth]{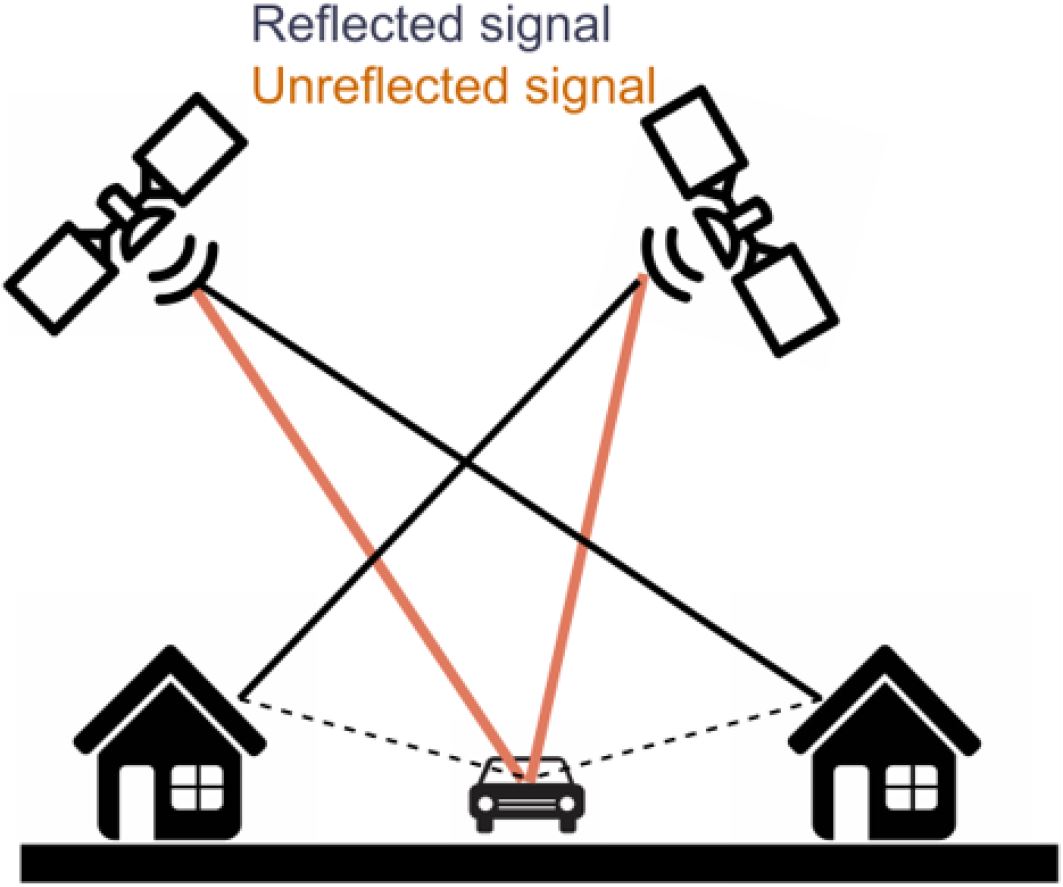}
        \caption{Multipath propagation.}
        \label{figure_multipath}
    \end{minipage}
\end{wrapfigure}
\paragraph{Real-World Data Discrepancies.} We collected several datasets for GNSS interference monitoring, comprising snapshot and low-cost (LC) data, as summarized in Figure~\ref{figure_overview_data_discrepancy}. Two real-world datasets were collected from different German highways. Additionally, datasets were collected in controlled indoor large-scale and small-scale environments using high-frequency and low-frequency antennas at the Fraunhofer Institute for Integrated Circuits IIS in Nuremberg, Germany. An LC dataset was also recorded in the Seetal Alps in Austria. The objective is to train an ML model to achieve high generalizability for interference classification. However, several challenges arise in training an ML model (see Figure~\ref{figure_overview_data_discrepancy}): (1) The data may exhibit substantial variance in jammer devices due to the increasing number of novel devices, leading to diverse interference characteristics. (2) Data is available in different formats; snapshot data is in the frequency domain, while LC data is in the time-series domain. (3) Sensors have varying characteristics such as recording frequency, antenna properties, and antenna directions. (4) Different distances between jammers and sensor stations result in varying intensities of interference, such as varying distances along highway lines. (5) Satellite constellations continuously vary. (6) Numerous environmental factors, such as indoor and outdoor settings, temperature changes, multipath effects (see Figure~\ref{figure_multipath} for multipath propagation among the sensor, jammer, and satellites), trees, and buildings, introduce additional variability. Therefore, it is paramount to assess the robustness of ML models in terms of their generalization capabilities against these variables. Ensuring the dependable deployment of ML models in real-world scenarios is of utmost importance. Assessing a model's robustness is critical and typically requires its application across multiple independent datasets. Given the constraints associated with collecting extensive datasets, focus often shifts to analyzing the distributions where the algorithm performs suboptimally~\citep{subbaswamy_adam,piratla}. Notably, shifts in datasets between training and testing sets commonly lead to a decline in model efficacy, prompting efforts to alleviate such distributional discrepancies~\citep{thamas_oberst,linden_forsberg}. One viable approach involves evaluating either the latent space or the predictive uncertainty \citep{wagh_wei}. However, model robustness is often assessed using synthetic datasets, leaving uncertainties regarding the model's resilience in the face of distributional shifts observed in authentic data scenarios~\citep{taori_dave}. Recording of datasets within controlled settings~\citep{ott_fusing} facilitates an exploration of the limitations of ML models. In the context of GNSS, particular emphasis is placed on the detection of non-line-of-sight~\citep{crespillo_ruiz} and multipath~\citep{yuan_shen} signals within specific environments~\citep{koiloth_achanta}, see Figure~\ref{figure_multipath}. However, datasets suitable for certain applications are often sparse, synthetic, or not publicly available. Our approach involves evaluating our models using data acquired in real-world highway environments, controlled indoor environments, and the Seetal Alps, with and without multipath effects.

\paragraph{Contributions.} The primary objective of this work is to evaluate both supervised and unsupervised ML methods for GNSS interference classification, taking into account real-world data discrepancies. We propose the following contributions: (1) We record several GNSS snapshot and LC datasets, including two datasets collected from highways, datasets from controlled indoor environments with different antennas, and a dataset recorded in the Seetal Alps. These datasets are described in detail, highlighting the challenges posed by real-world data discrepancies. (2) We analyze the influence of various interferences on the number of visible satellites and the positioning accuracy of smartphones in real-world scenarios. (3) We benchmark 20 supervised ML methods and cross-validate all snapshot datasets for GNSS interference classification to illustrate discrepancies in the datasets. (4) We benchmark 29 outlier detection methods to identify and classify LC GNSS interferences. (5) We benchmark 24 domain adaptation methods to facilitate adaptation from source to target domains, thereby reducing data discrepancies between LC datasets. (6) We propose a semi-supervised pseudo-labeling approach to effectively utilize only a subset of labeled data. (7) We classify the traveling direction of vehicles based on datasets from LC sensors.

\paragraph{Outlook.} The remainder of this paper is organized as follows. Section~\ref{label_related_work} provides an overview of the existing literature on GNSS interference classification, semi-supervised learning, and domain discrepancy. Section~\ref{label_methodology} details our ML model, pseudo-labeling method, and domain adaptation setup. Section~\ref{label_datasets} introduces several snapshot-based and LC-based datasets for GNSS interference classification. Section~\ref{label_experiments} presents an overview of all experiments conducted. Section~\ref{label_evaluation} summarizes the evaluation results, followed by the concluding remarks in Section~\ref{label_conclusion}.

\section{Related Work}
\label{label_related_work}

Initially, we present an overview of methods related to the detection, classification, and localization of GNSS interference (refer to Section~\ref{label_rw_gnss_monitoring}). Subsequently, we examine pertinent research on semi-supervised learning and pseudo-labeling (refer to Section~\ref{label_rw_pseudo_labeling}). Lastly, we provide a detailed background on domain shift and the discrepancies between synthetic data (i.e., data recorded in controlled environments) and real-world data (refer to Section~\ref{label_rw_domain_discrepancy}).

\subsection{GNSS Interference Monitoring}
\label{label_rw_gnss_monitoring}

\paragraph{Classical Approaches.} \cite{yang_kang} used automatic gain control (AGC) and an adaptive notch filter for interference detection and characterization according to interference types and power levels. While \cite{marcos_caizzone} characterized GNSS interferences in maritime applications, \cite{murrian_narula} proposed a situational awareness monitoring for the observation of terrestrial GNSS interference from low-earth orbit. For an overview of existing localization systems such as received signal strength (RSS), source angle of arrival (AoA), time difference of arrival (TDoA), and frequency difference of arrival (FDoA), refer to \cite{dempster_cetin}. \cite{biswas_cetin} proposed a particle filter for localizing a moving GNSS source using AoA and TDoA. \cite{borio_closas} used an orthonormal transformation to project the received GNSS samples into an appropriate transform domain. \cite{merwe_franco} introduced a low-cost GNSS interference monitoring, detection, and classification receiver. 

\paragraph{Machine Learning Techniques.} Recently, ML methods have been employed to analyze and process GNSS interference signals. \cite{swinney_woods} considered the jamming signal power spectral density, spectrogram, raw constellation, and histogram signal representations as images to utilize transfer learning from the imagery domain. They evaluated convolutional neural networks (CNNs), i.e., VGG16, support vector machines (SVMs), logistic regression, and random forests for the classification task. Inspired by \cite{swinney_woods}, we also employ spectrograms. \cite{ferre_fuente} introduced ML techniques, such as SVMs and CNNs, to classify jammer types in GNSS signals. Also, \cite{li_huang_lang,xu_ying_li} used a twin SVM-based method. \cite{mehr_dovis} classified chirp signals utilizing a CNN with representation transformed with the Wigner-Ville and Fourier method. \cite{ding_pham} exploited ML models in a single static (line-of-sight) propagation environment. \cite{gross_humphreys} applied a maximum likelihood method to determine whether a synthetic signal is affected by multipath or jamming. However, their method is not capable of classifying real-world multipath effects and jammer types. \cite{jdidi_brieger} proposed an unsupervised method to adapt to diverse, environment-specific factors, such as multipath effects, dynamics, and variations in signal strength. However, their pipeline requires human intervention to monitor the process and address uncertain classifications. In recent developments, few-shot learning~\citep{wang_yao,narwariya_malhotra,luo_si,tang_liu_long} has been used in the GNSS context to integrate new classes into a support set. \cite{ott_heublein_icl} proposed an uncertainty-based quadruplet loss aiming at a more continuous representation between positive and negative interference pairs. They utilize a dataset resembling a snapshot-based real-world dataset, featuring similar interference classes; however, with varying sampling rates. \cite{raichur_heublein} used the same dataset to adapt to novel interference classes through continual learning. \cite{brieger_ion_gnss} integrated both the spatial and temporal relationships between samples by using a joint loss function and a late fusion technique. Their dataset was acquired in a controlled indoor small-scale environment. As ResNet18~\citep{he_zhang} proved robust for interference classification, we also employ ResNet18 for feature extraction. \cite{raichur_ion_gnss} introduced a crowdsourcing method utilizing smartphone-based features to localize the source of detected interference.

\subsection{Semi-supervised Learning \& Pseudo-Labeling}
\label{label_rw_pseudo_labeling}

Semi-supervised learning~\citep{engelen_hoos} combines a small amount of labeled data with a large amount of unlabeled data during training. This technique leverages the labeled data to guide the learning process while also using the abundant unlabeled data to improve the model's performance. The key idea is to exploit the structure of the unlabeled data to enhance learning, making it particularly useful in scenarios where labeled data is scarce or expensive to obtain, such as in GNSS interference monitoring~\citep{berthelot_carlini}. Semi-supervised learning reduces the need for extensive labeling, which can be expensive and time-consuming, and can achieve better performance than using labeled data alone by harnessing patterns in the unlabeled data. In pseudo-labeling~\citep{bonilla_tan_yi}, the model itself generates labels for the unlabeled data based on its predictions, which are then used to train the model further. Pseudo-labeling increases the effective size of the training dataset by adding pseudo-labeled examples and helps the model generalize better by leveraging additional data. \cite{ivanov_scaramuzza} presented a one-class classification approach to classify GNSS interferences in aircraft data. To the best of our knowledge, no further related work on semi-supervised learning and pseudo-labeling exists in the context of GNSS monitoring. Virtual adversarial training (VAT), proposed by \cite{miyato_maeda_koyama}, is a regularization method based on a virtual adversarial loss -- a new measure of local smoothness of the conditional label distribution given input. Virtual adversarial loss is defined as the robustness of the conditional label distribution around each input data point against local perturbation. VAT defines the adversarial direction without label information and is hence applicable to semi-supervised learning. We propose a simple, yet effective pseudo-labeling approach based on several ML models and voting to select pseudo-labels. In the experiments, we compare our approach to VAT for classifying interferences and multipath scenarios.

\subsection{Domain Discrepancy \& Synthetic Data}
\label{label_rw_domain_discrepancy}

The issue of domain discrepancy in GNSS data has been a significant area of research due to the critical importance of accurate and reliable satellite-based positioning in various applications, ranging from navigation to geospatial sciences. Several studies have investigated the sources of discrepancies in GNSS data. \cite{misra_enge} identified atmospheric disturbances, satellite clock errors, and multipath effects as primary sources of GNSS data errors. \cite{gao_chen} explored the use of carrier-phase based differential GPS (DGPS) techniques to enhance positioning accuracy. Their findings highlighted the effectiveness of DGPS in reducing errors caused by satellite and receiver clock discrepancies. Moreover, the integration of GNSS data with other sensors' data has been proposed as a solution to mitigate discrepancies. The work by \cite{shin_el_sheimy} demonstrated the advantages of integrating inertial navigation systems (INS) with GNSS data. Their results indicated that such integration significantly improves positioning accuracy, especially in environments where GNSS signals are weak or obstructed. \cite{shen_chen_wang} utilized ML models to predict and correct GNSS errors. \cite{koiloth_achanta} benchmarked 14 ML methods to investigate the analysis of multipath data induced by sea waves, however, solely encompassed the consideration of elevation angle, signal strength, and pseudorange residuals. \cite{yuan_shen} focused on alleviating estimated multipath biases. \cite{crespillo_ruiz} proposed a logistic regression approach to mitigate biased estimation, thereby enhancing ML generalization. In assessing model robustness, rather than merely detecting outliers within datasets~\citep{linden_forsberg,piratla}, it is imperative to analyze the distributions among evaluation data~\citep{subbaswamy_adam}. While \cite{taori_dave} evaluated the distributional shift from synthetic to real data, our evaluation focuses on assessing model robustness across different scenarios within a controlled environment. \cite{wagh_wei} analyzed the predictive uncertainty between dataset shifts. \cite{wang_chen_guo} proposed a domain adaptation (DA) framework that leverages labeled GNSS data from a source domain to improve positioning accuracy in a target domain. Their framework used adversarial learning to minimize domain discrepancy and enhance the generalizability of GNSS models across different geographical regions. Despite significant advancements, challenges remain in fully addressing domain discrepancies in GNSS data.
\section{Methodology}
\label{label_methodology}

Firstly, we present a detailed notation of the classification task along with an in-depth description of the ML model employed for GNSS interference monitoring. Subsequently, we delineate the outlier detection task. Furthermore, we introduce a pseudo-labeling method aimed at reducing the proportion of labeled data required in the training set. Lastly, we elaborate on the domain adaptation task designed to facilitate adaptation to new datasets.

\paragraph{Notation.} For the snapshot data, let $\mathbf{X} \in \mathbb{R}^{h \times w}$ with entries $x_{i,j} \in [0, 255]$ with height $h$ and width $w$, represents an image from the image training set. The images are obtained by calculating the magnitude spectrogram of snapshots of IQ-samples. The image training set is a subset of the array $\mathcal{X} = \{\mathbf{X}_1,\ldots,\mathbf{X}_{n_X}\} \in \mathbb{R}^{n_X \times h \times w}$, where $n_X$ is the number of images in the training set. The goal is to predict an unknown class label $y \in \Omega$. We define the classification task as a multi-class task with one (or more) '\textit{no interference}' classes with varying background intensity and several '\textit{interference}' classes, e.g., '\textit{chirp}'~\citep{ott_heublein_icl}. For the low-cost classification task, we define a multivariate time-series $\mathbf{U} = \{\mathbf{u}_1,\ldots,\mathbf{u}_m\} \in \mathbb{R}^{m \times l}$ as an ordered sequence of $l \in \mathbb{N}$ streams with $\mathbf{u}_i = (u_{i,1},\ldots, u_{i,l}), i\in \{1,\ldots,m\}$, where $m \in \mathbb{N}$ is the length of the time-series. The multivariate time-series training set is a subset of the array $\mathcal{U} = \{\mathbf{U}_1,\ldots,\mathbf{U}_{n_U}\} \in \mathbb{R}^{n_U \times m \times l}$, where $n_U$ is the number of time-series. The aim of joint multivariate time-series classification task is to predict an unknown class label $y \in \Omega$ for a given multivariate time-series~\citep{ott_access}.

\begin{figure*}[!t]
    \centering
    \includegraphics[width=1.0\linewidth]{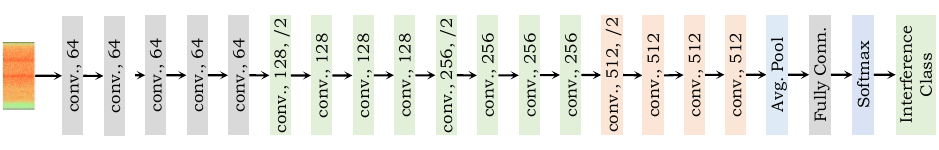}
    \caption{ResNet18 model consisting of several convolutional, average pooling, fully connected, and softmax layers.}
    \label{figure_resnet}
\end{figure*}

\paragraph{ML Model.} In the following, we describe the ML model that we utilize as a baseline for the classification task. The input is a snapshot image $\mathbf{X} \in \mathbb{R}^{h \times w}$ with height $h$ and width $w$. The output is the interference class label, where the number of class labels depends on the dataset. As ResNet18~\citep{he_zhang}, an abbreviation for residual network with 18 layers, proved robust for interference classification, we also employ ResNet18 for feature extraction with an additional softmax layer, see Figure~\ref{figure_resnet} for more details. This architecture addresses the challenge of training very deep neural networks, which tend to suffer from problems such as vanishing gradients. ResNet18 utilizes residual learning to ease the training of deep networks. Instead of learning a direct mapping, the network learns residuals (the difference between the input and the desired output). This is achieved through the introduction of shortcut connections (or \textit{skip connections}) that bypass one or more layers. ResNet18's relatively shallow depth makes it suitable for applications requiring a balance between computational efficiency and performance. For our benchmark, we compare the ResNet18 model to a large variety of ML models (refer to Section~\ref{section_exp_ml_models} for more details).

\paragraph{Outlier Detection.} In the context of GNSS interference classification, outlier detection involves identifying abnormal signal patterns that deviate from the typical GNSS signal behavior which may be caused by factors such as unusual interference sources, multipath effects, or hardware malfunctions. Accurate detection of these outliers is essential to maintain the integrity and reliability of GNSS systems, as undetected anomalies can lead to significant errors in positioning and timing applications~\citep{richards_scheer}. Methods for outlier detection in GNSS interference classification include statistical techniques, ML models, and advanced signal processing methods designed to recognize and isolate irregularities in GNSS data~\citep{pehlivan,qu_ding_xu_yu}. Outlier detection methods often use an outlier score for each data point $x_i$ based on its deviation from the expected distribution. A common measure used in statistical methods is the score: $\text{score}(x_i) = \frac{|x_i - \mu|}{\sigma}$, where $\mu$ and $\sigma$ are the mean and standard deviation of the data, respectively. Distance-based methods compute, e.g., $\text{score}(x_i) = \text{dist}\big(x_i, \text{kNN}(x_i)\big)$, where $\text{dist}$ is the distance metric (for example, the mean squared error) and $(x_i)$ denotes the distances to the $k$-nearest neighbors of $x_i$. Density-based methods compute the $\text{score} = \frac{1}{\text{density}(x_i)}$ based on the density around $x_i$. We benchmark a large variety of outlier detection methods on the low-cost GNSS datasets. Therefore, we utilize \textit{PyOD}, proposed by \cite{pyod}, which is an open-source Python toolbox for performing scalable outlier detection on multivariate data that provides access to a wide range of outlier detection algorithms (ranging from outlier ensembles techniques to recent neural network-based approaches). For an overview, refer to Section~\ref{section_exp_outlier_detection}.

\paragraph{Semi-Supervised Learning.} In \textit{supervised learning}, a set of data points consisting of an input $x$ and a corresponding output value $y$ is provided. The objective is to construct a classifier or regressor capable of estimating the output value for previously unseen inputs. In \textit{unsupervised learning}, no specific output value is given. Instead, the aim is to infer some underlying structure from the inputs. \textit{Semi-supervised learning} involves using both labeled and unlabeled data to perform certain learning tasks \citep{engelen_hoos}. Recent research in this area has aligned with general trends in machine learning, with significant focus on neural network-based models and generative learning. A necessary condition for semi-supervised learning is that the underlying \textit{marginal data distribution} $p(x)$ over the input space contains information about the \textit{posterior distribution} $p(y|x)$. If this condition is met, it may be possible to use unlabeled data to gain insights into $p(x)$ and, consequently, into $p(y|x)$~\citep{engelen_hoos}. Additionally, semi-supervised learning research often includes the cluster assumption, which posits that data points within the same cluster belong to the same class~\citep{chapelle_chi_zien}.

\begin{figure*}[!t]
    \centering
    \includegraphics[width=0.45\linewidth]{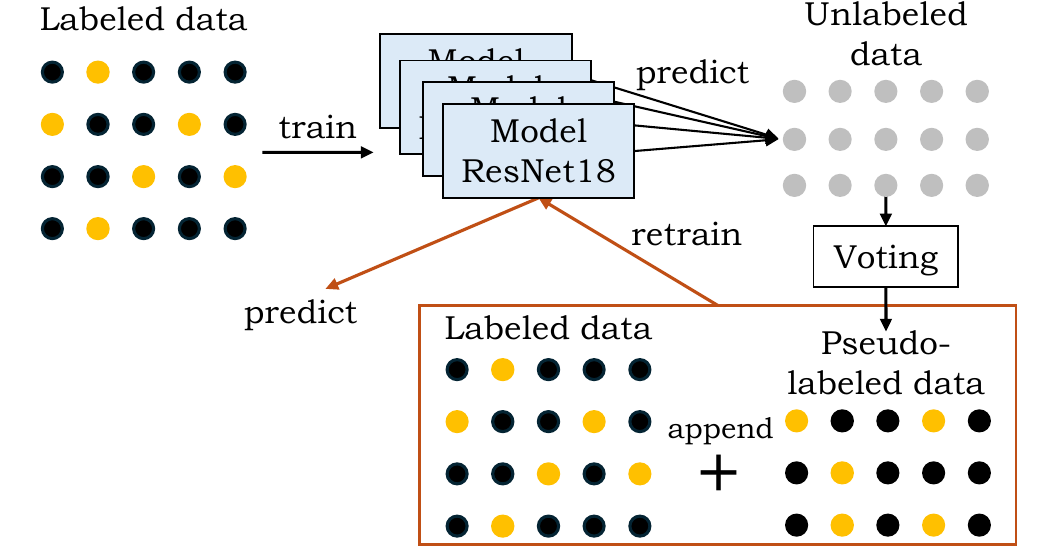}
    \caption{Pseudo-labeling consisting of four ResNet18 models that vote based on predictions for the unlabeled dataset. The pseudo-labeled data is appended to the labeled dataset to retrain the models~\citep{kim_lee_tama}.}
    \label{figure_pseudo_labeling}
\end{figure*}

\paragraph{Pseudo-Labeling.} In the context of semi-supervised learning, pseudo-labeling leverages a small set of labeled samples alongside a large set of unlabeled samples. This technique involves applying pseudo-labels to the unlabeled samples using a model trained on both the labeled samples and any previously pseudo-labeled samples, iteratively repeating this process in a self-training cycle~\citep{bonilla_tan_yi}. The objective is to utilize only a subset of labeled data from the dataset, reflecting real-world scenarios where much of the data may be unlabeled. By incorporating additional unlabeled data, we aim to develop robust models capable of handling data discrepancies between real-world and controlled environments and across different highway locations. In this paper, we propose a pseudo-labeling approach based on multiple ML models. Figure~\ref{figure_pseudo_labeling} illustrates our pipeline. Multiple models $M$ (here, $M=4$) are trained on a labeled subset (represented by orange dots) of the dataset and subsequently make predictions on the remaining unlabeled data (represented by black dots). A voting mechanism, based on the softmax output meeting a certain threshold, is employed to generate pseudo-labels. These pseudo-labeled data are then appended to the labeled set, and the models are retrained. Ultimately, a single ML model makes a prediction based on the final model weights. As baseline comparison, we utilize virtual adversarial training (VAT), as proposed by \cite{miyato_maeda_koyama}. VAT is a regularization method based on virtual adversarial loss. Unlike adversarial training, VAT determines the adversarial direction without label information, making it suitable for semi-supervised learning.

\begin{figure*}[!t]
    \centering
    \includegraphics[width=0.6\linewidth]{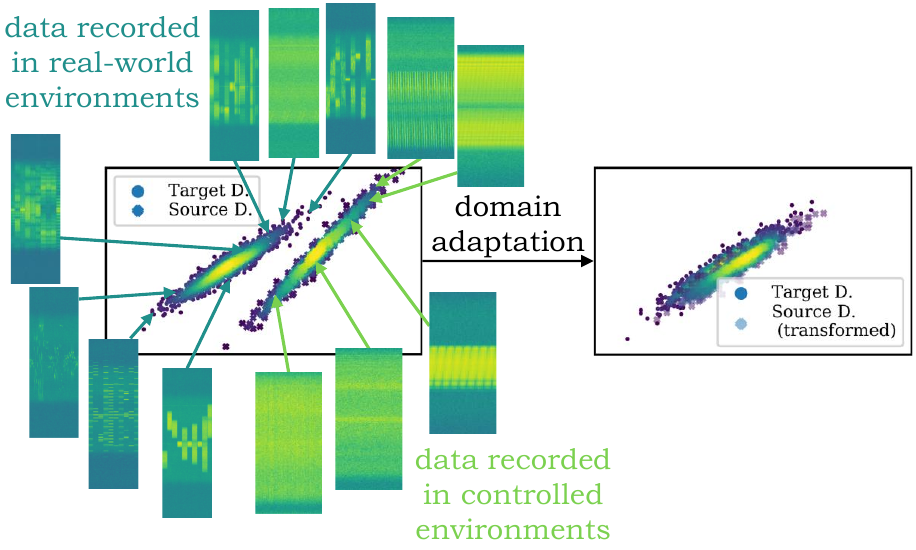}
    \caption{The left plot shows the discrepancy of features between data recorded in real-world environments (target domain) and data recorded in controlled environments (source domain). Domain adaptation methods reduce the discrepancy between source and target domains by minimizing the distance between their domain distributions. Refer to \cite{ott_acmmm}, for more details.}
    \label{figure_domain_adaptation}
\end{figure*}

\paragraph{Domain Adaptation.} Traditional ML algorithms assume that the training and test datasets are \textit{independent and identically distributed} (i.i.d.). Consequently, supervised ML performs effectively only when the test data originates from the same distribution as the training data. However, in real-world scenarios, data often varies across time and space, making this assumption rarely valid in practice~\citep{ott_acmmm}. Domain adaptation (DA) addresses this issue by compensating for the \textit{domain shift} between the \textit{source} and \textit{target} domains. The \textit{source} domain is where the model is initially trained, typically featuring well-annotated and abundant data. In contrast, the \textit{target} domain is where the model is intended to be applied or tested, often exhibiting a different data distribution due to variations in feature space, data characteristics, or environmental factors. In the context of GNSS applications, such distribution shifts occur due to atmospheric disturbances, satellite clock errors, and multipath effects~\citep{misra_enge,gao_chen,shen_chen_wang,yuan_shen}. For our speed jammer trap, domain shifts arise between data from different stations because the environments at these stations vary. Additionally, there is a domain shift between data recorded in controlled environments, where arbitrary interferences can be modeled, and real-world environments (i.e., highways), where recording data with jammers is challenging or prohibited. Figure~\ref{figure_domain_adaptation} illustrates an exemplary shift between the feature distributions of data recorded in real-world environments and those recorded in controlled environments, highlighting differences in jammer characteristics, image sizes, and sampling rates. The objective of DA is to mitigate the domain shift by minimizing the discrepancy between the feature embeddings of the source and target domains. This is achieved by reducing the distance between their feature distributions. Numerous DA methods have been developed, focusing on specific loss functions to compute an optimal distance between the source and target domain features. In our study, we evaluate 24 DA methods on our low-cost GNSS datasets. Section~\ref{section_exp_da_models} provides a detailed description of these methods.

\section{Datasets}
\label{label_datasets}

\begin{table}[t!]
\begin{center}
    \caption{Overview of datasets, their recording antenna, number of classes, and the number of samples in the train and test sets.}
    \label{table_overview_datasets}
    \small \begin{tabular}{ p{1.3cm} | p{0.5cm} | p{0.5cm} | p{0.5cm} | p{0.5cm} | p{0.5cm} }
    \multicolumn{1}{c|}{\textbf{Dataset}} & \multicolumn{1}{c|}{\textbf{Antenna}} & \multicolumn{1}{c|}{\textbf{Number Classes}} & \multicolumn{1}{c|}{\textbf{Train Set}} & \multicolumn{1}{c|}{\textbf{Test Set}} & \multicolumn{1}{c}{\textbf{Total}} \\ \hline
    \multicolumn{1}{l|}{Real-world highway dataset 1} & \multicolumn{1}{l|}{High-frequency} & \multicolumn{1}{r|}{11} & \multicolumn{1}{r|}{158,262} & \multicolumn{1}{r|}{39,565} & \multicolumn{1}{r}{39,565} \\
    \multicolumn{1}{l|}{Real-world highway dataset 2} & \multicolumn{1}{l|}{High-frequency} & \multicolumn{1}{r|}{9} & \multicolumn{1}{r|}{13,184} & \multicolumn{1}{r|}{3,296} & \multicolumn{1}{r}{16,480} \\
    \multicolumn{1}{l|}{Controlled large-scale dataset 1} & \multicolumn{1}{l|}{Low-frequency} & \multicolumn{1}{r|}{7} & \multicolumn{1}{r|}{33,544} & \multicolumn{1}{r|}{8,424} & \multicolumn{1}{r}{41,968} \\
    \multicolumn{1}{l|}{Controlled large-scale dataset 2} & \multicolumn{1}{l|}{Low-frequency} & \multicolumn{1}{r|}{7} & \multicolumn{1}{r|}{77,116} & \multicolumn{1}{r|}{19,328} & \multicolumn{1}{r}{96,444} \\
    \multicolumn{1}{l|}{Controlled large-scale dataset 2} & \multicolumn{1}{l|}{High-frequency} & \multicolumn{1}{r|}{7} & \multicolumn{1}{r|}{847,974} & \multicolumn{1}{r|}{211,993} & \multicolumn{1}{r}{1,059,967} \\
    \multicolumn{1}{l|}{Controlled [large+small]-scale dataset} & \multicolumn{1}{l|}{High-frequency} & \multicolumn{1}{r|}{9} & \multicolumn{1}{r|}{5,381} & \multicolumn{1}{r|}{1,345} & \multicolumn{1}{r}{6,726} \\
    \multicolumn{1}{l|}{Low-cost datasets} & \multicolumn{1}{l|}{Low-cost} & \multicolumn{1}{r|}{2} & \multicolumn{1}{r|}{68,226} & \multicolumn{1}{r|}{17,059} & \multicolumn{1}{r}{85,285} \\ \hline
    \multicolumn{3}{r|}{\textbf{Total}} & \multicolumn{1}{r|}{1,203,687} & \multicolumn{1}{r|}{301,010} & \multicolumn{1}{r}{1,504,697} \\
    \end{tabular}
\end{center}
\end{table}

To address this gap in knowledge, this paper presents a comprehensive series of large-scale measurement campaigns. These campaigns were conducted in real-world settings with special permission from legal authorities to overcome specific limitations and in large-scale controlled indoor environments. In this section, we provide a detailed description of all our GNSS datasets, which we have made publicly available, refer to \cite{ott_heublein_dataset}. Table~\ref{table_overview_datasets} provides an overview of all datasets. Initially, a sensor station with a high-frequency antenna, as depicted in Figure~\ref{figure_intro_jammer_trap1}, was mounted along a German highway to record data from eight different jammer types placed in a car (see Section~\ref{label_data_rw_1}). Subsequently, two sensor stations equipped with high-frequency antennas were installed next to a different highway, with several hundred meters separating the stations (see Section~\ref{label_data_rw_2}). Additionally, four datasets were recorded in controlled indoor environments. Two datasets were captured using a low-frequency antenna in the Fraunhofer IIS L.I.N.K.~test and application center (see Section~\ref{label_data_contr_lf1} and Section~\ref{label_data_contr_lf2}). The first dataset includes various multipath effects, while the second dataset was simultaneously recorded with a high-frequency antenna (see Section~\ref{label_data_contr_hf2}), facilitating a comparison of interferences between different antennas. Furthermore, a high-frequency antenna was placed both in the L.I.N.K.~center and in a small indoor lab, and these datasets were combined (see Section~\ref{label_data_contr_ls_hf}). Moreover, several low-cost GNSS datasets were recorded, including simultaneous recordings with the real-world highway dataset 1 and the controlled [large+small]-scale dataset. An additional low-cost GNSS dataset was recorded in the Seetal Alps in Austria, where various jamming devices were placed in a car to evaluate different driving directions (see Section~\ref{label_data_low_cost}). In total, our dataset comprises 1,504,697 snapshot and LC samples, which are divided into a 80\%/20\% train/test set. In the subsequent sections, we describe these datasets in greater detail and analyze the data discrepancy by examining the LC feature embeddings in Section~\ref{label_data_discrepancy}.

\subsection{Real-World Highway Dataset 1: High-Frequency Antenna}
\label{label_data_rw_1}

\begin{figure}[!t]
    \centering
	\begin{minipage}[t]{0.081\linewidth}
        \centering
    	\includegraphics[trim=218 40 118 44, clip, width=1.0\linewidth]{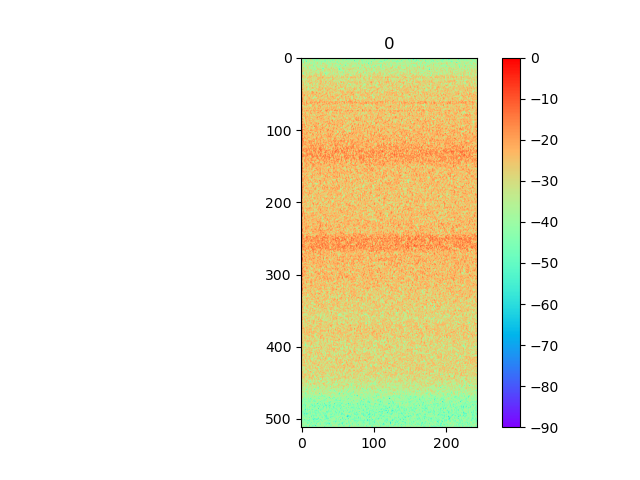}
    	\subcaption{None (low intensity)}
    	\label{figure_highway1_1}
    \end{minipage}
    \hfill
	\begin{minipage}[t]{0.081\linewidth}
        \centering
    	\includegraphics[trim=218 40 118 44, clip, width=1.0\linewidth]{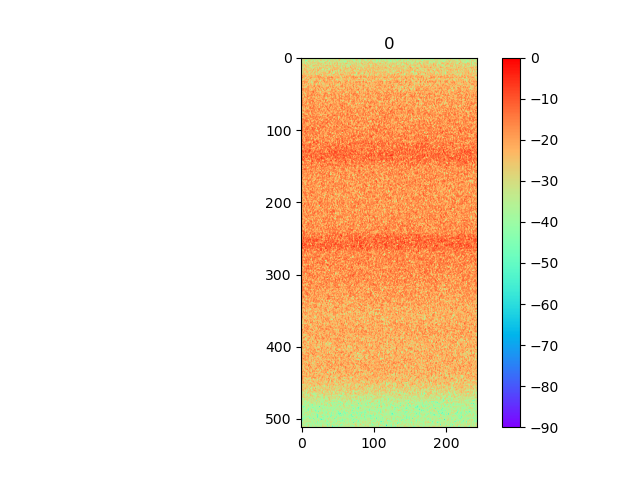}
    	\subcaption{None (medium intensity)}
    	\label{figure_highway1_2}
    \end{minipage}
    \hfill
	\begin{minipage}[t]{0.081\linewidth}
        \centering
    	\includegraphics[trim=218 40 118 44, clip, width=1.0\linewidth]{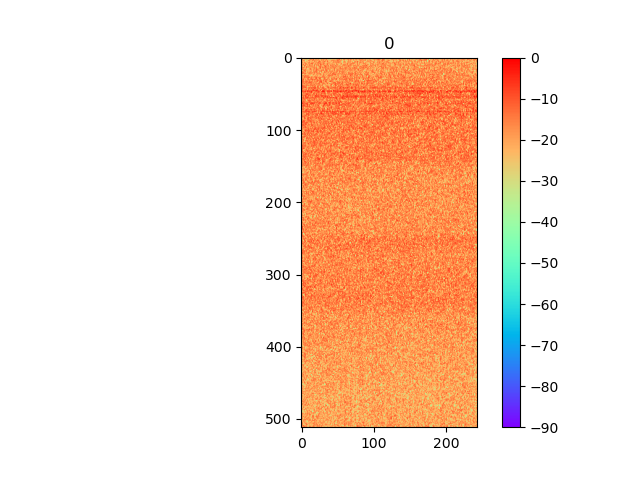}
    	\subcaption{None (high intensity)}
    	\label{figure_highway1_3}
    \end{minipage}
    \hfill
	\begin{minipage}[t]{0.081\linewidth}
        \centering
    	\includegraphics[trim=218 40 118 44, clip, width=1.0\linewidth]{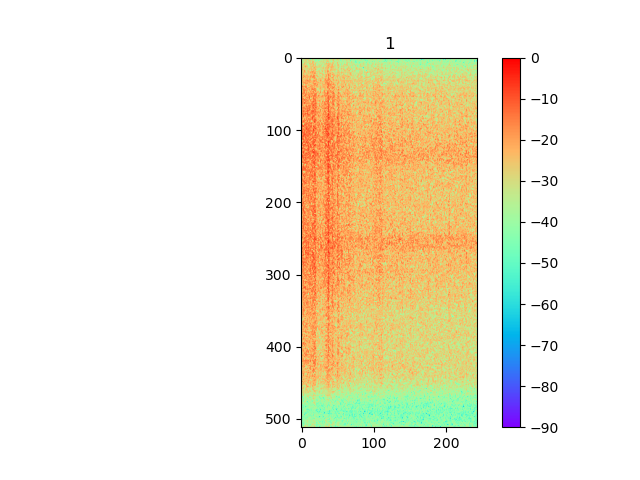}
    	\subcaption{Pulsed}
    	\label{figure_highway1_4}
    \end{minipage}
    \hfill
	\begin{minipage}[t]{0.081\linewidth}
        \centering
    	\includegraphics[trim=218 40 118 44, clip, width=1.0\linewidth]{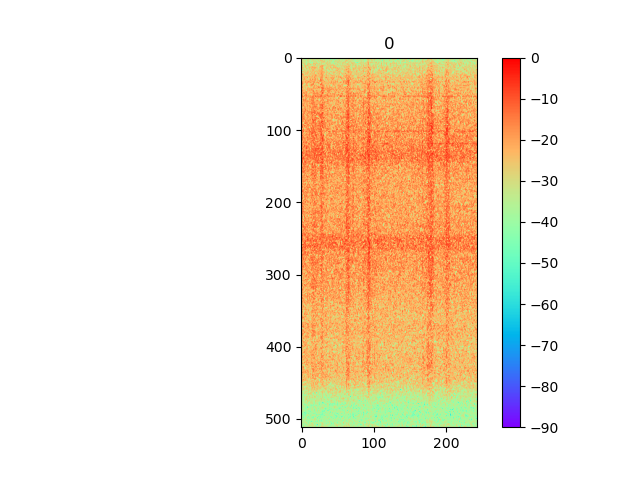}
    	\subcaption{Pulsed}
    	\label{figure_highway1_5}
    \end{minipage}
    \hfill
	\begin{minipage}[t]{0.081\linewidth}
        \centering
    	\includegraphics[trim=218 40 118 44, clip, width=1.0\linewidth]{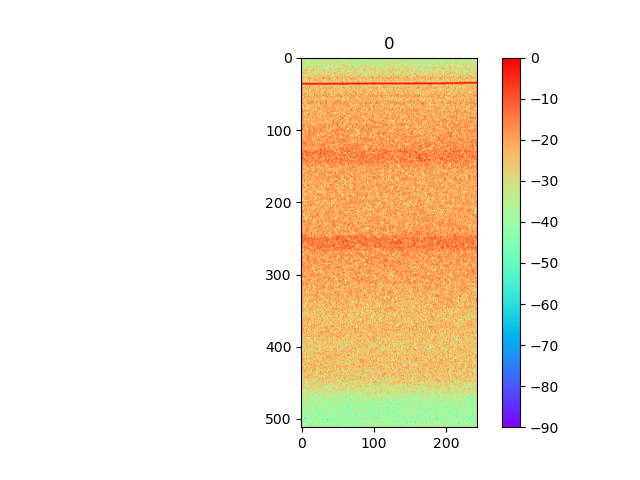}
    	\subcaption{Out-of-band tone}
    	\label{figure_highway1_6}
    \end{minipage}
    \hfill
	\begin{minipage}[t]{0.081\linewidth}
        \centering
    	\includegraphics[trim=218 40 118 44, clip, width=1.0\linewidth]{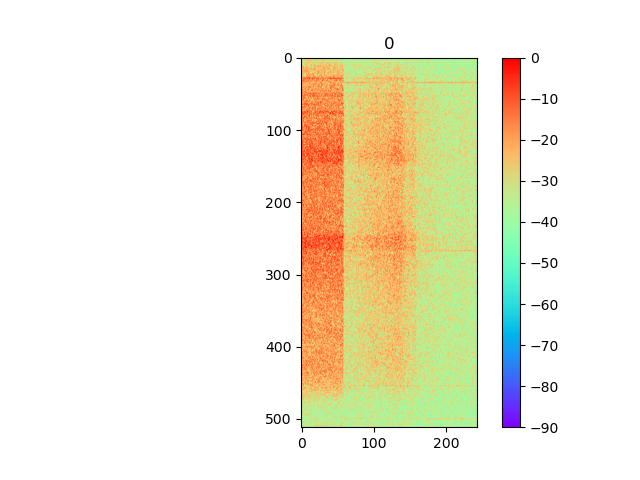}
    	\subcaption{Noise}
    	\label{figure_highway1_7}
    \end{minipage}
    \hfill
	\begin{minipage}[t]{0.081\linewidth}
        \centering
    	\includegraphics[trim=218 40 118 44, clip, width=1.0\linewidth]{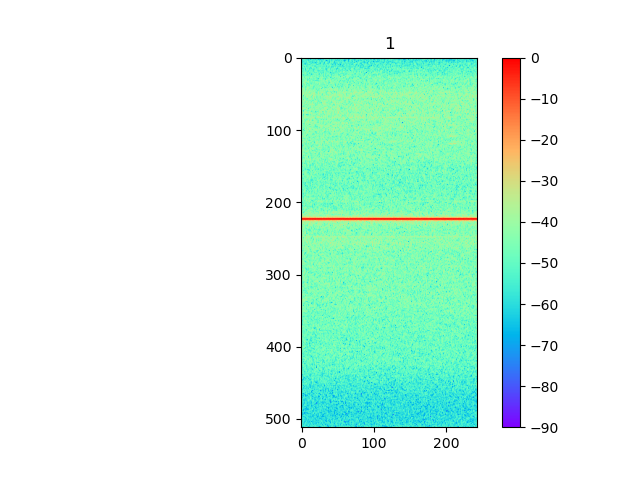}
    	\subcaption{Tone}
    	\label{figure_highway1_8}
    \end{minipage}
    \hfill
	\begin{minipage}[t]{0.081\linewidth}
        \centering
    	\includegraphics[trim=218 40 118 44, clip, width=1.0\linewidth]{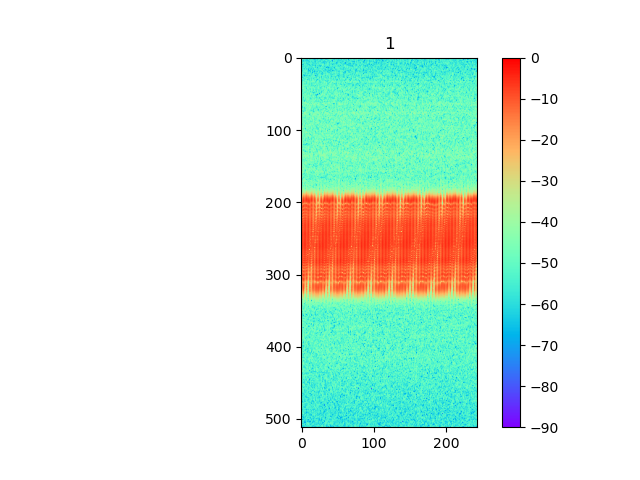}
    	\subcaption{Chirp}
    	\label{figure_highway1_9}
    \end{minipage}
    \hfill
	\begin{minipage}[t]{0.081\linewidth}
        \centering
    	\includegraphics[trim=218 40 118 44, clip, width=1.0\linewidth]{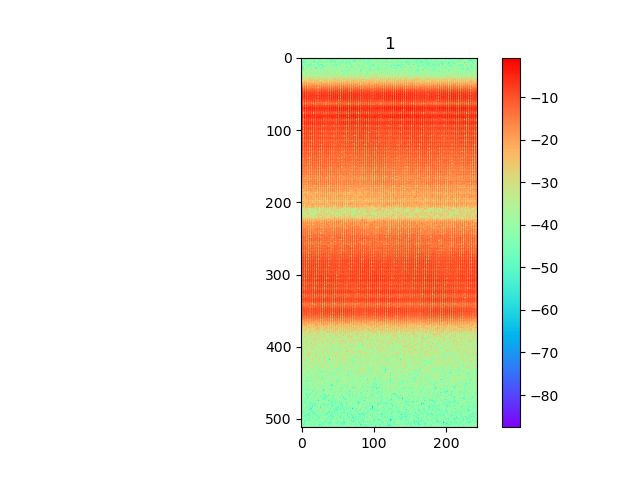}
    	\subcaption{Two chirps}
    	\label{figure_highway1_10}
    \end{minipage}
    \hfill
	\begin{minipage}[t]{0.081\linewidth}
        \centering
    	\includegraphics[trim=218 40 118 44, clip, width=1.0\linewidth]{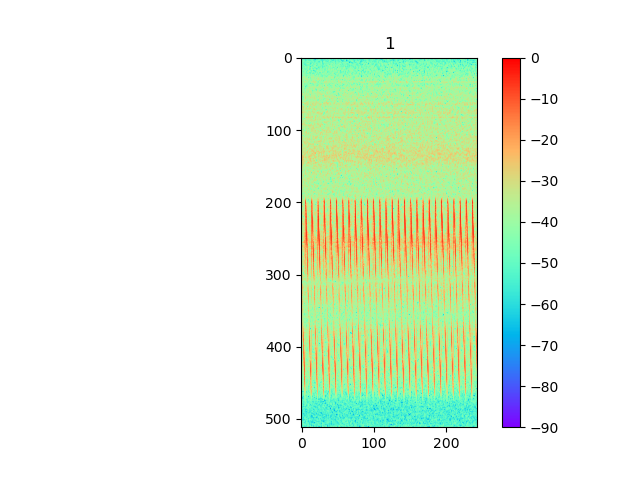}
    	\subcaption{Chirp}
    	\label{figure_highway1_11}
    \end{minipage}
    \caption{Exemplary spectrogram samples recorded on a bridge along a real-world German highway (first highway). The x-axis shows the time in $\text{ms}$. The y-axis shows the frequency in $\text{MHz}$.}
    \label{figure_highway1}
\end{figure}

For the first dataset~\citep{ott_heublein_icl}, a sensor station (see Figure~\ref{figure_recording_setup1} and Figure~\ref{figure_recording_setup2}) was positioned on a bridge along a highway to capture short, wideband snapshots in both the E1 and E6 GNSS bands. The setup records 20\,ms raw IQ (in-phase and quadrature-phase) snapshots triggered by energy, with a sample rate of 62.5\,MHz, an analog bandwidth of 50\,MHz, and an 8\,bit resolution. At certain frequencies, the GPS/Galileo or GLONASS signals are discernible as slight increases in the spectrum. The spectrogram images have dimensions of $512 \times 243$. Experts manually analyzed the data streams by thresholding C/N0 (carrier-to-noise density ratio) and AGC (automatic gain control) values, resulting in the manual labeling of the snapshots into 11 classes: Classes 0 to 2 represent samples with no interferences (Figure~\ref{figure_highway1_1} to Figure~\ref{figure_highway1_3}), distinguished by variations in background intensity, while classes 3 to 10 contain different interferences (Figure~\ref{figure_highway1_4} to Figure~\ref{figure_highway1_11})~\citep{ott_heublein_icl}. The dataset comprises a total of 39,565 samples, with 253 samples containing interference and 197,574 samples containing no interference. The dataset is highly imbalanced, presenting a challenge in adapting to positive class labels with only a limited number of samples available. The dataset is partitioned into an 80\% training set and a 20\% test set split (balanced across the classes).

\subsection{Real-World Highway Dataset 2: High-Frequency Antenna}
\label{label_data_rw_2}

\begin{figure}[!t]
    \centering
	\begin{minipage}[t]{0.10\linewidth}
        \centering
    	\includegraphics[trim=160 30 160 16, clip, width=1.0\linewidth]{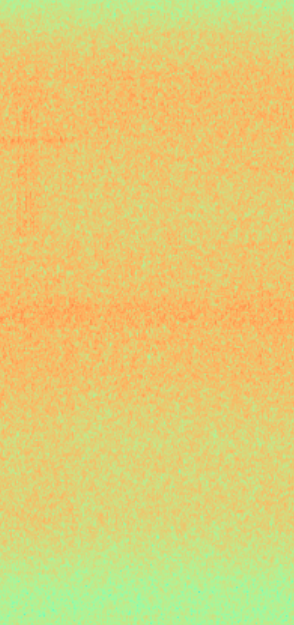}
    	\subcaption{None}
    	\label{figure_highway2_1}
    \end{minipage}
    \hfill
	\begin{minipage}[t]{0.10\linewidth}
        \centering
    	\includegraphics[trim=160 30 160 16, clip, width=1.0\linewidth]{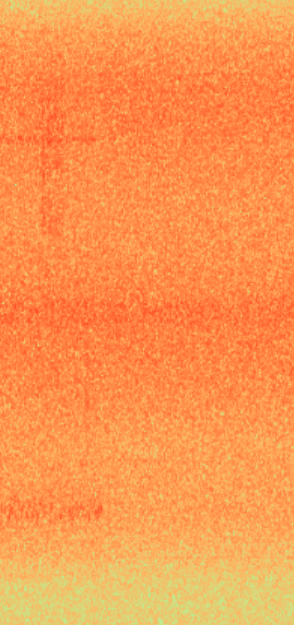}
    	\subcaption{None}
    	\label{figure_highway2_2}
    \end{minipage}
    \hfill
	\begin{minipage}[t]{0.10\linewidth}
        \centering
    	\includegraphics[trim=160 30 160 16, clip, width=1.0\linewidth]{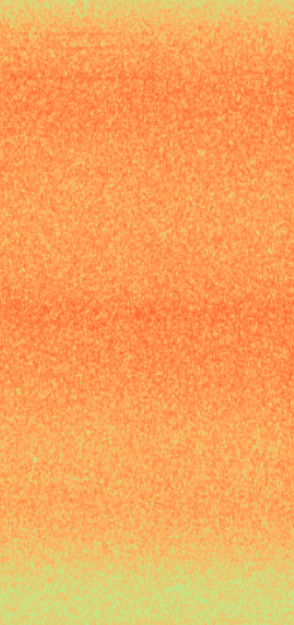}
    	\subcaption{None}
    	\label{figure_highway2_3}
    \end{minipage}
    \hfill
	\begin{minipage}[t]{0.10\linewidth}
        \centering
    	\includegraphics[trim=160 30 160 16, clip, width=1.0\linewidth]{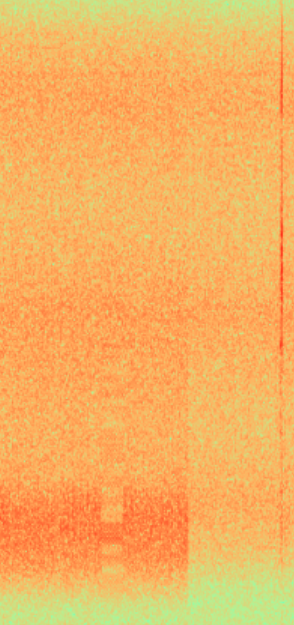}
    	\subcaption{None}
    	\label{figure_highway2_4}
    \end{minipage}
    \hfill
	\begin{minipage}[t]{0.10\linewidth}
        \centering
    	\includegraphics[trim=160 30 160 16, clip, width=1.0\linewidth]{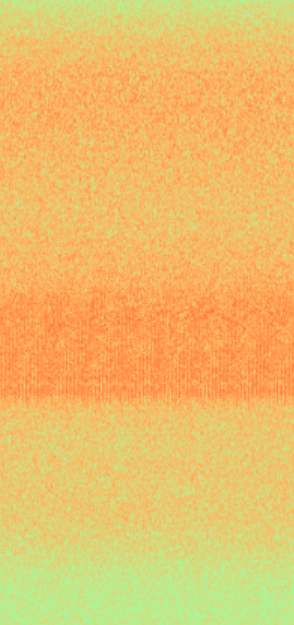}
    	\subcaption{Chirp, high distance}
    	\label{figure_highway2_5}
    \end{minipage}
    \hfill
	\begin{minipage}[t]{0.10\linewidth}
        \centering
    	\includegraphics[trim=160 30 160 16, clip, width=1.0\linewidth]{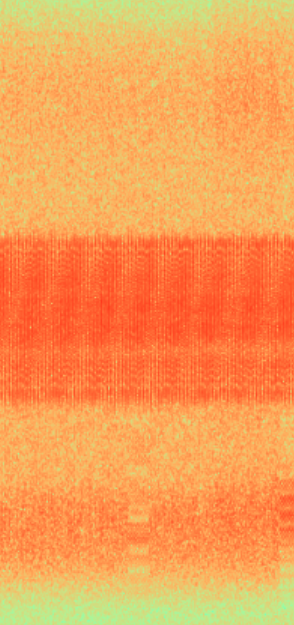}
    	\subcaption{Chirp, medium distance}
    	\label{figure_highway2_6}
    \end{minipage}
    \hfill
	\begin{minipage}[t]{0.10\linewidth}
        \centering
    	\includegraphics[trim=160 30 160 16, clip, width=1.0\linewidth]{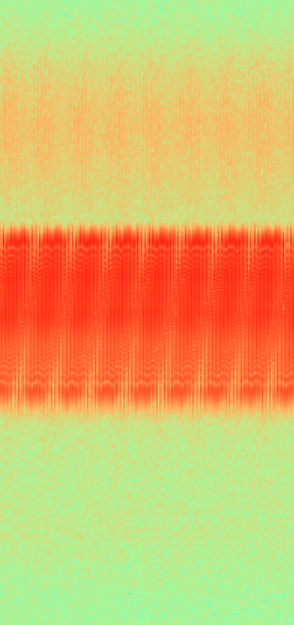}
    	\subcaption{Chirp, small distance}
    	\label{figure_highway2_7}
    \end{minipage}
    \hfill
	\begin{minipage}[t]{0.10\linewidth}
        \centering
    	\includegraphics[trim=160 30 160 16, clip, width=1.0\linewidth]{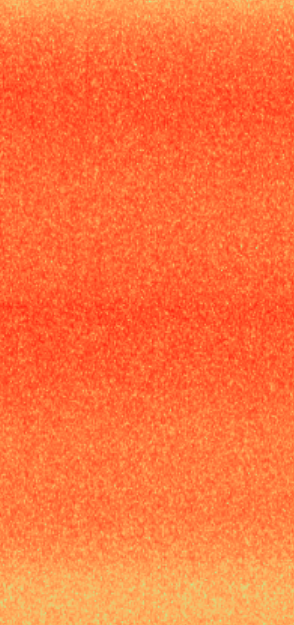}
    	\subcaption{Cigarette lighter 1}
    	\label{figure_highway2_8}
    \end{minipage}
    \hfill
	\begin{minipage}[t]{0.10\linewidth}
        \centering
    	\includegraphics[trim=160 30 160 16, clip, width=1.0\linewidth]{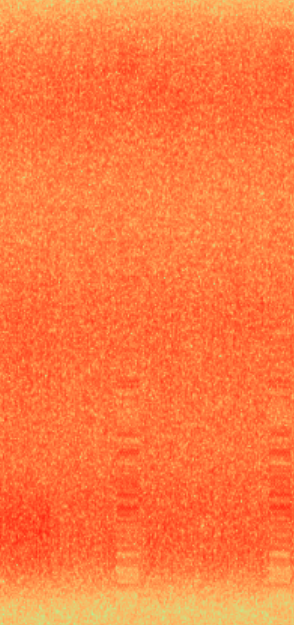}
    	\subcaption{Cigarette lighter 2}
    	\label{figure_highway2_9}
    \end{minipage}
    \caption{Exemplary spectrogram samples recorded on a bridge along a real-world German highway (second highway). The x-axis shows the time in $\text{ms}$. The y-axis shows the frequency in $\text{MHz}$.}
    \label{figure_highway2}
\end{figure}

\begin{figure}[!t]
    \centering
    \includegraphics[trim=0 30 0 20, clip, width=0.75\linewidth]{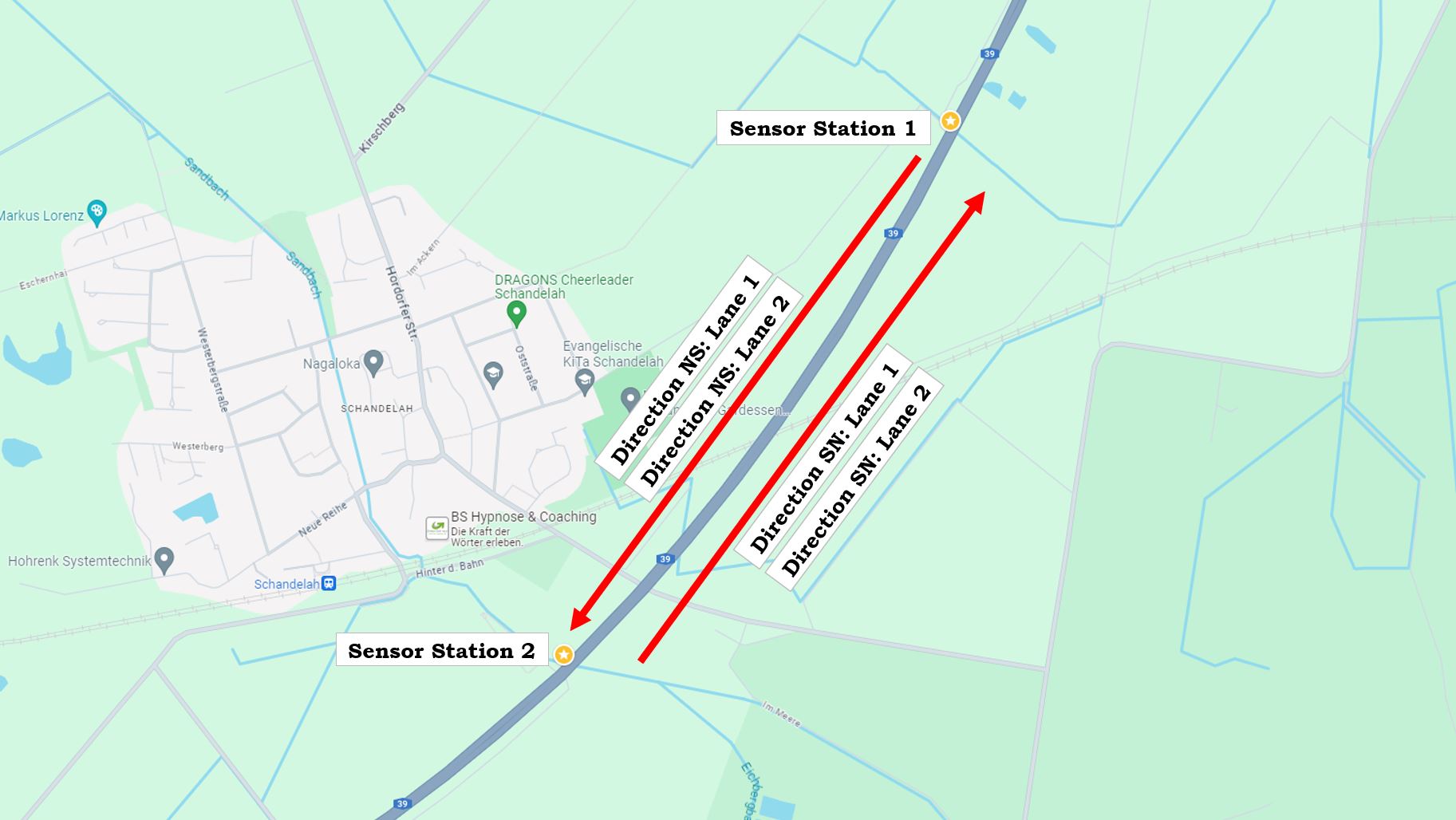}
    \caption{Visualization of the highway lanes of the real-world highway dataset 2 recorded at the German highway A39. We built two sensor stations on the west side of the highway. The car with integrated jammers drove from north to south and from south to north, each with two lanes (Map of Schandelah near Braunschweig, Germany. \textit{Google Maps}, August 2024, \url{https://maps.app.goo.gl/f8cxW3w2qSJV4FyR9}).}
    \label{figure_braunschweig_maps}
\end{figure}

Next, we introduce a second real-world GNSS highway dataset, recorded using the same platform as described in Section~\ref{label_data_rw_1}. Figure~\ref{figure_braunschweig_maps} illustrates the placement of two sensor stations adjacent to the highway at the \textit{Test Bed Lower Saxony for automated and connected mobility}~\citep{koester_mazzega,thonhofer_sigl} near Braunschweig, Germany: \newline
\url{https://verkehrsforschung.dlr.de/de/projekte/testfeld-niedersachsen-fuer-automatisierte-und-vernetzte-mobilitaet} \newline
We conducted 21 test drives in both directions and on both lanes. The blue jammer from Figure~\ref{figure_jamming_devices} was utilized at various locations within the vehicle, in addition to a jammer placed in the cigarette lighter, resulting in the snapshots shown in Figure~\ref{figure_highway2}. The \textit{chirp} interference is clearly visible in Figure~\ref{figure_highway2_5} to Figure~\ref{figure_highway2_7}, whereas the interferences from the cigarette lighter are less apparent (see Figure~\ref{figure_highway2_8} and Figure~\ref{figure_highway2_9}). The interference intensity decreases with increased distance between the vehicle and the sensor stations (see Figure~\ref{figure_highway2_5}), while it increases when the vehicle passes directly next to the stations (see Figure~\ref{figure_highway2_7}). Consequently, both the distance and driving direction of the vehicle can be predicted. We evaluated different driving velocities of $100\frac{km}{h}$, $130\frac{km}{h}$, $140\frac{km}{h}$, and $180\frac{km}{h}$. While snapshots can distinguish a slow velocity of $100\frac{km}{h}$ from a high velocity of $180\frac{km}{h}$, predicting velocity with high accuracy remains challenging. The dataset contains a total of 16,480 samples.

\subsection{Controlled Large-scale Dataset 1: Low-Frequency Antenna}
\label{label_data_contr_lf1}

\begin{figure}[!t]
	\begin{minipage}[t]{0.09\linewidth}
        \centering
    	\includegraphics[trim=170 55 90 45, clip, width=1.0\linewidth]{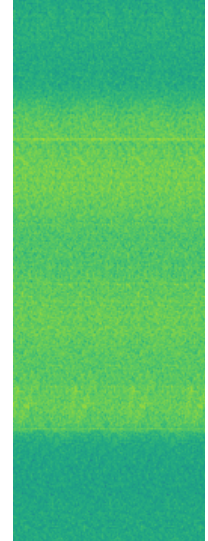}
    	\subcaption{None}
    	\label{figure_controlled_crpa1}
    \end{minipage}
    \hfill
	\begin{minipage}[t]{0.09\linewidth}
        \centering
    	\includegraphics[trim=170 55 90 45, clip, width=1.0\linewidth]{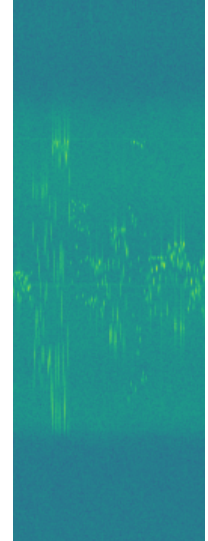}
    	\subcaption{Chirp, scenario 1}
    	\label{figure_controlled_crpa2}
    \end{minipage}
    \hfill
	\begin{minipage}[t]{0.09\linewidth}
        \centering
    	\includegraphics[trim=170 55 90 45, clip, width=1.0\linewidth]{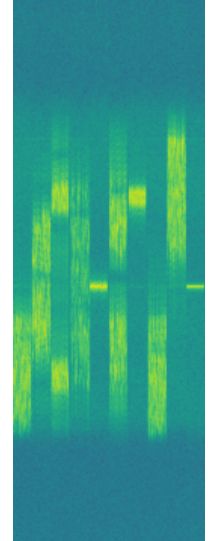}
    	\subcaption{FreqHopper}
    	\label{figure_controlled_crpa3}
    \end{minipage}
    \hfill
	\begin{minipage}[t]{0.09\linewidth}
        \centering
    	\includegraphics[trim=170 55 90 45, clip, width=1.0\linewidth]{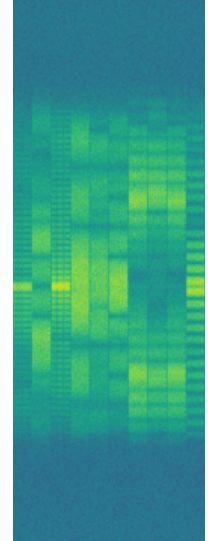}
    	\subcaption{Modulated}
    	\label{figure_controlled_crpa4}
    \end{minipage}
    \hfill
	\begin{minipage}[t]{0.09\linewidth}
        \centering
    	\includegraphics[trim=170 55 90 45, clip, width=1.0\linewidth]{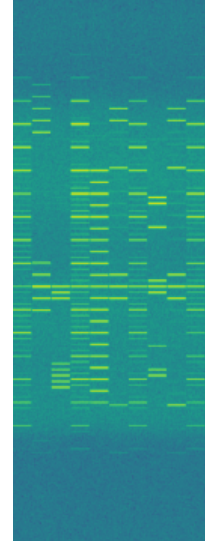}
    	\subcaption{Multitone}
    	\label{figure_controlled_crpa5}
    \end{minipage}
    \hfill
	\begin{minipage}[t]{0.09\linewidth}
        \centering
    	\includegraphics[trim=170 55 90 45, clip, width=1.0\linewidth]{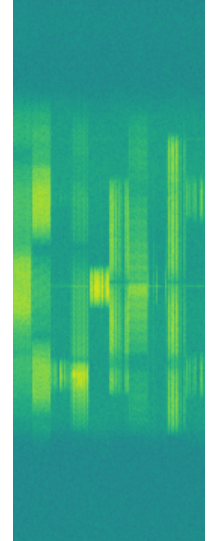}
    	\subcaption{Pulsed}
    	\label{figure_controlled_crpa6}
    \end{minipage}
    \hfill
	\begin{minipage}[t]{0.09\linewidth}
        \centering
    	\includegraphics[trim=170 55 90 45, clip, width=1.0\linewidth]{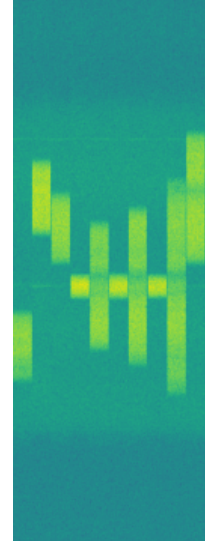}
    	\subcaption{Noise}
    	\label{figure_controlled_crpa7}
    \end{minipage}
    \hfill
	\begin{minipage}[t]{0.09\linewidth}
        \centering
    	\includegraphics[trim=170 55 90 45, clip, width=1.0\linewidth]{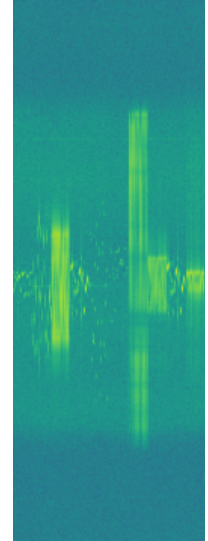}
    	\subcaption{Chirp, scenario 3}
    	\label{figure_controlled_crpa8}
    \end{minipage}
    \hfill
	\begin{minipage}[t]{0.09\linewidth}
        \centering
    	\includegraphics[trim=170 55 90 45, clip, width=1.0\linewidth]{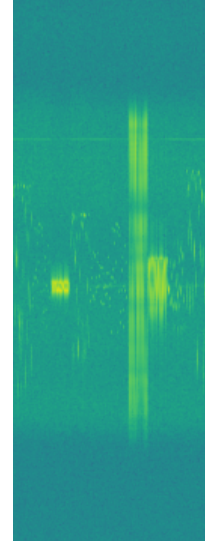}
    	\subcaption{Chirp, scenario 6}
    	\label{figure_controlled_crpa9}
    \end{minipage}
    \hfill
	\begin{minipage}[t]{0.09\linewidth}
        \centering
    	\includegraphics[trim=170 55 90 45, clip, width=1.0\linewidth]{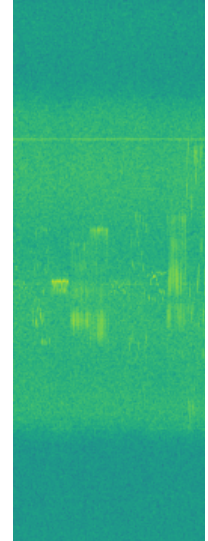}
    	\subcaption{Chirp, scenario 8}
    	\label{figure_controlled_crpa10}
    \end{minipage}
    \caption{Exemplary spectrogram samples recorded in a controlled large-scale environment with a low-frequency antenna. The x-axis shows the time in $\text{ms}$. The y-axis shows the frequency in $\text{MHz}$.}
    \label{figure_controlled_crpa}
\end{figure}

Furthermore, we recorded a GNSS dataset in the Fraunhofer IIS L.I.N.K.~test center, which mimics a controlled indoor environment. The primary aim is to develop ML models resilient to various jammer types, interference profiles, antenna variations, environmental fluctuations, shifts in location, and disparate receiver stations. The data recording setup is structured as follows: A spacious indoor hall measuring $1,320\,\textit{m}^2$ is designated as the recording environment, enabling controlled data collection that encompass multipath effects. A receiver antenna is situated on one end of the hall, and an MXG vector-signal generator is on the opposite end. The signal generator is designed to produce high-quality radio frequency (RF) signals across a wide range of frequencies. Subsequently, the antenna's signals are recorded as snapshots that include various interferences from the signal generator. A substantial dataset is recorded under diverse setups, including scenarios within an unoccupied environment and configurations featuring absorber walls interposed between the antenna and the generator. This setup facilitates recording distinct multipath effects, spanning from minor to pronounced effects, and instances of significant absorption. The antenna captures snapshots at a frequency of $100\,\text{MHz}$ with a duration of $10\,\mu s$. Each snapshot comprises 1,024 timesteps and has dimensions of $1,024 \times \text{N}_{\text{t}}$, where $\text{N}_{\text{t}}$ is the snapshot length, set to a maximum of 34. Figure~\ref{figure_controlled_crpa} illustrates a variety of snapshots, with each image displaying 10 randomly selected samples. Figure~\ref{figure_controlled_crpa1} depicts snapshots without interference. To introduce interference, six distinct types are generated (refer to Figure~\ref{figure_controlled_crpa2} to Figure~\ref{figure_controlled_crpa7}). The primary goal is to detect and accurately classify these interferences. Additionally, the jammer types necessitate characterization, including the bandwidth (BW) and signal-to-noise (StN) ratio. Thus, we record various BWs, ranging from 0.1 to 60, and StN ratios, ranging from -10 to 10. Enhanced multipath effects correspond to decreased interference intensity (refer to Figure~\ref{figure_controlled_crpa8} for scenario 3, Figure~\ref{figure_controlled_crpa9} for scenario 6, and Figure~\ref{figure_controlled_crpa10} for scenario 8). In total, the dataset comprises 41,968 samples, of which 576 samples are without any interferences.

\subsection{Controlled Large-scale Dataset 2: Low-Frequency Antenna}
\label{label_data_contr_lf2}

We conducted a second measurement campaign using the same low-frequency antenna and signal generator setup described in Section~\ref{label_data_contr_lf1} at the Fraunhofer IIS L.I.N.K.~test center. This campaign focused on recording more continuous BW and StN ratios without multipath effects. A continuous recording over 14 days yielded a total of 96,444 samples. For experimental purposes, we either combine the controlled large-scale datasets 1 and 2 or consider them separately.

\begin{figure}[!t]
	\begin{minipage}[t]{0.13\linewidth}
        \centering
    	\includegraphics[trim=160 30 160 16, clip, width=1.0\linewidth]{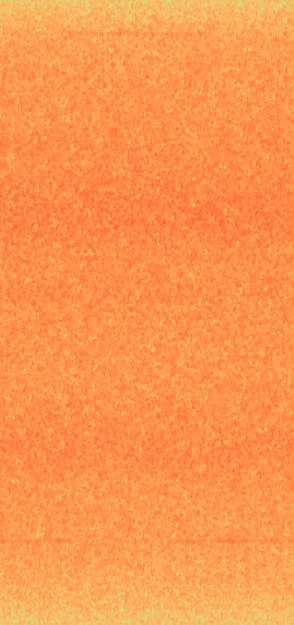}
    	\subcaption{None}
    	\label{figure_controlled_fiot1}
    \end{minipage}
    \hfill
	\begin{minipage}[t]{0.13\linewidth}
        \centering
    	\includegraphics[trim=160 30 160 16, clip, width=1.0\linewidth]{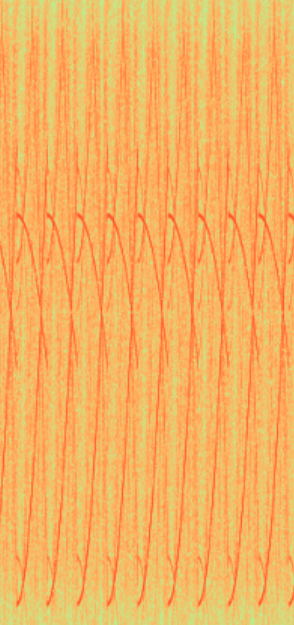}
    	\subcaption{Chirp}
    	\label{figure_controlled_fiot2}
    \end{minipage}
    \hfill
	\begin{minipage}[t]{0.13\linewidth}
        \centering
    	\includegraphics[trim=160 30 160 16, clip, width=1.0\linewidth]{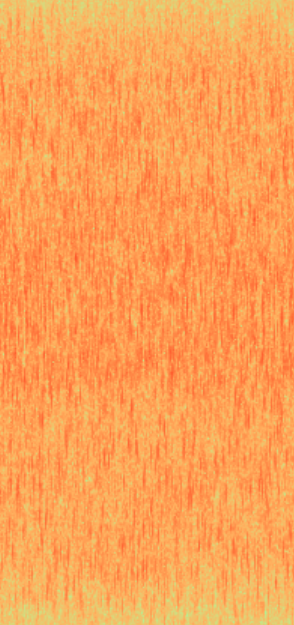}
    	\subcaption{FreqHopper}
    	\label{figure_controlled_fiot3}
    \end{minipage}
    \hfill
	\begin{minipage}[t]{0.13\linewidth}
        \centering
    	\includegraphics[trim=160 30 160 16, clip, width=1.0\linewidth]{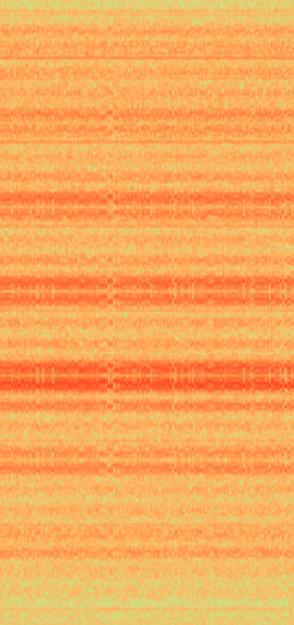}
    	\subcaption{Modulated}
    	\label{figure_controlled_fiot4}
    \end{minipage}
    \hfill
	\begin{minipage}[t]{0.13\linewidth}
        \centering
    	\includegraphics[trim=160 30 160 16, clip, width=1.0\linewidth]{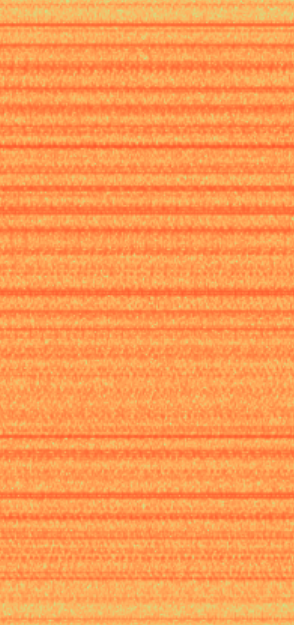}
    	\subcaption{Multitone}
    	\label{figure_controlled_fiot5}
    \end{minipage}
    \hfill
	\begin{minipage}[t]{0.13\linewidth}
        \centering
    	\includegraphics[trim=160 30 160 16, clip, width=1.0\linewidth]{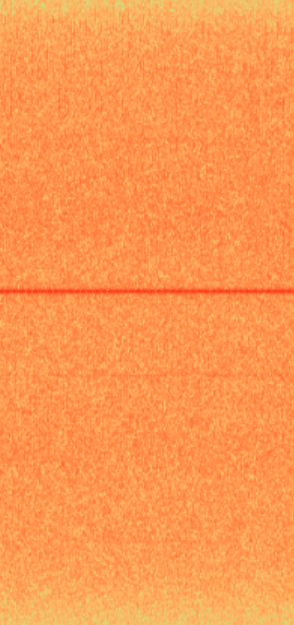}
    	\subcaption{Pulsed}
    	\label{figure_controlled_fiot6}
    \end{minipage}
    \hfill
	\begin{minipage}[t]{0.13\linewidth}
        \centering
    	\includegraphics[trim=160 30 160 16, clip, width=1.0\linewidth]{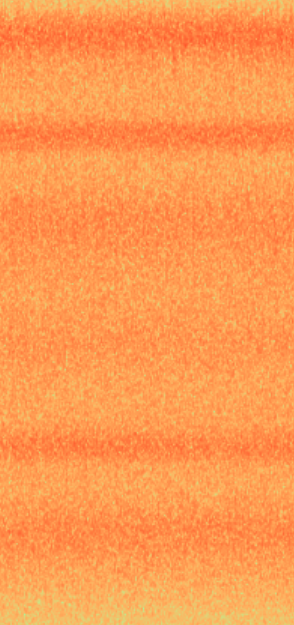}
    	\subcaption{Noise}
    	\label{figure_controlled_fiot7}
    \end{minipage}
    \caption{Exemplary spectrogram samples recorded in a controlled large-scale environment with a high-frequency antenna. The x-axis shows the time in $\text{ms}$. The y-axis shows the frequency in $\text{MHz}$.}
    \label{figure_controlled_fiot}
\end{figure}

\subsection{Controlled Large-scale Dataset 2: High-Frequency Antenna}
\label{label_data_contr_hf2}

Concurrently with the measurement campaign described in Section~\ref{label_data_contr_lf2}, we placed a high-frequency antenna next to the low-frequency antenna in the indoor environment. We recorded the same interferences, resulting in the snapshots presented in Figure~\ref{figure_controlled_fiot}. This allows for a comparison of snapshots from the two different antennas. For example, the \textit{chirp} interference produces different patterns for each antenna; compare Figure~\ref{figure_controlled_crpa2} from the low-frequency antenna with Figure~\ref{figure_controlled_fiot2} from the high-frequency antenna. This discrepancy in patterns poses a challenge for adapting ML models to different setups (i.e., different antennas). Semi-supervised learning and pseudo-labeling are potential solution that we evaluate in Section~\ref{label_eval_pseudo_labeling}. The dataset contains a total of 1,059,967 samples.

\begin{figure}[!t]
	\begin{minipage}[t]{0.14\linewidth}
        \centering
    	\includegraphics[trim=150 30 90 38, clip, width=1.0\linewidth]{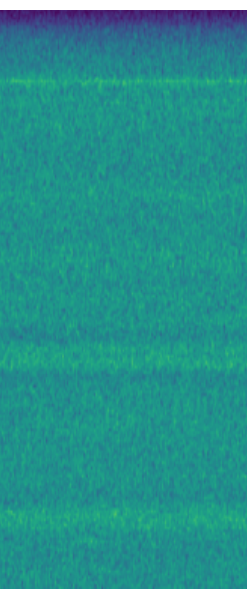}
    	\subcaption{None}
    	\label{figure_controlled_large_scale1}
    \end{minipage}
    \hfill
	\begin{minipage}[t]{0.14\linewidth}
        \centering
    	\includegraphics[trim=150 30 90 38, clip, width=1.0\linewidth]{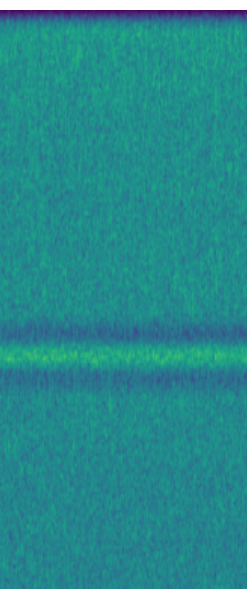}
    	\subcaption{Class 1}
    	\label{figure_controlled_large_scale2}
    \end{minipage}
    \hfill
	\begin{minipage}[t]{0.14\linewidth}
        \centering
    	\includegraphics[trim=150 30 90 38, clip, width=1.0\linewidth]{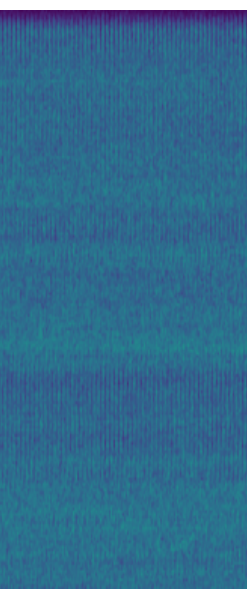}
    	\subcaption{Class 2}
    	\label{figure_controlled_large_scale3}
    \end{minipage}
    \hfill
	\begin{minipage}[t]{0.14\linewidth}
        \centering
    	\includegraphics[trim=150 30 90 38, clip, width=1.0\linewidth]{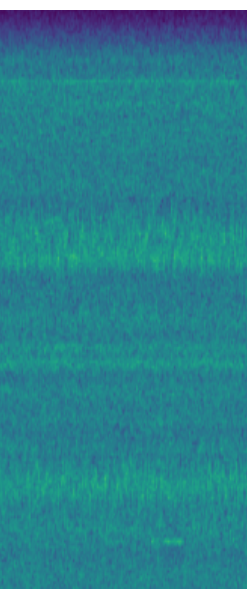}
    	\subcaption{Class 3}
    	\label{figure_controlled_large_scale4}
    \end{minipage}
    \hfill
	\begin{minipage}[t]{0.14\linewidth}
        \centering
    	\includegraphics[trim=150 30 90 38, clip, width=1.0\linewidth]{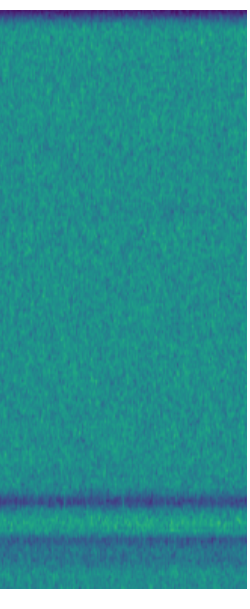}
    	\subcaption{Class 4}
    	\label{figure_controlled_large_scale5}
    \end{minipage}
    \hfill
	\begin{minipage}[t]{0.14\linewidth}
        \centering
    	\includegraphics[trim=150 30 90 38, clip, width=1.0\linewidth]{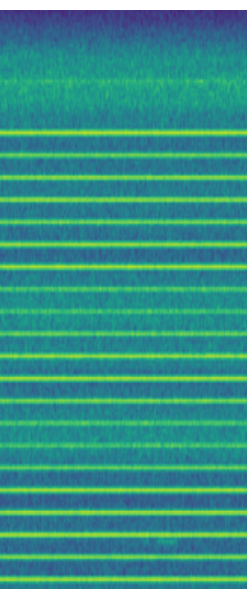}
    	\subcaption{Class 5}
    	\label{figure_controlled_large_scale6}
    \end{minipage}
    \caption{Exemplary spectrogram samples recorded in a controlled small-scale environment with a high-frequency antenna. The x-axis shows the time in $\text{ms}$. The y-axis shows the frequency in $\text{MHz}$.}
    \label{figure_controlled_large_scale}
\end{figure}

\subsection{Controlled [Large+Small]-scale Dataset: High-Frequency Antenna}
\label{label_data_contr_ls_hf}

Using an initial high-frequency prototype, we conducted recordings in the controlled large-scale environment. To facilitate a comparison of snapshots across differently scaled environments, we also recorded snapshots with interferences in a small-scale environment, specifically a small indoor lab. Exemplary snapshots are displayed in Figure~\ref{figure_controlled_large_scale}. The dataset is available in \cite{ott_heublein_dataset}.

\begin{figure}[!t]
	\begin{minipage}[t]{0.145\linewidth}
        \centering
    	\includegraphics[trim=0 30 0 0, clip, width=1.0\linewidth]{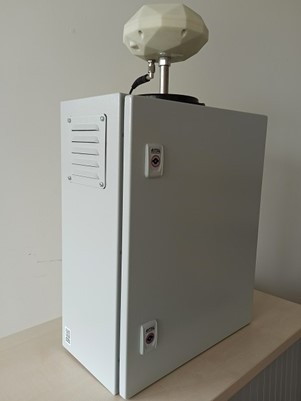}
    	\subcaption{High-frequency antenna.}
    	\label{figure_recording_setup1}
    \end{minipage}
    \hfill
	\begin{minipage}[t]{0.145\linewidth}
        \centering
    	\includegraphics[trim=0 30 0 0, clip, width=1.0\linewidth]{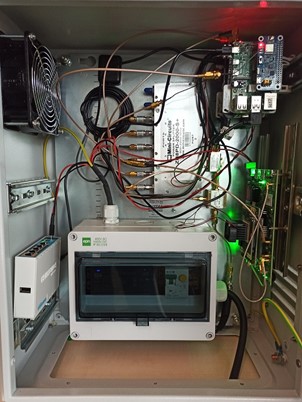}
    	\subcaption{Interior of the recording setup.}
    	\label{figure_recording_setup2}
    \end{minipage}
    \hfill
	\begin{minipage}[t]{0.23\linewidth}
        \centering
    	\includegraphics[trim=13 0 13 0, clip, width=1.0\linewidth]{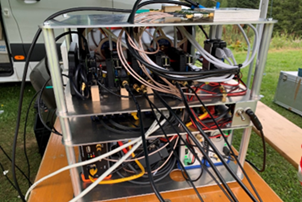}
    	\subcaption{Recording setup with low-cost sensors.}
    	\label{figure_recording_setup3}
    \end{minipage}
    \hfill
	\begin{minipage}[t]{0.23\linewidth}
        \centering
    	\includegraphics[trim=150 0 276 0, clip, width=1.0\linewidth]{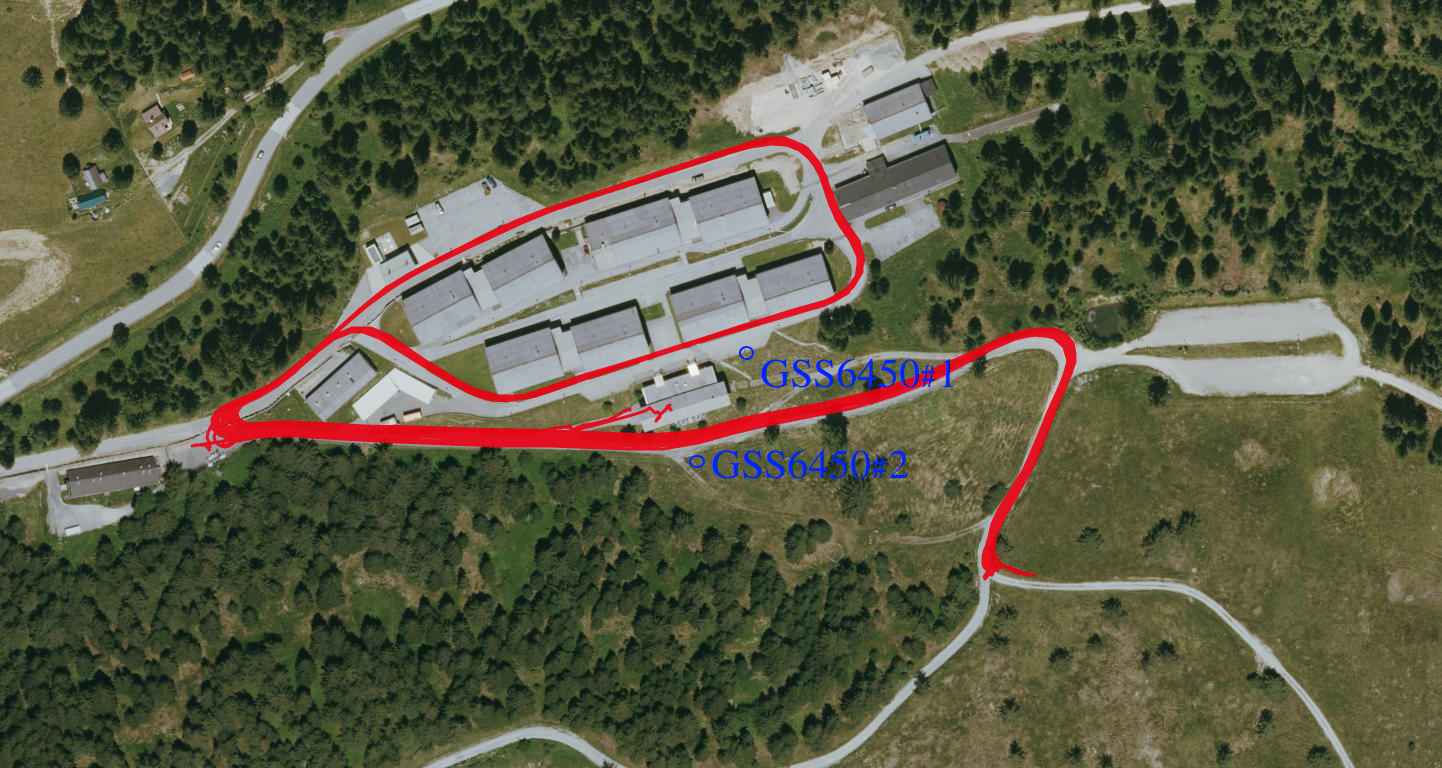}
    	\subcaption{Trajectory of the car in the Seetal Alps.}
    	\label{figure_recording_setup4}
    \end{minipage}
    \hfill
	\begin{minipage}[t]{0.23\linewidth}
        \centering
    	\includegraphics[trim=0 0 8 0, clip, width=1.0\linewidth]{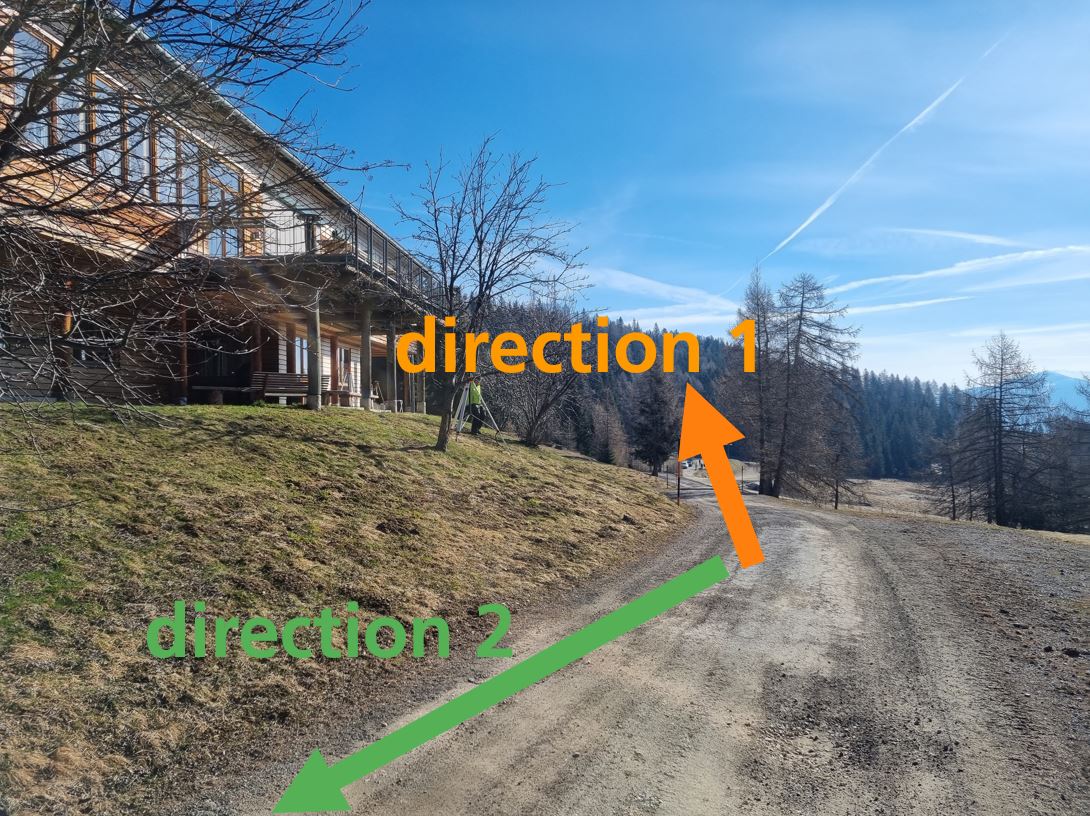}
    	\subcaption{Driving direction of the car with integrated jammers.}
    	\label{figure_recording_setup5}
    \end{minipage}
    \caption{Recording setup of our sensor station used as measurement platform at the German highways (a and b) and recording setup used as measurement platform at the Seetal Alps in Austria (c to e).}
    \label{figure_recording_setup}
\end{figure}

\subsection{Low-cost Datasets}
\label{label_data_low_cost}

\paragraph{Recording Setup.} Additionally, we collected several low-cost (LC) datasets with higher-rate samples of 1\,Hz. The central component of the hardware platform is a Raspberry Pi 4GB single-board computer (SBC), which handles primary processing, interfacing, and networking. GNSS processing is provided by a GPS Raspberry Pi hat (pHat) containing a u-blox MAX-M8Q receiver. This u-blox receiver processes GPS, Galileo, Beidou, GLONASS, QZSS, and SBAS signals in the L1 band, but can only handle a maximum of three concurrent GNSS. Therefore, it is configured to use GPS L1 C/A, Galileo E1 OS-B/C, and GLONASS G1 OS to maximize spectrum diversity. The SBC connects to the pHat via a universal asynchronous receiver-transmitter (UART) bus over the general-purpose input/output (GPIO) interface and decodes the National Marine Electronics Association (NMEA) messages at a 1\,Hz rate. The GNSS receiver is primarily managed by the GPS service daemon (GPSD). The second sensor is a NeSDR SMArt v4, containing an RTL2832U digital video broadcasting (DVB) radio-frequency front-end (RFFE). This sensor can be reconfigured as a software-defined radio (SDR) RFFE and delivers 8-bit complex IQ samples at a maximum sample rate of 3.2\,MHz. While this is sufficient to receive narrow GNSS signals, such as GPS L1 C/A (1.023\,MHz chipping rate), it is insufficient to accurately process and localize most GNSS signals~\citep{brieger_ion_gnss}. However, the bandwidth suffices for detection, monitoring, and classification purposes~\citep{merwe_franco}. The received samples are further processed on the SBC. We employ the antennas provided by both the u-blox and NeSDR receivers. Using the LC sensor’s SDR to record 64 power spectral density (PSD) and 64 kurtosis values, this data is combined with mode, number of satellites, and C/N0 per second from the u-blox sensor~\citep{brieger_ion_gnss}. The recording platform is shown in Figure~\ref{figure_recording_setup3}.

\begin{figure}[!t]
    \centering
	\begin{minipage}[t]{0.495\linewidth}
        \centering
    	\includegraphics[width=1.0\linewidth]{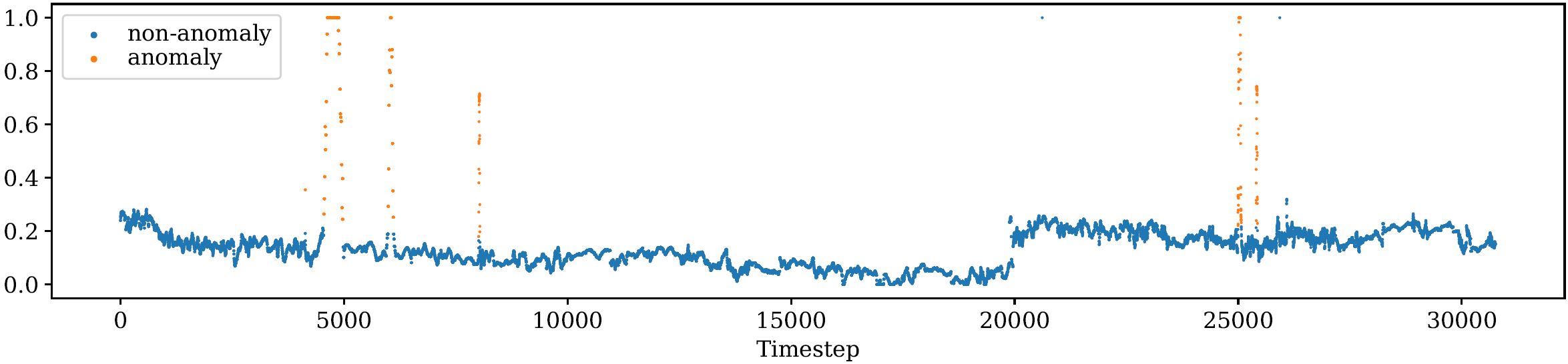}
    	\subcaption{Real-world highway dataset 1.}
    	\label{figure_lc_energy1}
    \end{minipage}
    \hfill
	\begin{minipage}[t]{0.495\linewidth}
        \centering
    	\includegraphics[width=1.0\linewidth]{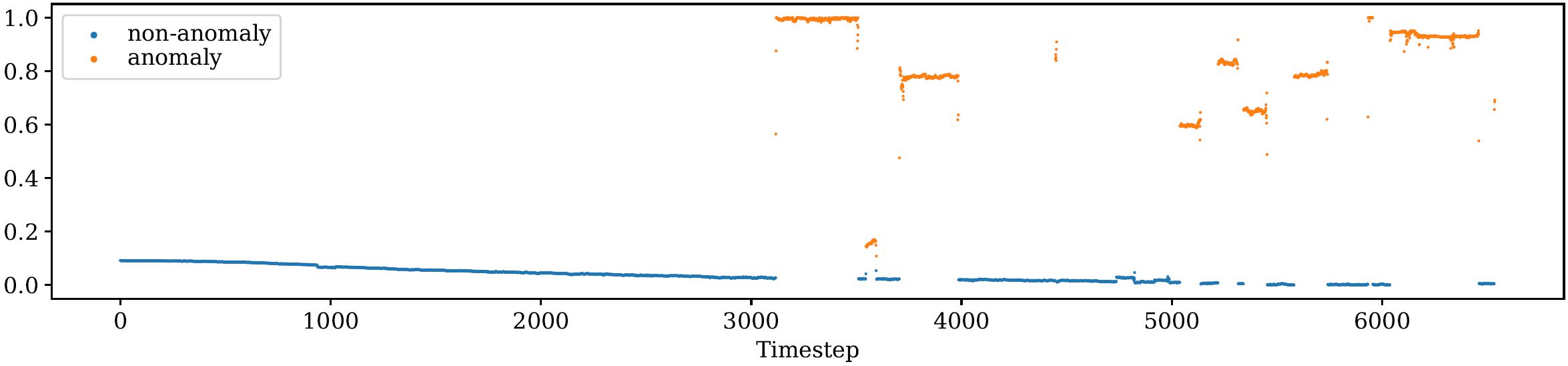}
    	\subcaption{Controlled [large+small]-scale dataset: large-scale.}
    	\label{figure_lc_energy2}
    \end{minipage}
    \hfill
	\begin{minipage}[t]{0.495\linewidth}
        \centering
    	\includegraphics[width=1.0\linewidth]{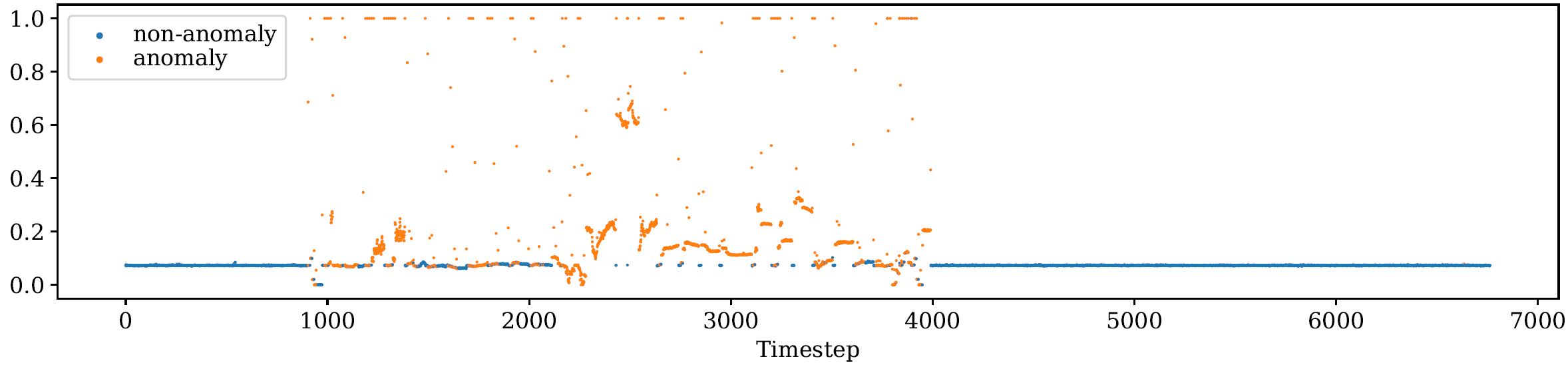}
    	\subcaption{Controlled [large+small]-scale dataset: small-scale.}
    	\label{figure_lc_energy3}
    \end{minipage}
	\begin{minipage}[t]{0.495\linewidth}
        \centering
    	\includegraphics[width=1.0\linewidth]{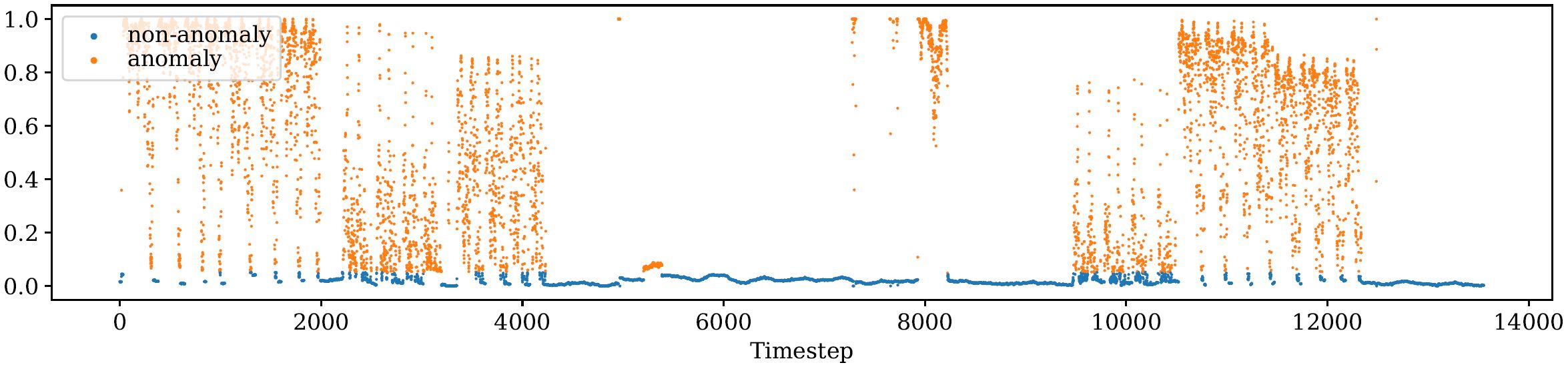}
    	\subcaption{Seetal Alps dataset: Sensor 1.}
    	\label{figure_lc_energy4}
    \end{minipage}
    \hfill
	\begin{minipage}[t]{0.495\linewidth}
        \centering
    	\includegraphics[width=1.0\linewidth]{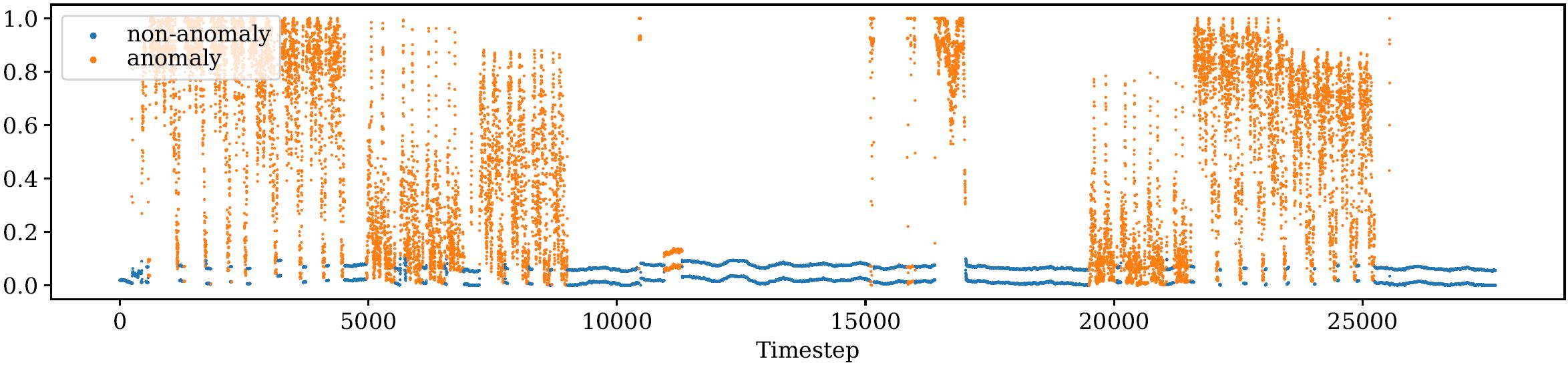}
    	\subcaption{Seetal Alps dataset: Sensor 2.}
    	\label{figure_lc_energy5}
    \end{minipage}
    \hfill
	\begin{minipage}[t]{0.495\linewidth}
        \centering
    	\includegraphics[width=1.0\linewidth]{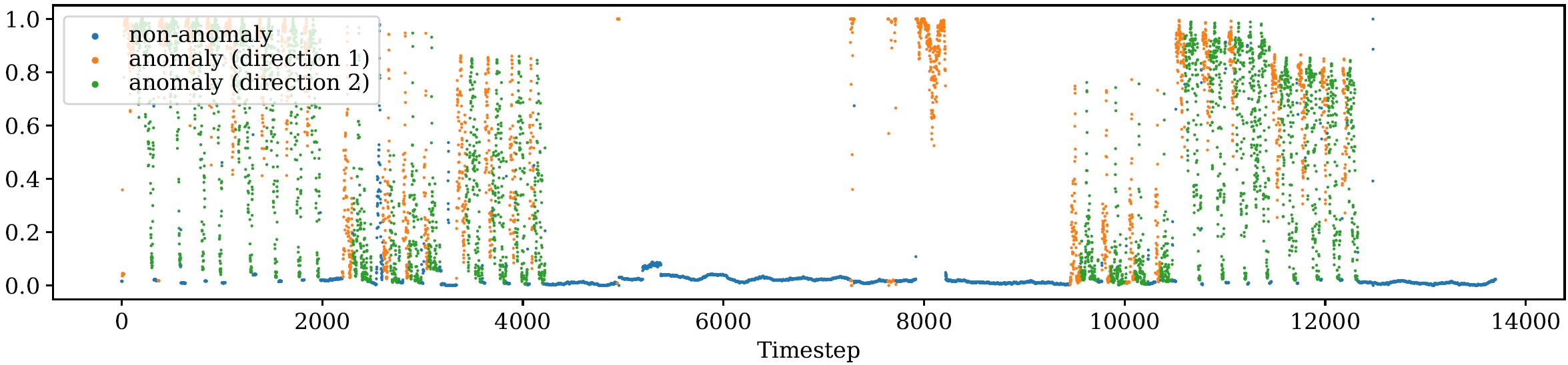}
    	\subcaption{Seetal Alps dataset: Direction.}
    	\label{figure_lc_energy6}
    \end{minipage}
    \caption{Visualization of the energy channel normalized between 0 and 1 of the low-cost sensor for four different datasets.}
    \label{figure_lc_energy}
\end{figure}

\paragraph{Measurement Campaigns.} We record four different LC GNSS datasets. The measurement setup, as integrated into the sensor station shown in Figure~\ref{figure_recording_setup2}, was used for this purpose. Simultaneously with the recording of the real-world highway dataset 1 described in Section~\ref{label_data_rw_1}, we also collected LC data. Figure~\ref{figure_lc_energy1} displays the corresponding normalized energy channel, where the orange dots indicate significant interferences. Additionally, we recorded LC data in the controlled large-scale environment (see Figure~\ref{figure_lc_energy2}) and in a small-scale environment (see Figure~\ref{figure_lc_energy3}). We also conducted recordings in the Seetal Alps in Austria, where we place the platform from Figure~\ref{figure_recording_setup3} with two LC sensors at a fixed position. A jamming device was placed in a vehicle, which was driven along the trajectory shown in Figure~\ref{figure_recording_setup4}. The LC data from both sensors are illustrated in Figure~\ref{figure_lc_energy4} and Figure~\ref{figure_lc_energy5}, respectively. Driving was conducted in both directions (as visualized in Figure~\ref{figure_recording_setup5}), and the direction was predicted as an additional task (see Figure~\ref{figure_lc_energy6}). The datasets comprise a total of 85,285 samples.

\begin{table}[t!]
\begin{center}
\setlength{\tabcolsep}{3.3pt}
    \caption{Overview of snapshot datasets for cross-validation with combined class labels and number of samples for each class.}
    \label{table_data_combination}
    \small \begin{tabular}{ p{1.3cm} | p{0.5cm} | p{0.5cm} | p{0.5cm} | p{0.5cm} | p{0.5cm} | p{0.5cm} | p{0.5cm} | p{0.5cm} | p{0.5cm} | p{0.5cm} | p{0.5cm} }
    \multicolumn{1}{c|}{\textbf{Dataset}} & \multicolumn{1}{c|}{\textbf{Antenna}} & \multicolumn{1}{c|}{\textbf{Set}} & \multicolumn{1}{c|}{\textbf{0}} & \multicolumn{1}{c|}{\textbf{1}} & \multicolumn{1}{c|}{\textbf{2}} & \multicolumn{1}{c|}{\textbf{3}} & \multicolumn{1}{c|}{\textbf{4}} & \multicolumn{1}{c|}{\textbf{5}} & \multicolumn{1}{c|}{\textbf{6}} & \multicolumn{1}{c|}{\textbf{7}} & \multicolumn{1}{c}{\textbf{8}} \\ \hline
    \multicolumn{1}{l|}{Real-world highway dataset 1} & \multicolumn{1}{l|}{High-frequency} & \multicolumn{1}{l|}{Train} & \multicolumn{1}{r|}{157,259} & \multicolumn{1}{r|}{7} & \multicolumn{1}{r|}{75} & \multicolumn{1}{r|}{} & \multicolumn{1}{r|}{} & \multicolumn{1}{r|}{} & \multicolumn{1}{r|}{30} & \multicolumn{1}{r|}{78} & \multicolumn{1}{r}{8} \\
    \multicolumn{1}{l|}{} & \multicolumn{1}{l|}{} & \multicolumn{1}{l|}{Test} & \multicolumn{1}{r|}{39,315} & \multicolumn{1}{r|}{2} & \multicolumn{1}{r|}{20} & \multicolumn{1}{r|}{} & \multicolumn{1}{r|}{} & \multicolumn{1}{r|}{} & \multicolumn{1}{r|}{11} & \multicolumn{1}{r|}{20} & \multicolumn{1}{r}{2} \\
    \multicolumn{1}{l|}{Real-world highway dataset 2} & \multicolumn{1}{l|}{High-frequency} & \multicolumn{1}{l|}{Train} & \multicolumn{1}{r|}{12,096} & \multicolumn{1}{r|}{} & \multicolumn{1}{r|}{800} & \multicolumn{1}{r|}{} & \multicolumn{1}{r|}{} & \multicolumn{1}{r|}{} & \multicolumn{1}{r|}{} & \multicolumn{1}{r|}{} & \multicolumn{1}{r}{} \\
    \multicolumn{1}{l|}{} & \multicolumn{1}{l|}{} & \multicolumn{1}{l|}{Test} & \multicolumn{1}{r|}{3,024} & \multicolumn{1}{r|}{} & \multicolumn{1}{r|}{200} & \multicolumn{1}{r|}{} & \multicolumn{1}{r|}{} & \multicolumn{1}{r|}{} & \multicolumn{1}{r|}{} & \multicolumn{1}{r|}{} & \multicolumn{1}{r}{} \\
    \multicolumn{1}{l|}{Controlled large-scale dataset 1} & \multicolumn{1}{l|}{Low-frequency} & \multicolumn{1}{l|}{Train} & \multicolumn{1}{r|}{460} & \multicolumn{1}{r|}{6,440} & \multicolumn{1}{r|}{22,728} & \multicolumn{1}{r|}{892} & \multicolumn{1}{r|}{76} & \multicolumn{1}{r|}{1,932} & \multicolumn{1}{r|}{1,016} & \multicolumn{1}{r|}{} & \multicolumn{1}{r}{} \\
    \multicolumn{1}{l|}{} & \multicolumn{1}{l|}{} & \multicolumn{1}{l|}{Test} & \multicolumn{1}{r|}{116} & \multicolumn{1}{r|}{1,620} & \multicolumn{1}{r|}{5,696} & \multicolumn{1}{r|}{224} & \multicolumn{1}{r|}{20} & \multicolumn{1}{r|}{492} & \multicolumn{1}{r|}{256} & \multicolumn{1}{r|}{} & \multicolumn{1}{r}{} \\
    \multicolumn{1}{l|}{Controlled large-scale dataset 2} & \multicolumn{1}{l|}{Low-frequency} & \multicolumn{1}{l|}{Train} & \multicolumn{1}{r|}{17,304} & \multicolumn{1}{r|}{9,072} & \multicolumn{1}{r|}{25,184} & \multicolumn{1}{r|}{10,428} & \multicolumn{1}{r|}{908} & \multicolumn{1}{r|}{2,336} & \multicolumn{1}{r|}{11,884} & \multicolumn{1}{r|}{} & \multicolumn{1}{r}{} \\
    \multicolumn{1}{l|}{} & \multicolumn{1}{l|}{} & \multicolumn{1}{l|}{Test} & \multicolumn{1}{r|}{4,332} & \multicolumn{1}{r|}{2,284} & \multicolumn{1}{r|}{6,308} & \multicolumn{1}{r|}{2,608} & \multicolumn{1}{r|}{228} & \multicolumn{1}{r|}{596} & \multicolumn{1}{r|}{2,972} & \multicolumn{1}{r|}{} & \multicolumn{1}{r}{} \\
    \multicolumn{1}{l|}{Controlled large-scale dataset 2} & \multicolumn{1}{l|}{High-frequency} & \multicolumn{1}{l|}{Train} & \multicolumn{1}{r|}{209,218} & \multicolumn{1}{r|}{106,584} & \multicolumn{1}{r|}{237,524} & \multicolumn{1}{r|}{122,862} & \multicolumn{1}{r|}{10,930} & \multicolumn{1}{r|}{21,266} & \multicolumn{1}{r|}{139,669} & \multicolumn{1}{r|}{} & \multicolumn{1}{r}{} \\
    \multicolumn{1}{l|}{} & \multicolumn{1}{l|}{} & \multicolumn{1}{l|}{Test} & \multicolumn{1}{r|}{52,646} & \multicolumn{1}{r|}{27,388} & \multicolumn{1}{r|}{58,583} & \multicolumn{1}{r|}{30,404} & \multicolumn{1}{r|}{2,388} & \multicolumn{1}{r|}{5,476} & \multicolumn{1}{r|}{35,029} & \multicolumn{1}{r|}{} & \multicolumn{1}{r}{} \\
    \multicolumn{1}{l|}{Controlled [large+small]-scale dataset} & \multicolumn{1}{l|}{High-frequency} & \multicolumn{1}{l|}{Train} & \multicolumn{1}{r|}{1,938} & \multicolumn{1}{r|}{810} & \multicolumn{1}{r|}{288} & \multicolumn{1}{r|}{306} & \multicolumn{1}{r|}{768} & \multicolumn{1}{r|}{324} & \multicolumn{1}{r|}{804} & \multicolumn{1}{r|}{30} & \multicolumn{1}{r}{} \\
    \multicolumn{1}{l|}{} & \multicolumn{1}{l|}{} & \multicolumn{1}{l|}{Test} & \multicolumn{1}{r|}{48} & \multicolumn{1}{r|}{228} & \multicolumn{1}{r|}{96} & \multicolumn{1}{r|}{102} & \multicolumn{1}{r|}{210} & \multicolumn{1}{r|}{90} & \multicolumn{1}{r|}{234} & \multicolumn{1}{r|}{12} & \multicolumn{1}{r}{} \\
    \end{tabular}
\end{center}
\end{table}

\begin{table}[t!]
\begin{center}
\setlength{\tabcolsep}{2.7pt}
    \caption{Visualization of embeddings of various low-cost sensor channels with t-distributed stochastic neighbor embedding (t-SNE) proposed by \cite{maaten_hinton}. \textit{Blue} shows samples without interference and \textit{red} shows samples with interference.}
    \label{table_tsne_plots}
    \small \begin{tabular}{ p{1.3cm} | p{0.5cm} | p{0.5cm} | p{0.5cm} | p{0.5cm} | p{0.5cm} | p{0.5cm} | p{0.5cm} }
    \multicolumn{1}{c|}{\textbf{Dataset}} & \multicolumn{1}{c}{\textbf{azimuth}} & \multicolumn{1}{c}{\textbf{elevation}} & \multicolumn{1}{c}{\textbf{kurtosis}} & \multicolumn{1}{c}{\textbf{PRN}} & \multicolumn{1}{c}{\textbf{SDR}} & \multicolumn{1}{c}{\textbf{SS}} & \multicolumn{1}{c}{\textbf{combined}} \\ \hline
    \multicolumn{1}{c|}{\rot{90}{\makecell{\textbf{Real-world} \\ \textbf{highway 1}}}} & \multicolumn{1}{c}{\centering\includegraphics[trim=57 37 45 42, clip, width=0.12\linewidth]{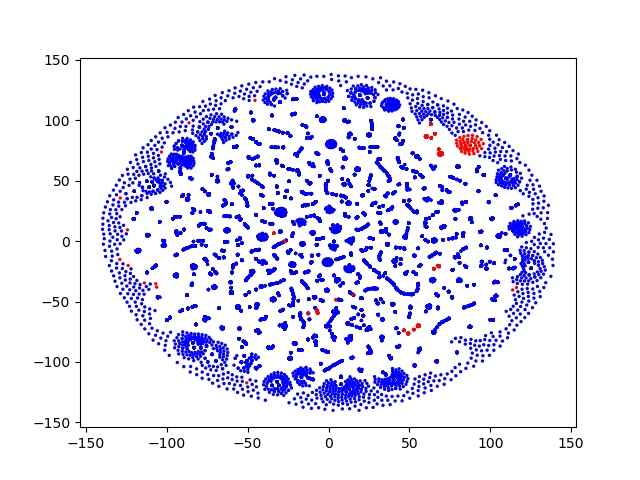}} & \multicolumn{1}{c}{\centering\includegraphics[trim=57 37 45 42, clip, width=0.12\linewidth]{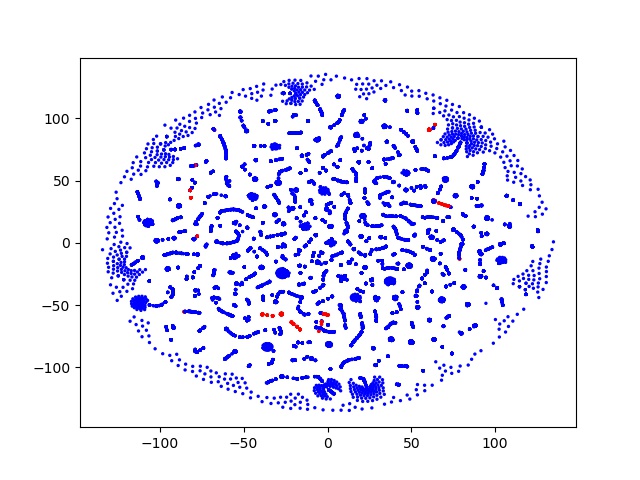}} & \multicolumn{1}{c}{\centering\includegraphics[trim=57 37 45 42, clip, width=0.12\linewidth]{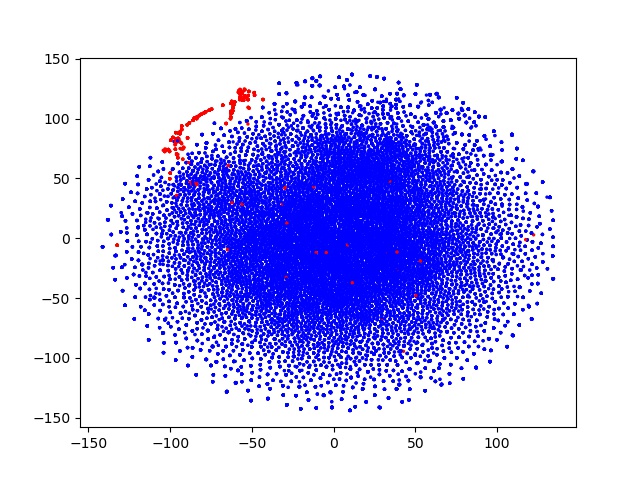}} & \multicolumn{1}{c}{\centering\includegraphics[trim=57 37 45 42, clip, width=0.12\linewidth]{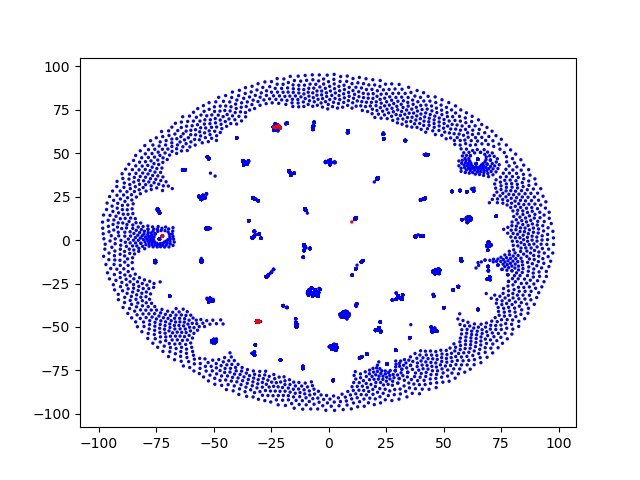}} & \multicolumn{1}{c}{\centering\includegraphics[trim=57 37 45 42, clip, width=0.12\linewidth]{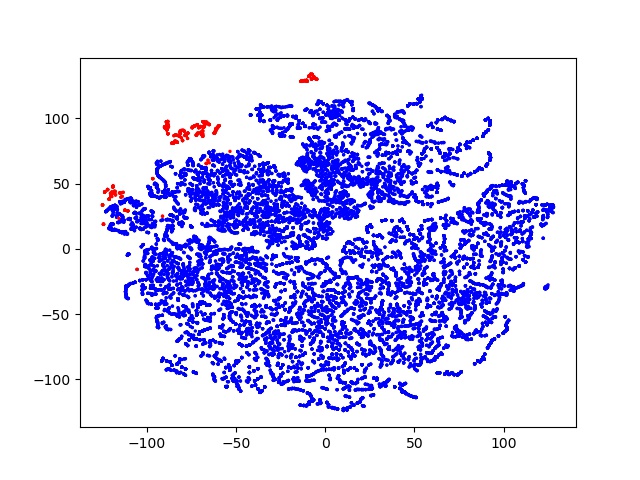}} & \multicolumn{1}{c}{\centering\includegraphics[trim=57 37 45 42, clip, width=0.12\linewidth]{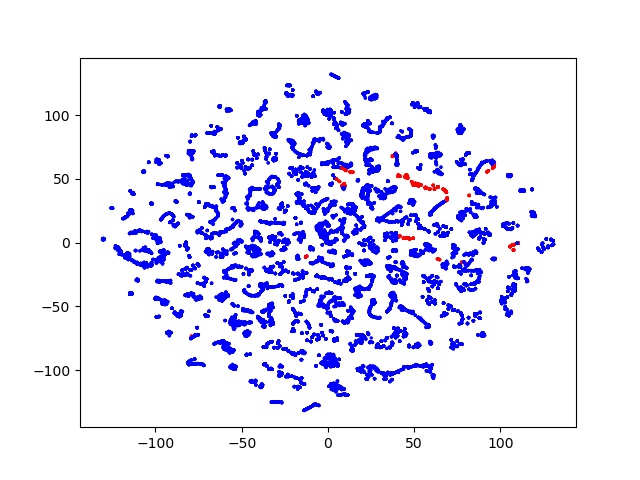}} & \multicolumn{1}{c}{\centering\includegraphics[trim=57 37 45 42, clip, width=0.12\linewidth]{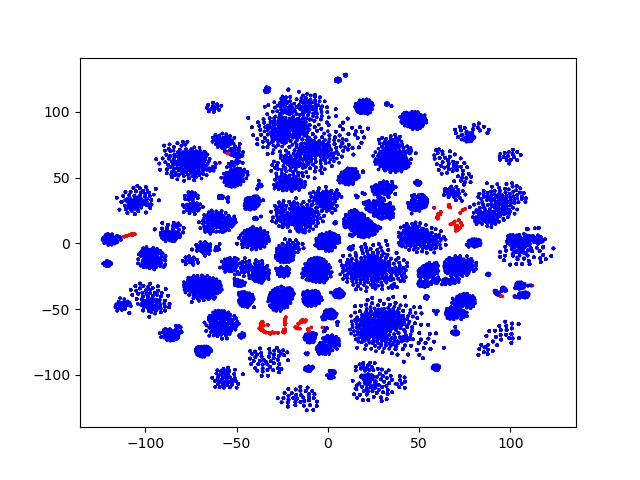}} \\
    \multicolumn{1}{c|}{\rot{90}{\makecell{\textbf{Controlled} \\ \textbf{large-scale}}}} & \multicolumn{1}{c}{\centering\includegraphics[trim=57 37 45 42, clip, width=0.12\linewidth]{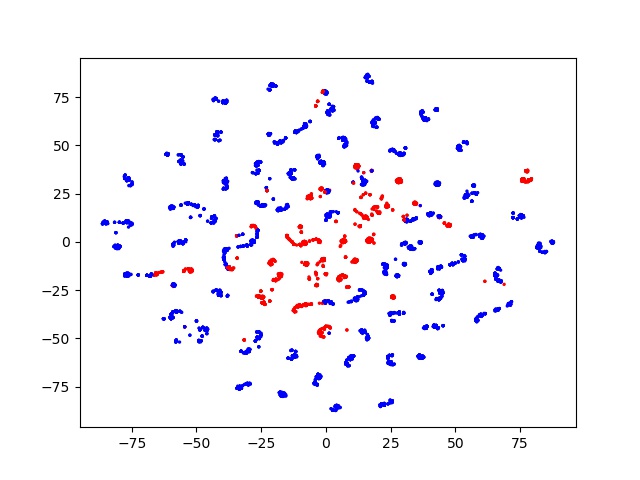}} & \multicolumn{1}{c}{\centering\includegraphics[trim=57 37 45 42, clip, width=0.12\linewidth]{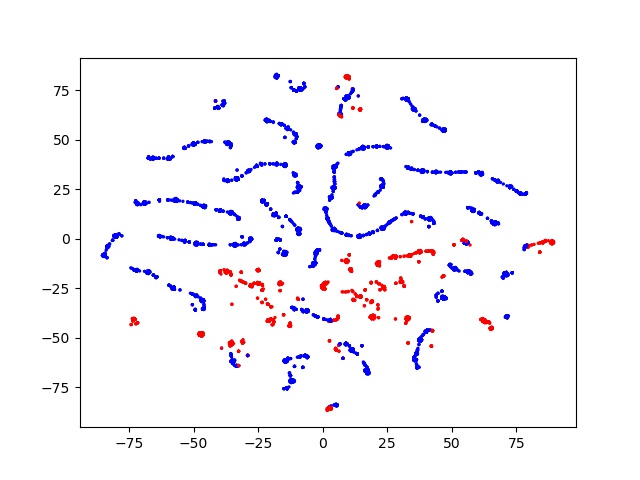}} & \multicolumn{1}{c}{\centering\includegraphics[trim=57 37 45 42, clip, width=0.12\linewidth]{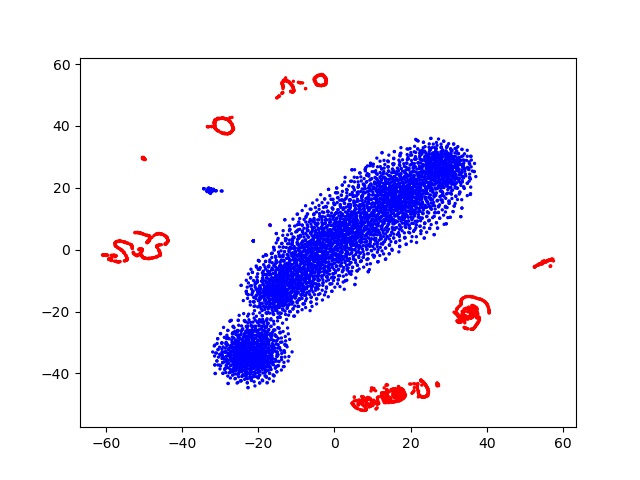}} & \multicolumn{1}{c}{\centering\includegraphics[trim=57 37 45 42, clip, width=0.12\linewidth]{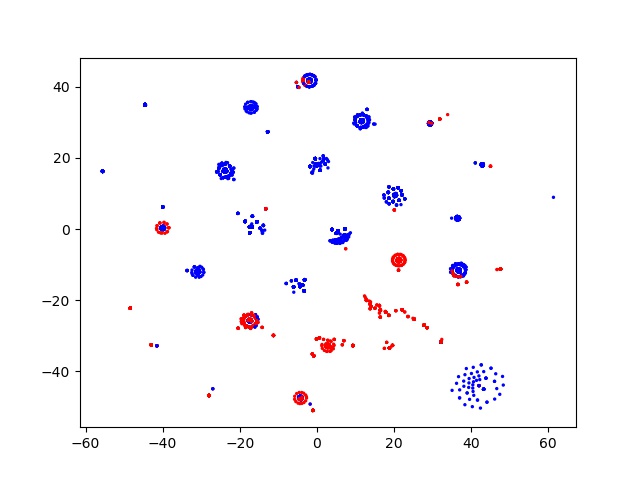}} & \multicolumn{1}{c}{\centering\includegraphics[trim=57 37 45 42, clip, width=0.12\linewidth]{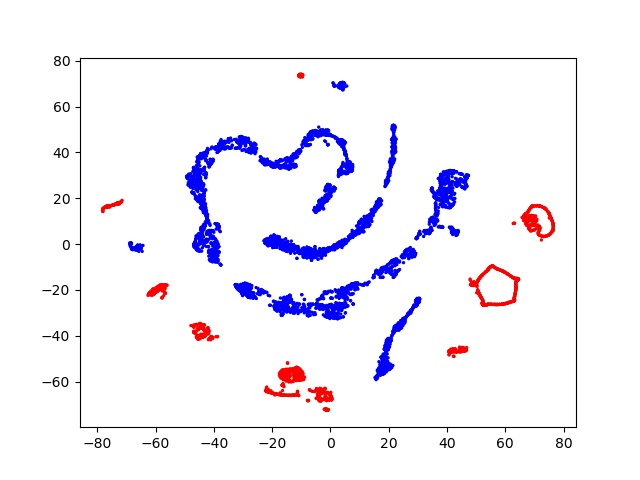}} & \multicolumn{1}{c}{\centering\includegraphics[trim=57 37 45 42, clip, width=0.12\linewidth]{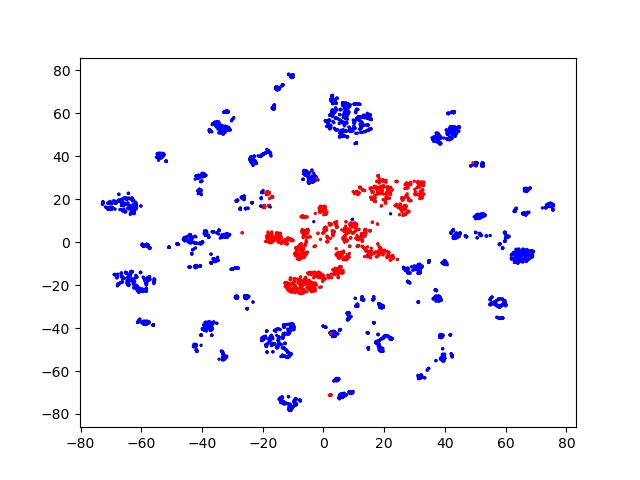}} & \multicolumn{1}{c}{\centering\includegraphics[trim=57 37 45 42, clip, width=0.12\linewidth]{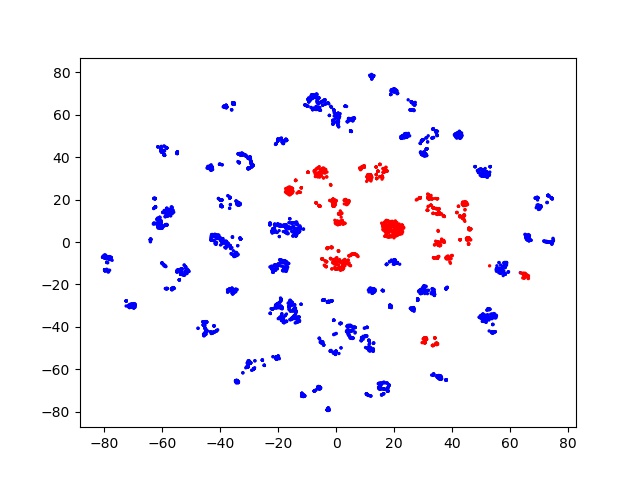}} \\
    \multicolumn{1}{c|}{\rot{90}{\makecell{\textbf{Controlled} \\ \textbf{small-scale}}}} & \multicolumn{1}{c}{\centering\includegraphics[trim=57 37 45 42, clip, width=0.12\linewidth]{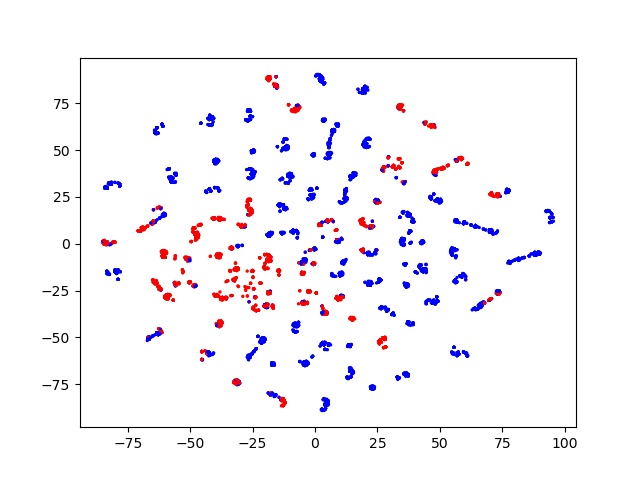}} & \multicolumn{1}{c}{\centering\includegraphics[trim=57 37 45 42, clip, width=0.12\linewidth]{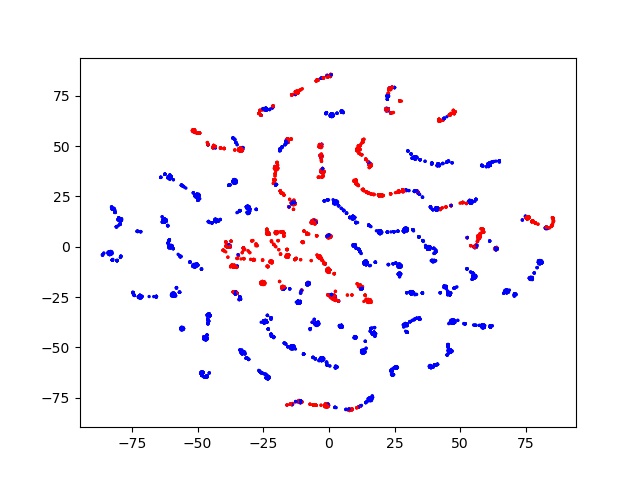}} & \multicolumn{1}{c}{\centering\includegraphics[trim=57 37 45 42, clip, width=0.12\linewidth]{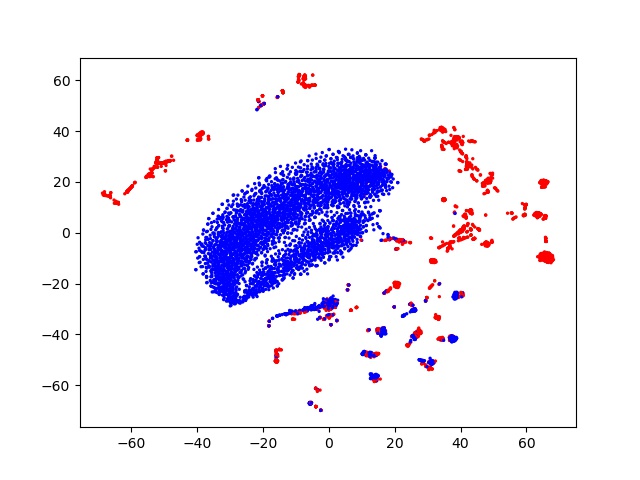}} & \multicolumn{1}{c}{\centering\includegraphics[trim=57 37 45 42, clip, width=0.12\linewidth]{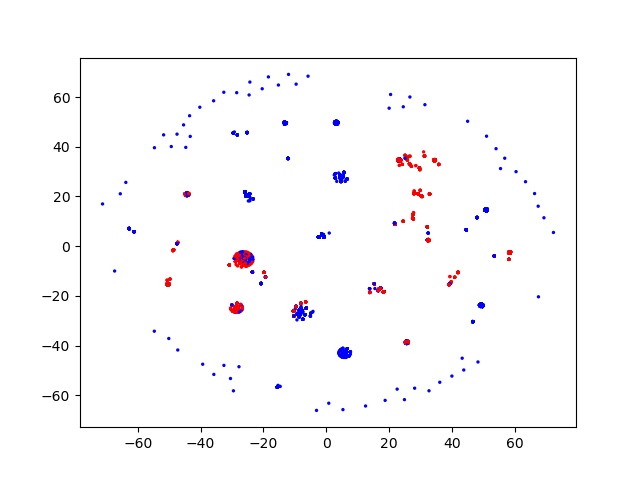}} & \multicolumn{1}{c}{\centering\includegraphics[trim=57 37 45 42, clip, width=0.12\linewidth]{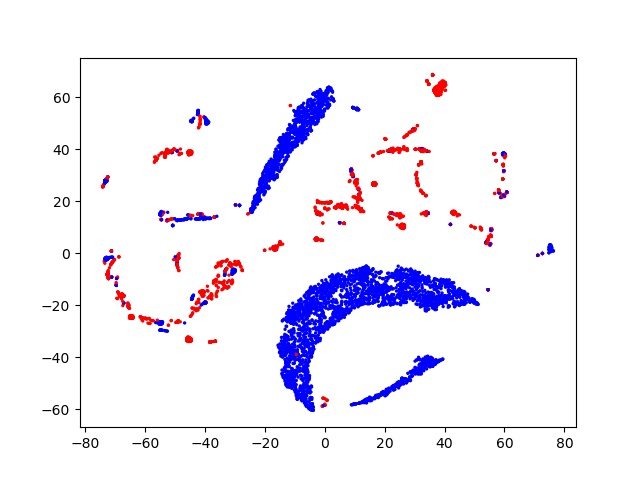}} & \multicolumn{1}{c}{\centering\includegraphics[trim=57 37 45 42, clip, width=0.12\linewidth]{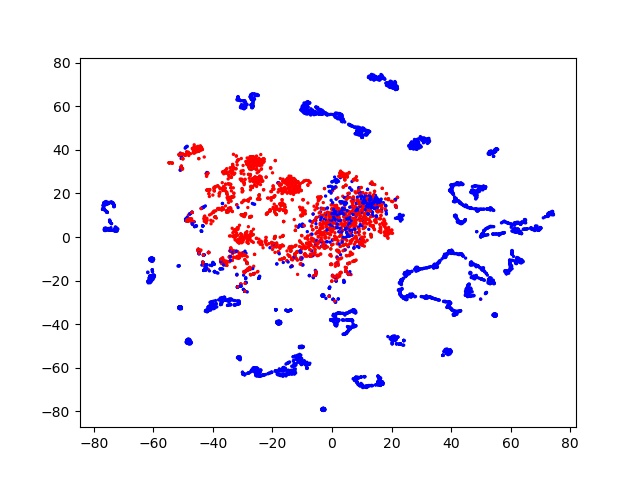}} & \multicolumn{1}{c}{\centering\includegraphics[trim=57 37 45 42, clip, width=0.12\linewidth]{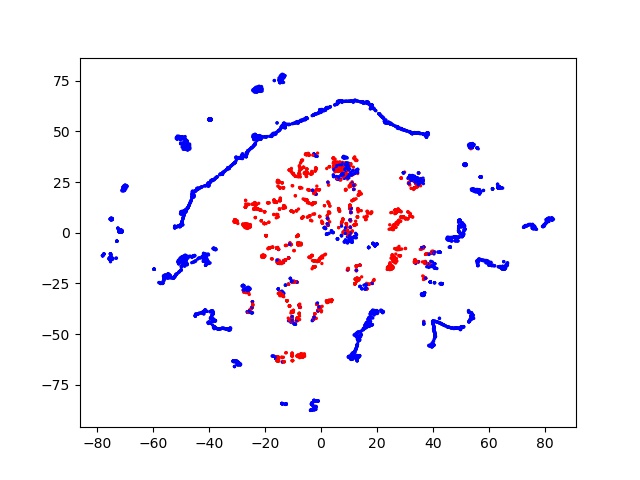}} \\
    \multicolumn{1}{c|}{\rot{90}{\makecell{\textbf{Seetal Alps} \\ \textbf{sensor 1}}}} & \multicolumn{1}{c}{\centering\includegraphics[trim=57 37 45 42, clip, width=0.12\linewidth]{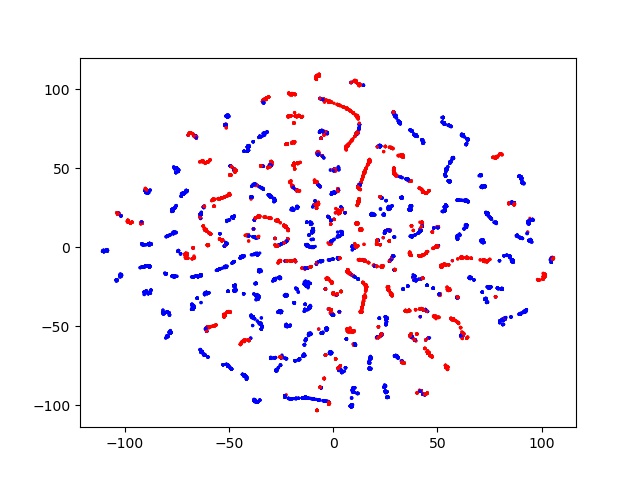}} & \multicolumn{1}{c}{\centering\includegraphics[trim=57 37 45 42, clip, width=0.12\linewidth]{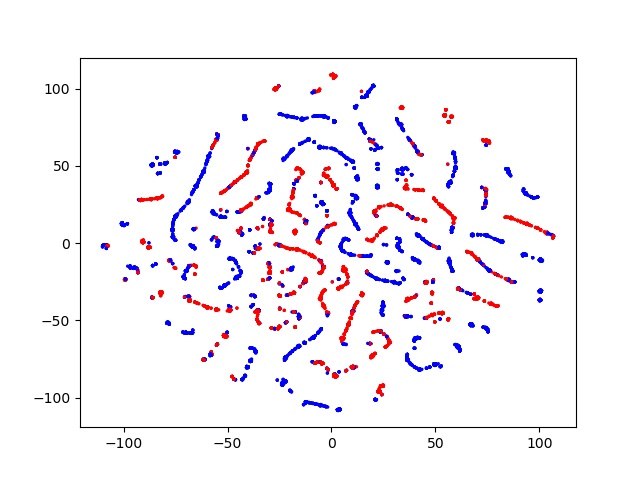}} & \multicolumn{1}{c}{\centering\includegraphics[trim=57 37 45 42, clip, width=0.12\linewidth]{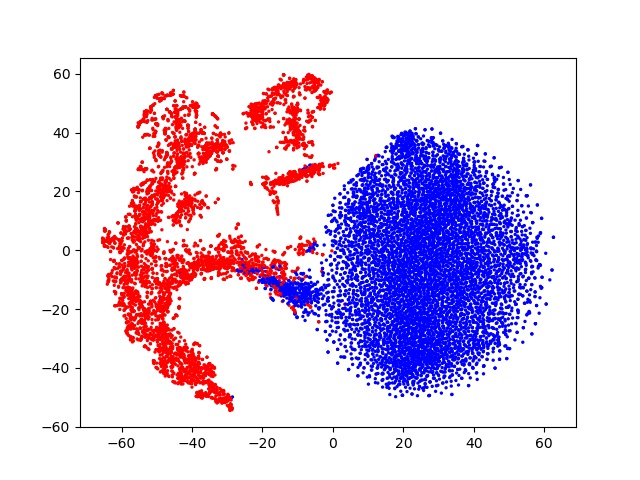}} & \multicolumn{1}{c}{\centering\includegraphics[trim=57 37 45 42, clip, width=0.12\linewidth]{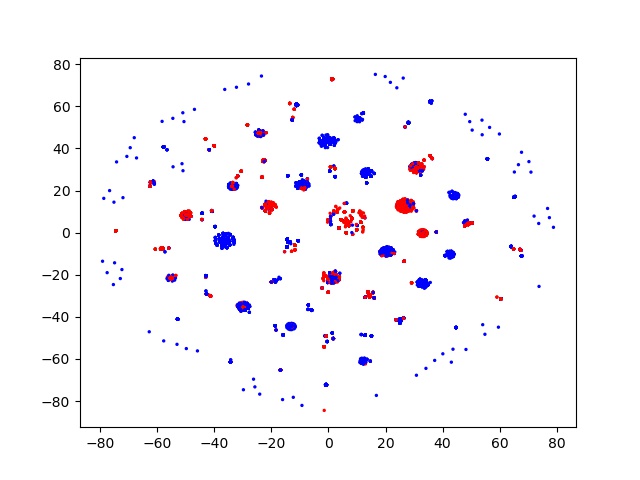}} & \multicolumn{1}{c}{\centering\includegraphics[trim=57 37 45 42, clip, width=0.12\linewidth]{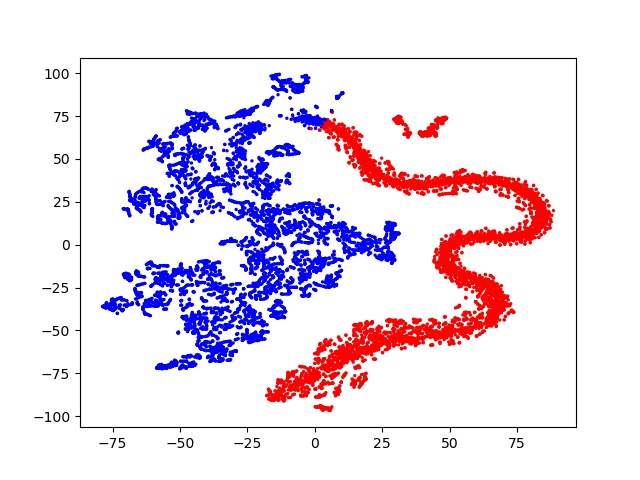}} & \multicolumn{1}{c}{\centering\includegraphics[trim=57 37 45 42, clip, width=0.12\linewidth]{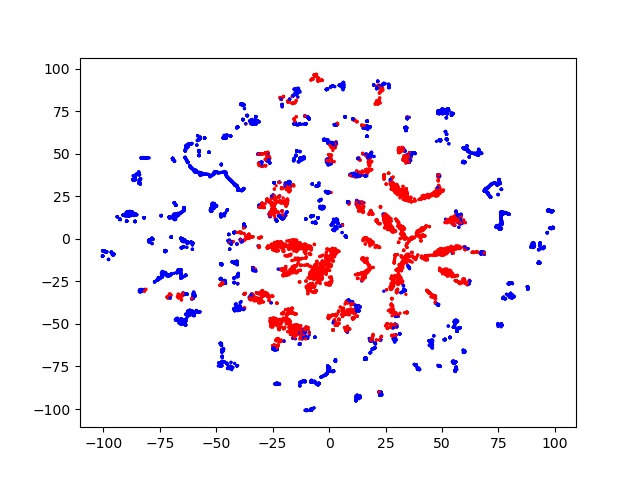}} & \multicolumn{1}{c}{\centering\includegraphics[trim=57 37 45 42, clip, width=0.12\linewidth]{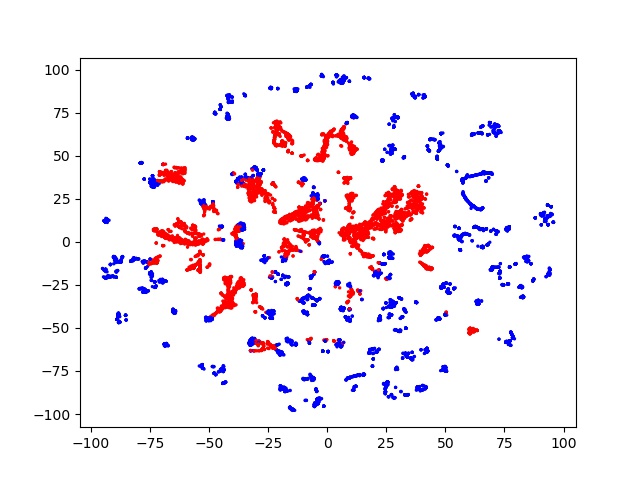}} \\
    \multicolumn{1}{c|}{\rot{90}{\makecell{\textbf{Seetal Alps} \\ \textbf{sensor 2}}}} & \multicolumn{1}{c}{\centering\includegraphics[trim=57 37 45 42, clip, width=0.12\linewidth]{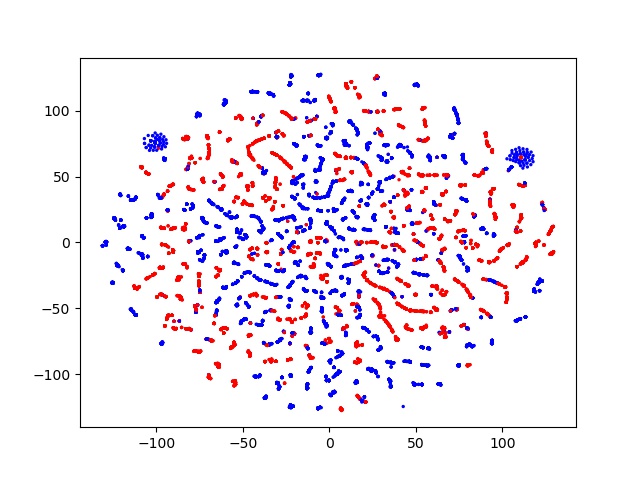}} & \multicolumn{1}{c}{\centering\includegraphics[trim=57 37 45 42, clip, width=0.12\linewidth]{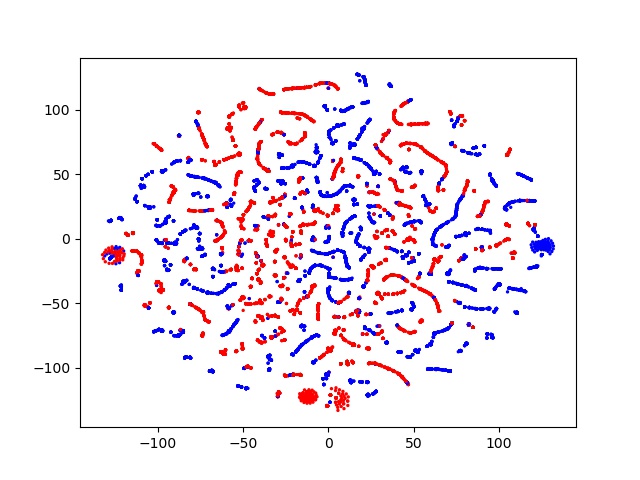}} & \multicolumn{1}{c}{\centering\includegraphics[trim=57 37 45 42, clip, width=0.12\linewidth]{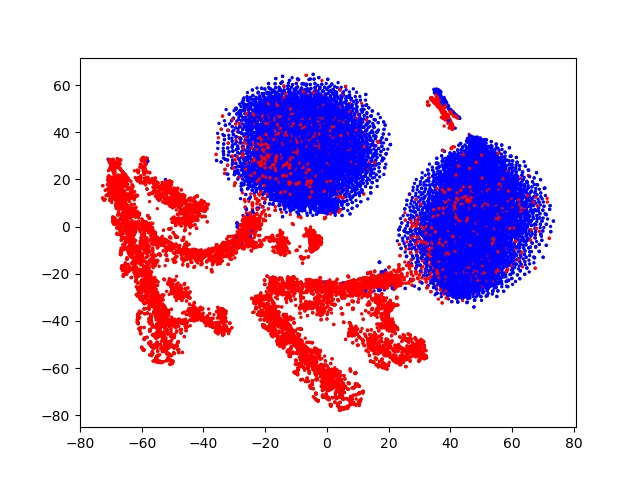}} & \multicolumn{1}{c}{\centering\includegraphics[trim=57 37 45 42, clip, width=0.12\linewidth]{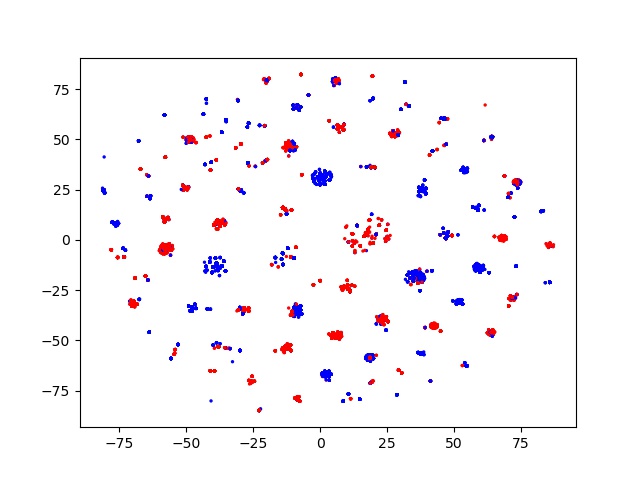}} & \multicolumn{1}{c}{\centering\includegraphics[trim=57 37 45 42, clip, width=0.12\linewidth]{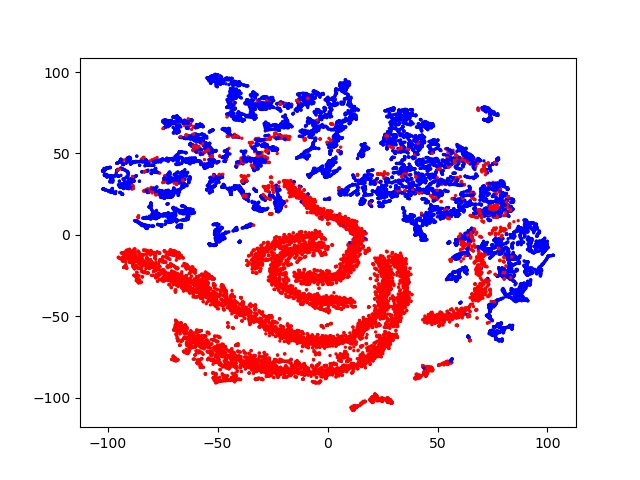}} & \multicolumn{1}{c}{\centering\includegraphics[trim=57 37 45 42, clip, width=0.12\linewidth]{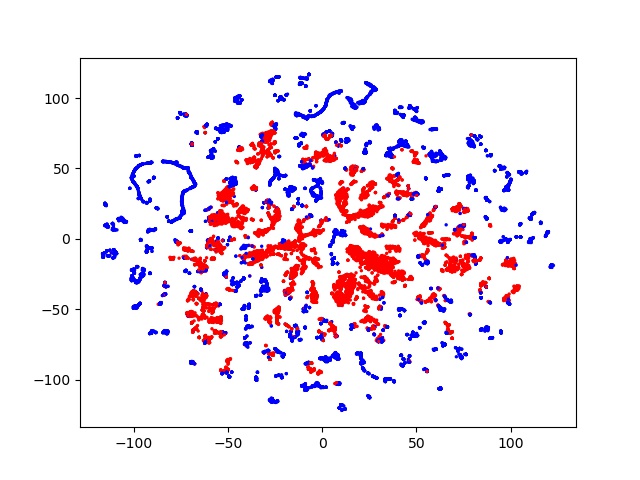}} & \multicolumn{1}{c}{\centering\includegraphics[trim=57 37 45 42, clip, width=0.12\linewidth]{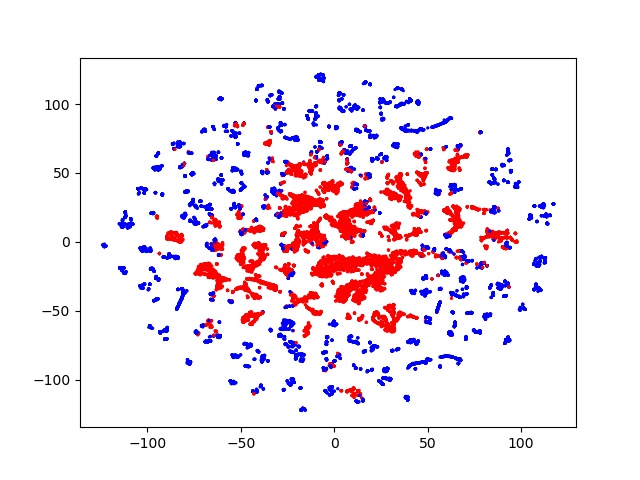}} \\
    \end{tabular}
\end{center}
\end{table}

\subsection{Dataset Combination \& Data Discrepancy}
\label{label_data_discrepancy}

Utilizing datasets from different domains in ML, particularly in the area of GNSS interference classification, enhances model performance and generalizability. Cross-domain data transfer allows for the incorporation of diverse features and patterns, which can help in training more robust and versatile models. In the context of GNSS interference classification, employing datasets from various domains can aid in identifying interference types and patterns that might not be present in GNSS-specific data alone. This approach can lead to the development of algorithms that are better equipped to detect and classify interference under varying conditions and in different environments.

Consequently, our objective is to integrate multiple datasets to enhance model generalizability. However, this generalization is compromised by discrepancies between datasets, such as those arising from the transition from controlled to real-world environments. To address this, we have created a snapshot dataset derived from the datasets discussed in Section~\ref{label_data_rw_1} to Section~\ref{label_data_contr_ls_hf}, ensuring overlapping labels. In this dataset, label 0 represents non-interference classes, while labels 1 to 8 denote interference classes (refer to Table~\ref{table_data_combination}). We perform cross-validation on all snapshot datasets and assess combined training to present the limitations of the ML model (see Section~\ref{label_eval_cross_validation}).

This data shift is also evident in the LC datasets. Table~\ref{table_tsne_plots} illustrates the lower-dimensional feature representations obtained through t-distributed stochastic neighbor embedding (t-SNE), as proposed by \cite{maaten_hinton}, for all sensor channels. It is apparent that there is a distinct separation between the non-interference classes (\textit{blue}) and the interference classes (\textit{red}). For all dataset, the SDR values between the classes are clearly separable. A noticeable shift in feature embeddings is observed between the real-world highway dataset 1 and the other datasets. This domain shift results in a decrease in the accuracy of ML models, necessitating the use of appropriate domain adaptation methods.
\section{Experiments}
\label{label_experiments}

First, we summarize all ML models that we cross-validate on our snapshot datasets (see Section~\ref{section_exp_ml_models}). Subsequently, we provide an overview of outlier detection methods for identifying GNSS interferences in LC data (see Section~\ref{section_exp_outlier_detection}). To minimize discrepancies between LC datasets, we evaluate various domain adaptation methods (see Section~\ref{section_exp_da_models}). Details regarding data augmentation techniques are provided in Section~\ref{section_exp_data_augmentation}.

\begin{table}[t!]
\begin{center}
    \caption{Overview of machine learning models. For an implementation, see the tsai~\citep{tsai} toolbox.}
    \label{table_ml_methods}
    \small \begin{tabular}{ p{1.3cm} | p{0.5cm} }
    \multicolumn{1}{c|}{\textbf{Method}} & \multicolumn{1}{c}{\textbf{Description}} \\ \hline
    \multicolumn{1}{l|}{Long short-term memory (LSTM) \citep{hochreiter_schmidhuber}} & \multicolumn{1}{l}{Type of recurrent neural network (RNN) aimed at dealing with} \\
    \multicolumn{1}{l|}{} & \multicolumn{1}{l}{\,\, -- the vanishing gradient problem} \\
    \multicolumn{1}{l|}{Gated recurrent unit (GRU) \citep{chung_gulcehre_cho}} & \multicolumn{1}{l}{An LSTM with a gating mechanism to input or forget certain} \\
    \multicolumn{1}{l|}{} & \multicolumn{1}{l}{\,\, -- features, but lacks a context vector or output gate} \\
    \multicolumn{1}{l|}{Fully convolutional network (FCN) \citep{wang_yan_oates}} & \multicolumn{1}{l}{A CNN without fully connected layers} \\
    \multicolumn{1}{l|}{Temporal convolutional network (TCN) \citep{bai_kolter_koltun}} & \multicolumn{1}{l}{Consists of dilated, causal 1D convolutional layers with the} \\
    \multicolumn{1}{l|}{} & \multicolumn{1}{l}{\,\, -- same input and output lengths} \\
    \multicolumn{1}{l|}{Residual network (ResNet) \citep{wang_yan_oates}} & \multicolumn{1}{l}{A seminal deep learning model in which the weight layers} \\
    \multicolumn{1}{l|}{} & \multicolumn{1}{l}{\,\, -- learn residual functions with reference to the layer inputs} \\
    \multicolumn{1}{l|}{ResCNN \citep{zou_wang_li}} & \multicolumn{1}{l}{A model based on one-dimensional residual convolutional} \\
    \multicolumn{1}{l|}{} & \multicolumn{1}{l}{\,\, -- neural networks (1D-ResCNN)} \\
    \multicolumn{1}{l|}{InceptionTime \citep{fawaz_lucas_forestier}} & \multicolumn{1}{l}{Ensemble of five deep learning models for time-series} \\
    \multicolumn{1}{l|}{} & \multicolumn{1}{l}{\,\, -- classification, each one created by cascading multiple} \\
    \multicolumn{1}{l|}{} & \multicolumn{1}{l}{\,\, -- Inception modules} \\
    \multicolumn{1}{l|}{XceptionTime \citep{rahimian_zabihi_atashzar}} & \multicolumn{1}{l}{Designed by integration of depthwise separable convolutions,} \\
    \multicolumn{1}{l|}{} & \multicolumn{1}{l}{\,\, -- adaptive average pooling, and a novel non-linear} \\
    \multicolumn{1}{l|}{} & \multicolumn{1}{l}{\,\, -- normalization technique} \\
    \multicolumn{1}{l|}{OmniScaleCNN \citep{tang_long_liu}} & \multicolumn{1}{l}{Simple and effective kernel size configuration for time series} \\
    \multicolumn{1}{l|}{} & \multicolumn{1}{l}{\,\, -- classification} \\
    \multicolumn{1}{l|}{LSTM-FCN \citep{karim_majumdar_darabi}} & \multicolumn{1}{l}{A combination of an LSTM with a FCN} \\
    \multicolumn{1}{l|}{GRU-FCN \citep{elsaye_maida_bayoumi}} & \multicolumn{1}{l}{A combination of a GRU with a FCN} \\
    \multicolumn{1}{l|}{MLSTM-FCN \citep{karim_majumdar}} & \multicolumn{1}{l}{A multivariate LSTM-FCN that augments the squeeze and} \\
    \multicolumn{1}{l|}{} & \multicolumn{1}{l}{\,\, -- excitation blocks} \\
    \multicolumn{1}{l|}{Time series Transformer (TST) \citep{zerveas_jayaraman_patel}} & \multicolumn{1}{l}{Adding positional encodings with 1D convolutions} \\
    \multicolumn{1}{l|}{TSPerceiver \citep{jaegle_borgeaud_alayrac}} & \multicolumn{1}{l}{Adapted from PerceiverIO~\citep{jaegle_borgeaud_alayrac}} \\
    \multicolumn{1}{l|}{TSSequencerPlus \citep{tatsunami_taki}} & \multicolumn{1}{l}{Adapted from Sequencer~\citep{tatsunami_taki}} \\
    \multicolumn{1}{l|}{Multilevel wavelet decomposition network (mWDN) \citep{wang_wang_li}} & \multicolumn{1}{l}{Preserves the advantage of multilevel discrete wavelet} \\
    \multicolumn{1}{l|}{} & \multicolumn{1}{l}{\,\, -- decomposition in frequency learning while enables} \\
    \multicolumn{1}{l|}{} & \multicolumn{1}{l}{\,\, -- the fine-tuning of all parameters under a deep neural} \\
    \multicolumn{1}{l|}{} & \multicolumn{1}{l}{\,\, -- network framework} \\
    \multicolumn{1}{l|}{Explainable CNN (XCM) \citep{fauvel_fromont_masson}} & \multicolumn{1}{l}{Compact convolutional neural network which extracts} \\
    \multicolumn{1}{l|}{} & \multicolumn{1}{l}{\,\, -- extracts information relative to the observed variables and} \\
    \multicolumn{1}{l|}{} & \multicolumn{1}{l}{\,\, -- time directly from the input data} \\
    \multicolumn{1}{l|}{Gated multilayer perceptron (gMLP) \citep{liu_dai_so_le}} & \multicolumn{1}{l}{An MLP-based alternative to Transformers without self-} \\
    \multicolumn{1}{l|}{} & \multicolumn{1}{l}{\,\, -- attention, which simply consists of channel projections and} \\
    \multicolumn{1}{l|}{} & \multicolumn{1}{l}{\,\, -- spatial projections with static parameterization} \\
    \end{tabular}
\end{center}
\end{table}

\subsection{Machine Learning Methods}
\label{section_exp_ml_models}

For our cross-validation of all snapshot datasets, we initially benchmark 20 ML models. Table~\ref{table_ml_methods} summarizes current state-of-the-art architectures. We evaluate convolutional networks, such as FCN~\citep{wang_yan_oates}, TCN~\citep{bai_kolter_koltun}, ResNet~\citep{wang_yan_oates}, ResCNN~\citep{zou_wang_li}, InceptionTime~\citep{fawaz_lucas_forestier}, XceptionTime~\citep{rahimian_zabihi_atashzar}, and OmniScaleCNN~\citep{tang_long_liu}. Recurrent neural networks (RNNs) are a class of neural networks that are particularly well-suited for sequential data (as GNSS snapshots are) and tasks where the order of the data points matters. Hence, we evaluate LSTM~\citep{hochreiter_schmidhuber} and GRU~\citep{chung_gulcehre_cho} networks. We benchmark a combination of recurrent with fully convolutional networks, i.e., LSTM-FCN~\citep{karim_majumdar_darabi}, GRU-FCN~\citep{elsaye_maida_bayoumi}, and MLSTM-FCN~\citep{karim_majumdar}. Transformer models have gained significant popularity in the field of deep learning and natural language processing due to several key advantages and innovations they offer over previous architectures, such as RNNs and LSTMs, such as parallelization, handling long-range dependencies, good scalability, pre-training, and fine-tuning. Well known models are TST~\citep{zerveas_jayaraman_patel}, TSPerceiver~\citep{jaegle_borgeaud_alayrac}, and TSSequencerPlus~\citep{tatsunami_taki}. Further methods that show good performance on time-series classification tasks are mWDN~\citep{wang_wang_li}, XCM~\citep{fauvel_fromont_masson}, and gMLP~\citep{liu_dai_so_le}. For evaluation results, see Section~\ref{label_eval_ml_models}.

\begin{table}[t!]
\begin{center}
\setlength{\tabcolsep}{3.2pt}
    \caption{Overview of outlier detection methods. For an implementation, see the python outlier detection (PyOD)~\citep{pyod} toolbox. For an outlier detection benchmark, see ADBench~\citep{han_hu_huang}.}
    \label{table_outlier_detection}
    \small \begin{tabular}{ p{1.3cm} | p{0.5cm} }
    \multicolumn{1}{c|}{\textbf{Method}} & \multicolumn{1}{c}{\textbf{Description}} \\ \hline
    \multicolumn{1}{l|}{ECOD \citep{li_zhao_hu}} & \multicolumn{1}{l}{Unsupervised outlier detection using empirical cumulative} \\
    \multicolumn{1}{l|}{} & \multicolumn{1}{l}{\,\, -- distribution functions} \\
    \multicolumn{1}{l|}{Copula-based outlier detection (COPOD) \citep{li_zhao_botta}} & \multicolumn{1}{l}{Empirical copula, predict tail probabilities of each given data point} \\
    \multicolumn{1}{l|}{} & \multicolumn{1}{l}{\,\, -- to determine its level of \textit{extremeness}} \\
    \multicolumn{1}{l|}{Local outlier factor (LOF) \citep{breunig_kriegel}} & \multicolumn{1}{l}{Computes how isolated the object is with respect to the} \\
    \multicolumn{1}{l|}{} & \multicolumn{1}{l}{\,\, -- surrounding neighborhood} \\
    \multicolumn{1}{l|}{Clustering-based local outlier factor (CBLOF) \citep{he_xu_deng}} & \multicolumn{1}{l}{Measure for identifying the physical significance of an outlier} \\
    \multicolumn{1}{l|}{Median absolute deviation (MAD) \citep{iglewicz_hoaglin}} & \multicolumn{1}{l}{Outlier labeling aims to flag observations as possible outliers} \\
    \multicolumn{1}{l|}{Quasi-Monte carlo discrepancy outlier detection} & \multicolumn{1}{l}{Wrap-around version of the $L_2$-discrepancy (WD)} \\
    \multicolumn{1}{l|}{\,\, -- (QMCD) \citep{fang_ma}} & \multicolumn{1}{l}{} \\
    \multicolumn{1}{l|}{Sampling \citep{sugiyama_borgwardt}} & \multicolumn{1}{l}{Rapid distance-based outlier detection via sampling} \\
    \multicolumn{1}{l|}{GMM \citep{aggarwal}} & \multicolumn{1}{l}{Probabilistic mixture modeling for outlier analysis} \\
    \multicolumn{1}{l|}{One-class support vector machine (OCSVM) \citep{schoelkopf_platt}} & \multicolumn{1}{l}{Data set drawn from an underlying probability distribution $P$,} \\
    \multicolumn{1}{l|}{} & \multicolumn{1}{l}{\,\, -- estimate a \textit{simple} subset $S$ of input space such that the probability} \\
    \multicolumn{1}{l|}{} & \multicolumn{1}{l}{\,\, -- that a test point drawn from $P$ lies outside of $S$ equals some a} \\
    \multicolumn{1}{l|}{} & \multicolumn{1}{l}{\,\, -- priori specified value between 0 and 1} \\
    \multicolumn{1}{l|}{LMDD \citep{arning_agrawal}} & \multicolumn{1}{l}{Deviation-based outlier detection} \\
    \multicolumn{1}{l|}{INNE \citep{bandaragoda_ting}} & \multicolumn{1}{l}{Isolation-based anomaly detection using nearest neighbor ensembles} \\
    \multicolumn{1}{l|}{k nearest neighbors (kNN) \citep{ramaswamy_rastogi}} & \multicolumn{1}{l}{Uses the distance to the k-th nearest neighbor as the outlier score} \\
    \multicolumn{1}{l|}{Lightweight online detector of anomalies (LODA) \citep{pevny}} & \multicolumn{1}{l}{An ensemble of very weak detectors can lead to a strong anomaly} \\
    \multicolumn{1}{l|}{} & \multicolumn{1}{l}{\,\, -- detector,low time and space complexity} \\
    \multicolumn{1}{l|}{Clustering-based local outlier factor (CBLOF) \citep{he_xu_deng}} & \multicolumn{1}{l}{Measure for identifying the physical significance of an outlier} \\
    \multicolumn{1}{l|}{Principal component analysis (PCA) \citep{shyu_chen_sarinnapakorn}} & \multicolumn{1}{l}{Sum of weighted projected distances to the eigenvector hyperplanes} \\
    \multicolumn{1}{l|}{Minimum covariance determinant (MCD) \citep{hardin_rocke}} & \multicolumn{1}{l}{Uses the Mahalanobis distances as the outlier scores} \\
    \multicolumn{1}{l|}{Feature bagging (FB) \citep{lazarevic_kumar}} & \multicolumn{1}{l}{Detecting outliers in very large, high dimensional and noisy datasets,} \\
    \multicolumn{1}{l|}{} & \multicolumn{1}{l}{\,\, -- combines results from multiple outlier detection methods using} \\
    \multicolumn{1}{l|}{} & \multicolumn{1}{l}{\,\, -- different set of features (randomly selected)} \\
    \multicolumn{1}{l|}{Angle-based outlier Detection (ABOD) \citep{kriegel_schubert_zimek}} & \multicolumn{1}{l}{Assessing the variance in the angles between the difference vectors} \\
    \multicolumn{1}{l|}{} & \multicolumn{1}{l}{\,\, -- of a point to the other points} \\
    \multicolumn{1}{l|}{Isolation forest (IForest) \citep{liu_ting_zhou}} & \multicolumn{1}{l}{Isolates anomalies instead of profiles normal points, exploit sub-} \\
    \multicolumn{1}{l|}{} & \multicolumn{1}{l}{\,\, -- sampling that is not feasible in existing methods} \\
    \multicolumn{1}{l|}{Histogram-based outlier score (HBOS) \citep{goldstein_dengel}} & \multicolumn{1}{l}{Assumes independence of features making it much faster than} \\
    \multicolumn{1}{l|}{} & \multicolumn{1}{l}{\,\, -- multivariate approaches at the cost of less precision} \\
    \multicolumn{1}{l|}{Stochastic outlier selection (SOS) \citep{hanssens_huszar}} & \multicolumn{1}{l}{Takes as input either a feature matrix or a dissimilarity matrix and} \\
    \multicolumn{1}{l|}{} & \multicolumn{1}{l}{\,\, -- outputs for each data point an outlier probability} \\
    \multicolumn{1}{l|}{AutoEncoder \citep{aggarwal_sathe}} & \multicolumn{1}{l}{Fully connected autoencoder (uses reconstruction error as the} \\
    \multicolumn{1}{l|}{} & \multicolumn{1}{l}{\,\, -- outlier score), variant of subsampling and feature bagging} \\
    \multicolumn{1}{l|}{Variational autoencoder (VAE) \citep{kingma_welling}} & \multicolumn{1}{l}{Uses reconstruction error as the outlier score} \\
    \multicolumn{1}{l|}{SO-GAAL \citep{liu_li_zhou}} & \multicolumn{1}{l}{Single-objective generative adversarial active learning} \\
    \multicolumn{1}{l|}{MO-GAAL \citep{liu_li_zhou}} & \multicolumn{1}{l}{Multiple-objective generative adversarial active learning} \\
    \multicolumn{1}{l|}{Deep one-class classification (DeepSVDD) \citep{ruff_vandermeulen}} & \multicolumn{1}{l}{Trained on an anomaly detection based objective} \\
    \multicolumn{1}{l|}{Adversarially learned anomaly detection (ALAD) \citep{zenati_romain_foo}} & \multicolumn{1}{l}{Based on bidirectional GANs} \\
    \multicolumn{1}{l|}{RGraph \citep{you_robinson_vidal}} & \multicolumn{1}{l}{Outlier detection by RGraph} \\
    \multicolumn{1}{l|}{LUNAR \citep{goodge_hooi}} & \multicolumn{1}{l}{Unifying local outlier detection methods via graph neural networks} \\
    \end{tabular}
\end{center}
\end{table}

\subsection{Outlier Detection Methods}
\label{section_exp_outlier_detection}

We briefly summarize all outlier detection methods that we use for our benchmark in Section~\ref{label_eval_outlier_detection}. Table~\ref{table_outlier_detection} presents a comprehensive overview of these methods. Unsupervised outlier detection techniques are developed based on various assumptions about data distributions, such as the expectation that anomalies are found in low-density regions. The effectiveness of these unsupervised methods depends on how well the input data aligns with the underlying assumption of the algorithms. Numerous unsupervised techniques exist, spanning from shallow approach to deep neural networks, including those proposed by \cite{shyu_chen_sarinnapakorn,breunig_kriegel,he_xu_deng,goldstein_dengel,ramaswamy_rastogi,li_zhao_botta,ruff_vandermeulen,pevny,liu_ting_zhou}. Shallow methods generally offer greater interpretability, while deep methods are better suited for handling large, high-dimensional data~\citep{han_hu_huang}. A notable shallow method that performs well on our GNSS datasets is Isolation Forest (IForest), as introduced by \cite{liu_ting_zhou}. This method constructs an ensemble of tress to isolate data points, with the anomaly score defined as the distance of an individual instance from the root. Supervised methods are often impractical because they require labeled training data containing both normal and anomalous samples. However, there is ongoing research into the relationship between fully-supervised and semi-supervised approaches. Semi-supervised methods, such as DeepSAD~\citep{ruff_vandermeulen}, efficiently utilize partially labeled data to enhance detection performance while also leveraging unlabeled data to improve representation learning~\citep{han_hu_huang}.

\begin{table}[t!]
\begin{center}
    \caption{Overview of domain adaptation (DA) methods. For an implementation, see the AdaTime toolbox~\citep{ragab_eldele_tan}.}
    \label{table_da_methods}
    \small \begin{tabular}{ p{1.3cm} | p{0.5cm} }
    \multicolumn{1}{c|}{\textbf{Method}} & \multicolumn{1}{c}{\textbf{Description}} \\ \hline
    \multicolumn{1}{l|}{Mean squared error (MSE)} & \multicolumn{1}{l}{Frobenius norm to calculate the distance} \\
    \multicolumn{1}{l|}{Cosine similarity (CS)} & \multicolumn{1}{l}{Measures the orthogonality or decorrelation between two vectors} \\
    \multicolumn{1}{l|}{Pearson correlation (PC) \citep{karl_pearson}} & \multicolumn{1}{l}{A correlation coefficient that measures linear correlation between} \\
    \multicolumn{1}{l|}{} & \multicolumn{1}{l}{\,\, -- two sets of data} \\
    \multicolumn{1}{l|}{Kullback-Leibler (KL) divergence \citep{kullback_leibler}} & \multicolumn{1}{l}{A measure of how one probability distribution is different from} \\
    \multicolumn{1}{l|}{} & \multicolumn{1}{l}{\,\, -- a second} \\
    \multicolumn{1}{l|}{Jensen-Shannon (JS) divergence} & \multicolumn{1}{l}{Quantify the difference between two probability distributions} \\
    \multicolumn{1}{l|}{} & \multicolumn{1}{l}{\,\, -- by measuring the relative entropy or information divergence} \\
    \multicolumn{1}{l|}{Linear maximum mean discrepancy (MMD) \citep{borgwardt_gretton}} & \multicolumn{1}{l}{Align the first order statistics} \\
    \multicolumn{1}{l|}{Kernelized MMD (kMMD) \citep{long_zhu_mmd}} & \multicolumn{1}{l}{Uses a Gaussian radial basis function (RBF) kernel} \\
    \multicolumn{1}{l|}{Deep correlation alignment (CORAL) \citep{sun_feng_saenko}} & \multicolumn{1}{l}{Align the second-order statistics} \\
    \multicolumn{1}{l|}{Jeffreys \& Stein CORAL \citep{cherian_sra,suvrit_sra}} & \multicolumn{1}{l}{Affine invariant variations of CORAL} \\
    \multicolumn{1}{l|}{\,\, -- \citep{harandi_salzmann}} & \multicolumn{1}{l}{Affine invariant variations of CORAL} \\
    \multicolumn{1}{l|}{Maximum mean \& covariance discrepancy (MMCD)} & \multicolumn{1}{l}{Combination of MMD and CORAL} \\
    \multicolumn{1}{l|}{\,\, -- \citep{zhang_zhang_lan,alipour_tahmoresnezhad}} & \multicolumn{1}{l}{} \\
    \multicolumn{1}{l|}{Minimum discrepancy estimation for deep domain adaptation} & \multicolumn{1}{l}{Leverages a conditional entropy minimization technique} \\
    \multicolumn{1}{l|}{\,\, -- (MMDA) \citep{rahman_fookes}} & \multicolumn{1}{l}{} \\
    \multicolumn{1}{l|}{Deep domain confusion (DDC) \citep{tzeng_hoffman_zhang}} & \multicolumn{1}{l}{Utilizes kMMD} \\
    \multicolumn{1}{l|}{Deep adaptation network (DAN) \citep{long_cao_wang_DAN}} & \multicolumn{1}{l}{Embeds the hidden representations of all task-specific layers} \\
    \multicolumn{1}{l|}{} & \multicolumn{1}{l}{\,\, -- layers in an RKHS} \\
    \multicolumn{1}{l|}{Higher-order moment matching (HoMM) \citep{chen_fu_chen}} & \multicolumn{1}{l}{Align domains based on higher-order statistics} \\
    \multicolumn{1}{l|}{Domain-adversarial neural network (DANN) \citep{ganin_ustinova_ajakan}} & \multicolumn{1}{l}{Encourage the emergence of features that are discriminative} \\
    \multicolumn{1}{l|}{} & \multicolumn{1}{l}{\,\, -- for the source domain but indiscriminative with respect to the} \\
    \multicolumn{1}{l|}{} & \multicolumn{1}{l}{\,\, -- shift between the source and target domains} \\
    \multicolumn{1}{l|}{Conditional domain adversarial network (CDAN) \citep{long_cao_wang}} & \multicolumn{1}{l}{Learns representations that are disentangled and transferable by} \\
    \multicolumn{1}{l|}{} & \multicolumn{1}{l}{\,\, -- conditioning the adversarial adaptation models} \\
    \multicolumn{1}{l|}{} & \multicolumn{1}{l}{\,\, -- on discriminative information} \\
    \multicolumn{1}{l|}{Decision-boundary iterative refinement training with a teacher} & \multicolumn{1}{l}{Combination of domain adversarial training and a penalty term} \\
    \multicolumn{1}{l|}{\,\, -- (DIRT-T) \citep{shu_bui_narui}} & \multicolumn{1}{l}{\,\, -- that penalizes violations of the cluster assumption} \\
    \multicolumn{1}{l|}{Subdomain adaptation network (DSAN) \citep{zhu_zhuang_wang}} & \multicolumn{1}{l}{Transfer network which aligns subdomain distributions of domain-} \\
    \multicolumn{1}{l|}{} & \multicolumn{1}{l}{\,\, -- specific layers across different domains based on a local kMMD} \\
    \multicolumn{1}{l|}{Convolutional deep domain adaptation model for time-series data} & \multicolumn{1}{l}{Uses a weak supervision method in the form of target-domain} \\
    \multicolumn{1}{l|}{\,\, -- (CoDATS) \citep{wilson_doppa_cook}} & \multicolumn{1}{l}{\,\, -- label distributions} \\
    \multicolumn{1}{l|}{Adversarial spectral kernel matching (AdvSKM) \citep{liu_xue}} & \multicolumn{1}{l}{Improves the MMD metric through the use of a hybrid kMMD} \\
    \multicolumn{1}{l|}{} & \multicolumn{1}{l}{\,\, -- spectral kernel that allows for a more accurate characterization} \\
    \multicolumn{1}{l|}{} & \multicolumn{1}{l}{\,\, -- of non-stationary and non-monotonic statistics within time-} \\
    \multicolumn{1}{l|}{} & \multicolumn{1}{l}{\,\, -- series distributions} \\
    \multicolumn{1}{l|}{Optimal transport-based DA \citep{ott_acmmm}} & \multicolumn{1}{l}{Domain adaptation based on optimal transport} \\
    \multicolumn{1}{l|}{Sinkhorn-based DA \citep{ott_dissertation}} & \multicolumn{1}{l}{Domain adaptation based on Sinkhorn} \\
    \end{tabular}
\end{center}
\end{table}

\subsection{Domain Adaptation Methods}
\label{section_exp_da_models}

Domain adaptation (DA) methods primarily aim to identify an optimal distance metric for measuring the discrepancy between feature embeddings of the source and target domains. Table~\ref{table_da_methods} offers a detailed overview of various DA methods. A commonly used metric is the mean squared error (MSE) distance. However, MSE has significant limitations: it is sensitive to outliers because it assigns greater weight to large errors by squaring the differences between predicted and actual values. In DA, where the data distributions of the source and target domains can differ, outliers in the target domain may disproportionately influence model training, resulting in poor generalization. MSE assumes that the data distribution is similar across both domains, which is often not the case in DA applications where the target domain's data distribution typically differs from that of the source domain. Moreover, MSE focuses on reducing average error, potentially overlooking important aspects such as distributional differences between domains. 

Cosine similarity, which measures the cosine of the angle between two non-zero vectors in multidimensional space, is frequently employed in cross-modal learning. As a result, distribution-based metrics like Kullback-Leibler (KL)~\citep{kullback_leibler} and Jensen-Shannon (JS) divergences are often used to quantify the differences between probability distributions. However, KL divergence has the drawback of being asymmetric and sensitive to zero probabilities, which can be particularly problematic in DA when the source and target domains have non-overlapping support (e.g., when certain features or data points are present in one domain but absent in the other). JS divergence, a symmetrized and smoothed version of KL divergence, is less sensitive to small discrepancies between distributions because it averages out differences.

A widely adopted metric in DA is the maximum mean discrepancy (MMD), as proposed by \cite{borgwardt_gretton}, which aligns the first-order statistics between domains. The kernelized MMD~\citep{long_zhu_mmd}, which utilizes a Gaussian radial basis function kernel, enhances this approach. Deep correlation alignment (CORAL) further refines the process by aligning the second-order statistics, often leading to improved DA accuracy~\citep{sun_feng_saenko}. MMCD~\citep{zhang_zhang_lan,alipour_tahmoresnezhad} combines MMD and CORAL into a single loss function. Numerous other advancements leverage the MMD loss for DA~\citep{rahman_fookes,tzeng_hoffman_zhang,long_cao_wang_DAN,ganin_ustinova_ajakan}. Higher-order moment matching (HoMM)~\citep{chen_fu_chen}, which aligns domains based on higher-order statistics, requires significant computational resources. Recent methods employ optimal transport techniques to map the source domain onto the target domain~\citep{ott_acmmm,ott_dissertation}. We benchmark all 24 DA methods and present the results in Section~\ref{label_eval_domain_adaptation}.

\subsection{Data Augmentation}
\label{section_exp_data_augmentation}

Data augmentation is a technique employed in image classification to artificially increase the size and diversity of a training dataset by generating modified versions of existing images. This method enhances the performance of ML models by increasing the dataset size, improving model generalization by reducing overfitting, introducing variability (simulating real-world variations), and balancing classes (addressing the significant class imbalance often present in GNSS datasets, see Table~\ref{table_data_combination}). First, we normalize the GNSS snapshots to the range $[0, 1]$. Subsequently, with a probability of 50\%, the following data augmentation techniques are applied to every snapshot:
\begin{enumerate}
    \item \textbf{Flip:} Random horizontal and vertical flipping of the image.
    \item \textbf{Noise:} Addition of a normally distributed noise, with the noise parameter randomly selected between $[-0.1, 0.1]$.
    \item \textbf{Multiply intensity:} Multiplication of the entire snapshot by a normally distributed value randomly selected between $[0.8, 1.2]$.
    \item \textbf{Zoom:} Creating a new sample by selecting a subset of the snapshot. Specifically, the length parameter of the height and width of the snapshot is randomly selected between $[0.7, 1.0]$.
\end{enumerate}
Following the application of these data augmentation techniques, the possible range of snapshot values extends to $[-0.12, 1.32]$. For evaluation results, see Section~\ref{label_eval_data_augmentation}.
\section{Evaluation}
\label{label_evaluation}

First, we examine the impact of interferences on both visible satellites and snapshot data using a controlled indoor dataset (refer to Section~\ref{label_eval_infl1}), subsequently assess the impact on positiong, velocity, and time (PVT) and snapshot data within a highway dataset (refer to Section~\ref{label_eval_infl2}). Following this, we analyze the evaluation results of cross-validation for the ResNet18 model across all snapshot datasets (refer to Section~\ref{label_eval_cross_validation}) and compare them with state-of-the-art ML methods (refer to Section~\ref{label_eval_ml_models}). Next, we assess the effectiveness of our proposed pseudo-labeling method in Section~\ref{label_eval_pseudo_labeling}. We then present the evaluation results for various outlier detection and domain adaptation methods applied to the LC datasets in Section~\ref{label_eval_outlier_detection} and Section~\ref{label_eval_domain_adaptation}, respectively. Finally, we predict the driving direction using the Seetal Alps dataset (refer to Section~\ref{label_eval_driving_direction}) and discuss the results of data augmentation experiments (refer to Section~\ref{label_eval_data_augmentation}). All experiments were conducted using Nvidia Tesla V100-SXM2 GPUs with 32 GB VRAM, alongside Core Xeon CPUs and 192 GB RAM. The vanilla SGD optimizer was utilized, with a learning rate of $0.01$, a decay rate of $5 \times 10^4$, a momentum of $0.9$, and a batch size of 64. We train for 200 epochs, and apply a multistep learning rate with the milestones $[120, 160]$ and a gamma of $0.1$. Throughout the experiments, we report the accuracy after the final epoch or for the epoch with the lowest loss (in \%), as well as the F1-score and F2-score, with the best results highlighted in \textbf{bold}.

\begin{figure}[!t]
    \centering
	\begin{minipage}[t]{1.0\linewidth}
    \includegraphics[trim=0 292 0 0, clip, width=0.9\linewidth]{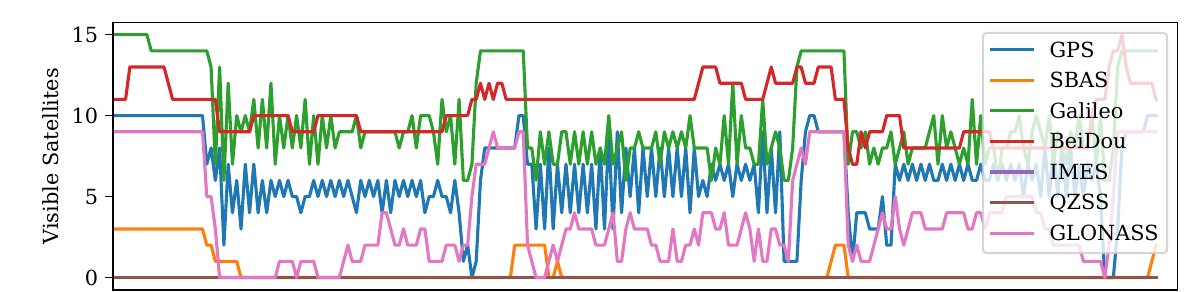}
    \end{minipage}
	\begin{minipage}[t]{1.0\linewidth}
    \includegraphics[trim=0 176 0 176, clip, width=0.9\linewidth]{images/influence_jammer/influence_jammer_lc_snapshot.pdf}
    \end{minipage}
	\begin{minipage}[t]{1.0\linewidth}
    \includegraphics[trim=0 0 0 300, clip, width=0.9\linewidth]{images/influence_jammer/influence_jammer_lc_snapshot.pdf}
    \end{minipage}
    \caption{Visualization of the influence of different \textit{chirp} interferences with three different bandwidths on the satellites for the controlled large-scale dataset 2. Top figure shows the number of visible satellites. Middle figure shows the snapshots of a high-frequency antenna. Bottom figure shows the snapshots of a low-frequency antenna.}
    \label{figure_influence_lc_snapshot}
\end{figure}

\subsection{Influence of Interferences on Visible Satellites \& Snapshot Data}
\label{label_eval_infl1}

First, we examine the impact of jamming devices on GNSS antennas and the satellite signals received from a u-blox module. Figure~\ref{figure_influence_lc_snapshot} illustrates the number of visible satellites over time for various systems: GPS, SBAS, Galileo, BeiDou, IMES, QZSS, and GLONASS (top), during the recording of the controlled large-scale dataset 2 from Section~\ref{label_data_contr_lf2} and Section~\ref{label_data_contr_hf2}. The middle and bottom panels of the figure depict synchronized snapshots of signals from the high-frequency and low-frequency antennas, respectively, corresponding with the satellite signals. These snapshots reveal three distinct \textit{chirp} interferences, each characterized by different bandwidths (as indicated by the varying widths of the interferences). Notably, the number of visible satellites decreases significantly during periods of active interference. For Galileo, the number of visible satellites decreases more with higher interference levels, while for GPS and GLONASS, the number of visible satellites increases. This indicates that the impact of interference on satellite signals depends on the specific characteristics of the interference.

\begin{figure}[!t]
    \centering
    \includegraphics[trim=10 28 10 40, clip, width=0.75\linewidth]{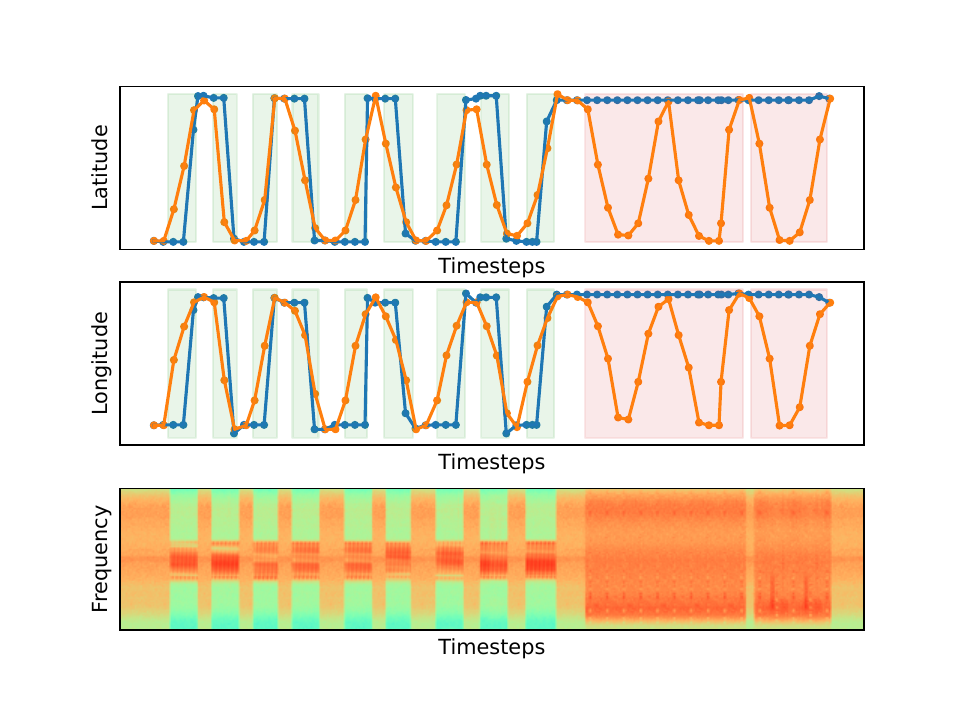}
    \caption{Visualization of the influence of interferences on the \textit{latitude} (top) and \textit{longitude} (middle) of a smartphone for the real-world highway dataset 2. The jammer and smartphone is integrated in the vehicle. The orange line shows the vehicle reference position. The blue line is the smartphone position. The green (handheld jammer) and red (cigarette lighter) marked backgrounds are the interfered timesteps.}
    \label{figure_influence_braunschweig}
\end{figure}

\subsection{Influence of Interferences on PVT \& Snapshot Data}
\label{label_eval_infl2}

Furthermore, we examine the effects of jamming devices on PVT in real-world applications. Specifically, we analyze the latitude and longitude recorded by smartphones equipped with a jammer within the same vehicle during the collection of the real-world highway dataset 2, as described in Section~\ref{label_data_rw_2}. Figure~\ref{figure_influence_braunschweig} presents the latitude (top), longitude (middle), and snapshot data (bottom) for several test drives in both directions. The orange line represents the reference position of the vehicle, while the blue line represents the position reported by the smartphone. When the jamming device is activated, the smartphone's reported position remains unchanged (blue line). We also tested two different jamming devices: the blue handheld jammer depicted in Figure~\ref{figure_jamming_devices} (green area) and a jammer installed in the vehicle's cigarette lighter (red area). Since the jammer in the cigarette lighter was not turned off, the smartphone's reported position was continuously disrupted. The handheld jammer was observed to have a narrower bandwidth. Both jamming devices are clearly identifiable in the GNSS snapshots and cause interference with the smartphone's position.

\begin{figure}[!t]
	\begin{minipage}[t]{1.0\linewidth}
        \centering
    	\includegraphics[trim=0 0 0 0, clip, width=1.0\linewidth]{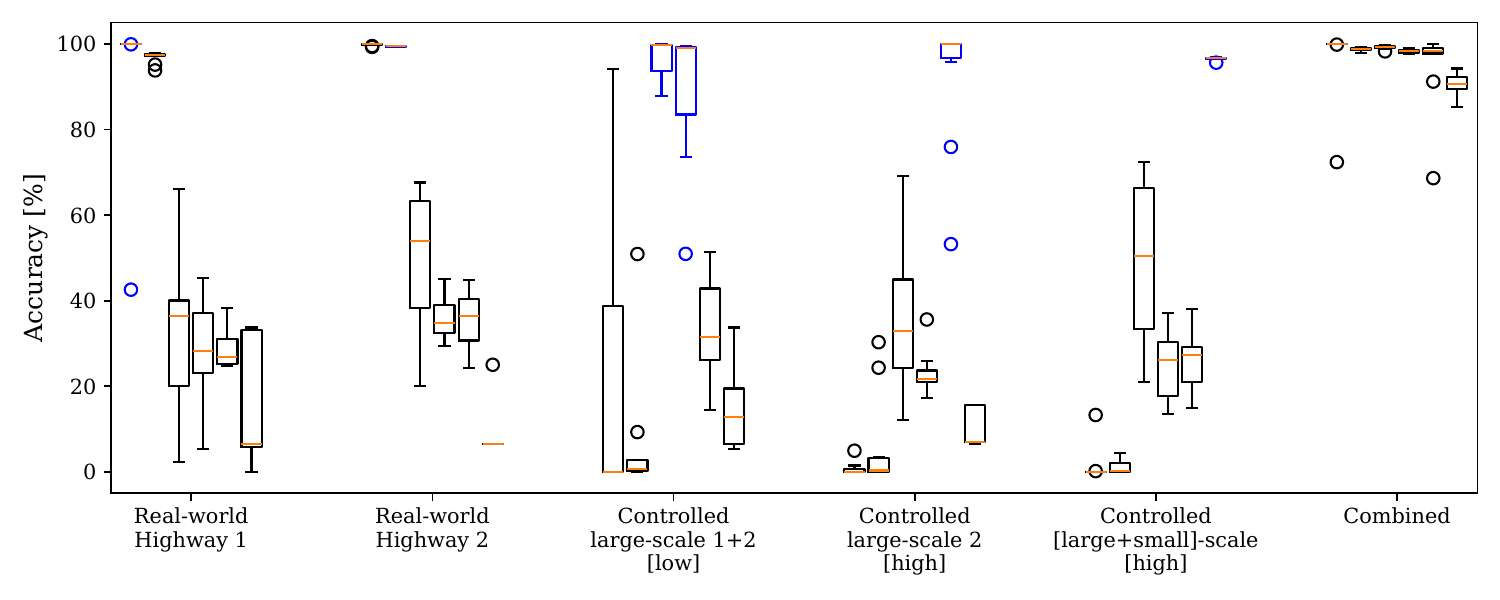}
        \subcaption{Accuracy [\%].}
        \label{figure_cross_validation1}
    \end{minipage}
    \hfill
	\begin{minipage}[t]{1.0\linewidth}
        \centering
    	\includegraphics[trim=0 0 0 0, clip, width=1.0\linewidth]{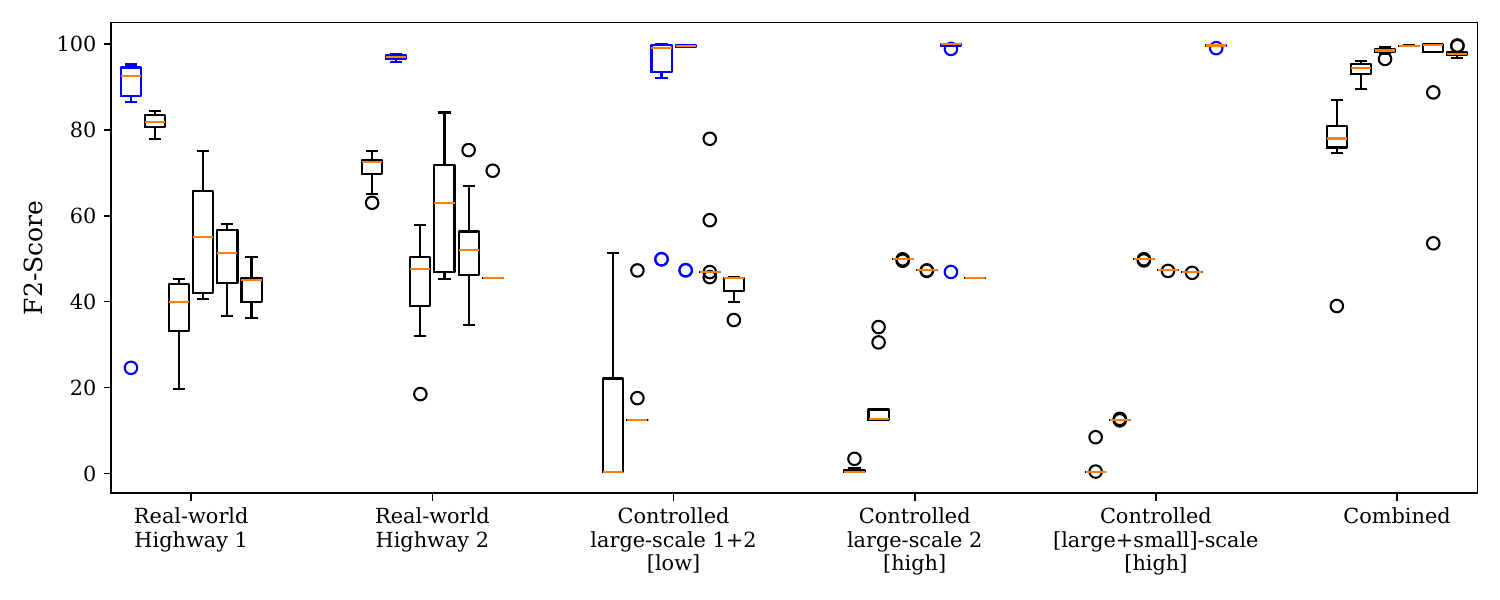}
        \subcaption{F2-score.}
        \label{figure_cross_validation2}
    \end{minipage}
    \caption{Cross-validation of all snapshot datasets. Blue marks an equal train and test dataset. The box plots show the mean and standard deviation averaged over 10 trainings of the same model with randomly chosen seed.}
    \label{figure_cross_validation}
\end{figure}

\subsection{Experimental Results on Cross-Validating Snapshot Data}
\label{label_eval_cross_validation}

Figure~\ref{figure_cross_validation} presents a summary of the cross-validation results obtained using the ResNet18 model. Additionally, we trained the model on a combined dataset and conducted tests on individual datasets. The test datasets were evaluated in the following order: real-world highway dataset 1 ($1^{\text{st}}$), real-world highway dataset 2 ($2^{\text{nd}}$), controlled large-scale dataset 1+2 with a low-frequency antenna tested separately ($3^{\text{rd}}$ and $4^{\text{th}}$), controlled large-scale dataset 2 with a high-frequency antenna ($5^{\text{th}}$), and the controlled [large+small]-scale dataset with a high-frequency antenna ($6^{\text{th}}$). Generally, the F2-score is slightly lower than the accuracy, attributed to the unbalanced nature of the dataset and the prevalence of non-interference classes. We achieved 99.9\% classification accuracy on the real-world datasets; however, this accuracy decreased for the controlled large-scale datasets due to the presence of multipath effects. Testing on a dataset different from the training set significantly reduced performance, indicating a substantial data discrepancy between the datasets. Specifically, training on indoor datasets and testing on real-world highway datasets proved ineffective. When trained on the combined data, the model achieved an accuracy of approximately 91\%, demonstrating its capability to learn features across different environments. Nevertheless, the use of a robust domain adaptation and transfer learning approach could further mitigate the observed data discrepancies.

\begin{figure}[!t]
	\begin{minipage}[t]{0.495\linewidth}
        \centering
    	\includegraphics[trim=11 11 10 11, clip, width=1.0\linewidth]{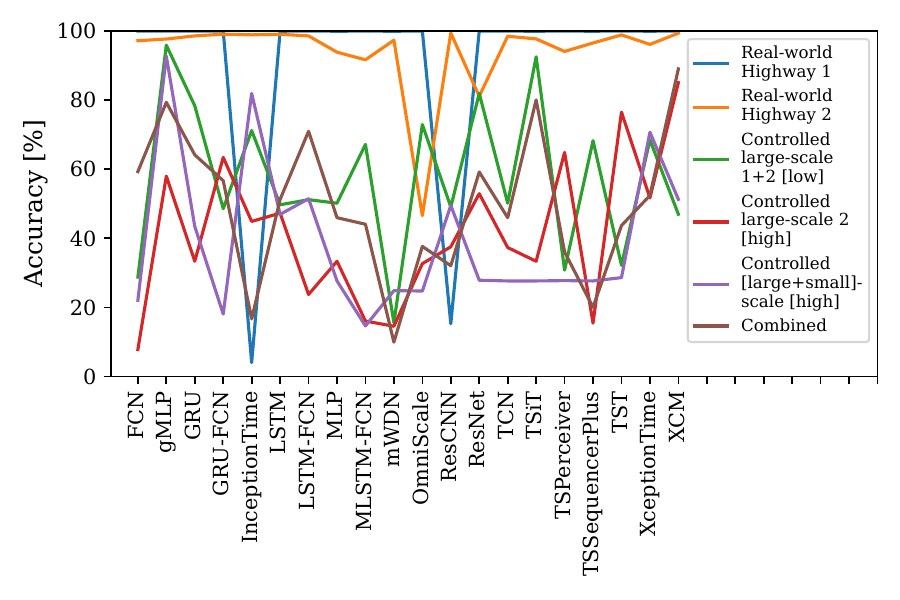}
        \subcaption{Accuracy [\%].}
        \label{figure_tsai_results1}
    \end{minipage}
    \hfill
	\begin{minipage}[t]{0.495\linewidth}
        \centering
    	\includegraphics[trim=11 11 10 11, clip, width=1.0\linewidth]{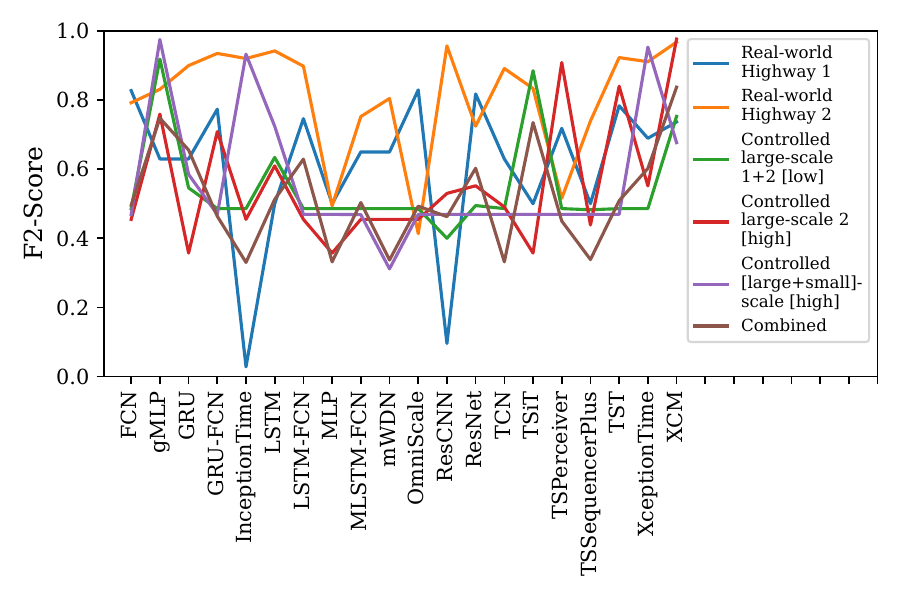}
        \subcaption{F2-score.}
        \label{figure_tsai_results2}
    \end{minipage}
    \caption{Evaluation results for all ML methods for all datasets tested on the equal train/test set.}
    \label{figure_tsai_results}
\end{figure}

\subsection{Evaluation of Machine Learning Methods on Snapshot Data}
\label{label_eval_ml_models}

\small \paragraph{Hyperparameters.} For the ML models, we use the following default hyperparameters: FCN ($\text{layers = [128, 256, 128]}$, $\text{kss = [7, 5, 3]}$), InceptionTime ($\text{ks} = 40$, $\text{bottleneck} = \text{True}$), MLP ($\text{layers = [50, 50]}$, $\text{ps = [0.1, 0.2]}$), OmniScaleCNN ($\text{layers} = [8 \cdot 128, 5 \cdot 128 \cdot 256 + 2 \cdot 256 \cdot 128]$), ResCNN ($\text{ks} = [7, 5, 3]$), ResNet ($\text{ks} = [7, 5, 3]$), TSPerceiver ($n_{\text{latents}} = 512$, $d_{\text{latent}} = 128$, $n_{\text{self\_heads}} = 8$). Besides that, we set for several models specific parameters, such as the hidden\_size to 50 for LSTM, GRU, MLP, LSTM-FCN, GRU-FCN, MLSTM-FCN. For mWDN, we set the levels to 1. For TSPerceiver and TST, we set n\_layers to 2. For TST, TSiT, gMLP, and TSSequencerPlus, we set d\_model to 64. For InceptionTime, TSiT, gMLP, and TSSequencerPlus, we set depth to 4. For InceptionTime, XceptionTime, and XCM, we set the parameter nf to 64.

\normalsize Figure~\ref{figure_tsai_results} presents the benchmark results of all ML models discussed in Section~\ref{section_exp_ml_models} on our GNSS datasets. The results indicate that most models perform well on the real-world highway datasets 1 and 2. However, their performance significantly declines on the controlled indoor datasets. Moreover, when combining the training and test datasets, there is a marked decrease in accuracy, although our ResNet18-based model continues to demonstrate robust performance (see Figure~\ref{figure_cross_validation1}). Not all models are capable of addressing the data discrepancies inherent in the GNSS datasets. The models gMLP~\citep{liu_dai_so_le} and TSiT~\citep{zerveas_jayaraman_patel} consistently perform well. Time-series-based models such as GRU~\citep{chung_gulcehre_cho}, LSTM~\citep{hochreiter_schmidhuber}, MLSTM-FCN~\citep{karim_majumdar}, and TCN~\citep{bai_kolter_koltun} tend to overfit, as the time-series component is less significant than the frequency component in this context. Smaller networks, such as FCN~\citep{wang_yan_oates} and MLP, are significantly outperformed by larger networks like ResNet~\citep{wang_yan_oates} and ResCNN~\citep{zou_wang_li}. Our ResNet model has a number of 11,181,129 parameters. In conclusion, we recommend utilizing a large architecture with a high parameter count for GNSS interference monitoring. For the real-world highway dataset 1, there exist significantly more samples from non-interference classes, and hence, the F2-score is more meaningful (see Figure~\ref{figure_tsai_results2}). Here, we achieve 0.95 F2-score with gMLP. The controlled [large+small]-scale dataset is balanced over all classes, and hence, the accuracy is more meaningful.

\begin{figure}[!t]
    \centering
	\begin{minipage}[t]{0.45\linewidth}
        \centering
        \includegraphics[trim=25 11 10 11, clip, width=1.0\linewidth]{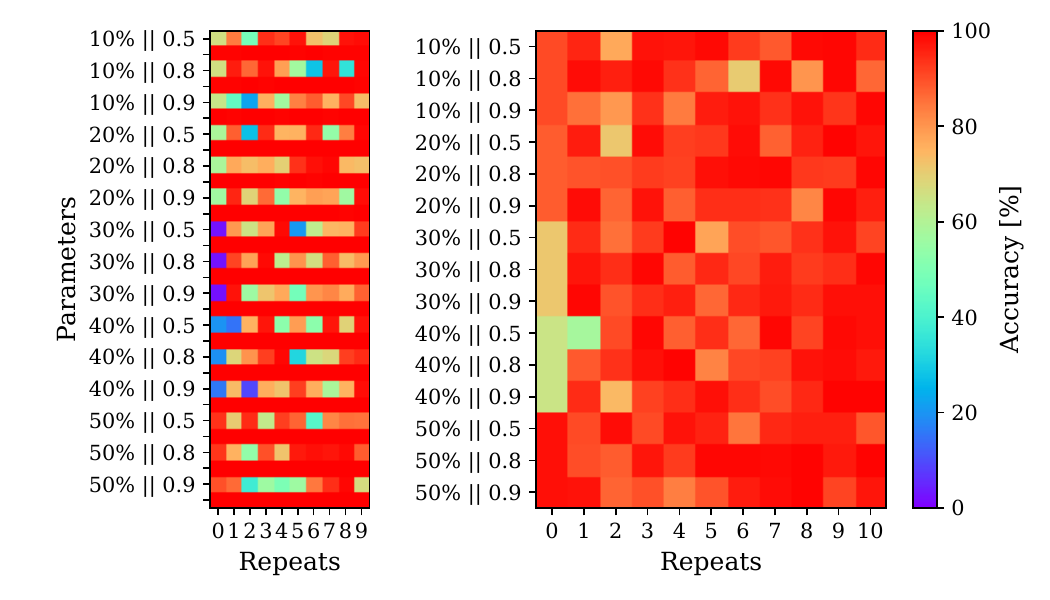}
        \subcaption{Pseudo-labeling for the interference classification task. The first parameter denotes the percentage of pre-defined labels used from the model. The second parameter is the threshold for the softmax output.}
        \label{figure_results_pseudo_labeling1}
    \end{minipage}
    \hfill
	\begin{minipage}[t]{0.54\linewidth}
        \centering
        \includegraphics[trim=0 11 10 11, clip, width=1.0\linewidth]{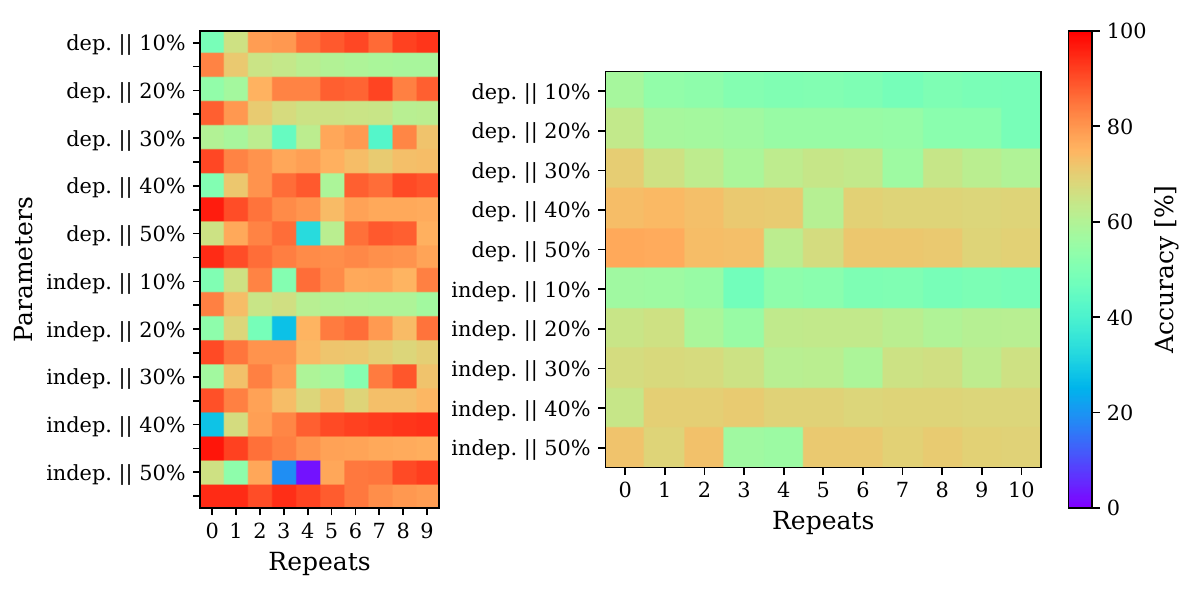}
        \subcaption{Pseudo-labeling for the multipath scenario classification task. For the dependent and independent scenario classification tasks, we evaluate 10\% to 50\% of labeled data.}
        \label{figure_results_pseudo_labeling2}
    \end{minipage}
    \caption{Evaluation results for our pseudo-labeling method on the combined controlled large-scale dataset 1+2 for four ResNet18 models.}
    \label{figure_results_pseudo_labeling}
\end{figure}

\subsection{Evaluation of Pseudo-Labeling Methods on Snapshot Data}
\label{label_eval_pseudo_labeling}

Figure~\ref{figure_results_pseudo_labeling} presents the results of our pseudo-labeling method applied to the combined large-scale controlled dataset 1+2. Specifically, Figure~\ref{figure_results_pseudo_labeling1} illustrates the performance of pseudo-labeling in the interference classification task. The first row indicates the percentage of unlabeled data on which all four ResNet18 models reached a consensus. The second row shows the percentage of data on which all four ResNet18 models agreed and were actually correct. Notably, when all four models agree, the percentage of correctly classified samples reaches 99.98\%. The final accuracy only marginally increases with a higher percentage of labeled data (10\%, 20\%, 30\%, 40\%, and 50\%). Thus, the portion of labeled data can be reduced, thereby decreasing the time required for data labeling. However, with 50\% labeled data, the models demonstrate greater robustness due to continuous training progress (as shown in the last row of Figure~\ref{figure_results_pseudo_labeling}, right). Regarding the number of iterations, the models converge after seven repetitions. Without labels (i.e., at iteration 0), the models fail to perform effectively. A higher softmax threshold of 0.9 consistently results in higher accuracy, as it leads to fewer but better-labeled data. Figure~\ref{figure_results_pseudo_labeling2} depicts the results of pseudo-labeling for the multipath scenario classification task. In this context, the percentage of labeled data plays a more critical role, as 10\% labeled data leads to a final accuracy of 40\%, while 50\% labeled data results in an accuracy between 70\% and 80\%. The final accuracy is notably lower for the independent dataset. At the early stages of the training process, the model struggles to learn meaningful patterns, making pseudo-labeling less effective than in the interference classification task. The baseline accuracy of a single ResNet model trained on 100\% labeled data, representing the upper bound, is 81\%. Therefore, further improvements could be achieved through data augmentation, for instance.

\begin{figure}[!t]
	\begin{minipage}[t]{0.495\linewidth}
    \centering
        \includegraphics[trim=11 11 0 11, clip, width=1.0\linewidth]{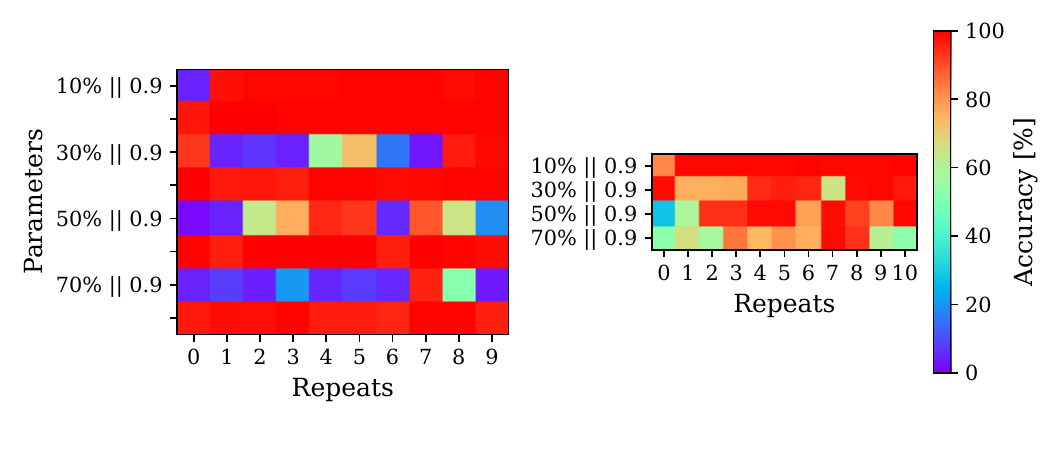}
        \subcaption{Accuracy [\%].}
        \label{figure_results_pseudo_labeling_tc1}
    \end{minipage}
    \hfill
	\begin{minipage}[t]{0.495\linewidth}
    \centering
        \includegraphics[trim=11 11 0 11, clip, width=1.0\linewidth]{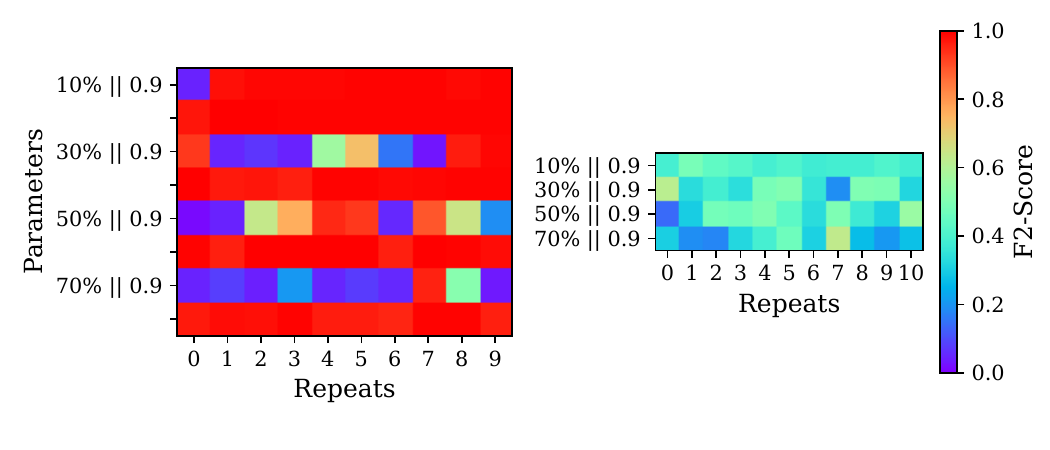}
        \subcaption{F2-score.}
        \label{figure_results_pseudo_labeling_tc2}
    \end{minipage}
    \caption{Evaluation results for our pseudo-labeling method on the real-world highway dataset 1 for four ResNet18 models. The first parameter denotes the percentage of pre-defined labels used from the model. The second parameter is the threshold for the softmax output.}
    \label{figure_results_pseudo_labeling_tc}
\end{figure}

Figure~\ref{figure_results_pseudo_labeling_tc} presents the results of pseudo-labeling on the real-world highway dataset 1. When using 10\% of labeled data, the models predominantly predict non-interference classes, with only a few predictions in the interference classes. However, with 70\% of labeled data, the models become less robust, as some predict only non-interference classes, while others predict more interference classes compared to the 10\% labeled data scenario. This variability is not apparent in Figure~\ref{figure_results_pseudo_labeling_tc} because the pseudo-labeling method was trained 10 times, and the results were averaged.

\begin{table}[t!]
\begin{center}
    \caption{Results for VAT~\citep{miyato_maeda_koyama} on the controlled large-scale low-frequency dataset for adapting to new scenarios. We evaluate different percentages of labeled data, and interference type dependency between training and test set.}
    \label{table_results_vat}
    \small \begin{tabular}{ p{1.3cm} | p{0.5cm} | p{0.5cm} | p{0.5cm} | p{0.5cm} | p{0.5cm} | p{0.5cm} }
    \multicolumn{1}{c||}{\textbf{Percentage}} & \multicolumn{1}{c|}{\textbf{Dependency}} & \multicolumn{1}{c|}{\textbf{Baseline}} & \multicolumn{1}{c||}{\textbf{VAT}} & \multicolumn{1}{c|}{\textbf{Dependency}} & \multicolumn{1}{c|}{\textbf{Baseline}} & \multicolumn{1}{c}{\textbf{VAT}} \\ \hline
    \multicolumn{1}{r||}{10\%} & \multicolumn{1}{l|}{dependent} & \multicolumn{1}{r|}{59.252} & \multicolumn{1}{r||}{\textbf{62.022}} & \multicolumn{1}{l|}{independent} & \multicolumn{1}{r|}{\textbf{60.049}} & \multicolumn{1}{r}{55.338} \\
    \multicolumn{1}{r||}{20\%} & \multicolumn{1}{l|}{dependent} & \multicolumn{1}{r|}{64.522} & \multicolumn{1}{r||}{\textbf{73.248}} & \multicolumn{1}{l|}{independent} & \multicolumn{1}{r|}{\textbf{64.951}} & \multicolumn{1}{r}{62.139} \\
    \multicolumn{1}{r||}{30\%} & \multicolumn{1}{l|}{dependent} & \multicolumn{1}{r|}{\textbf{73.105}} & \multicolumn{1}{r||}{69.873} & \multicolumn{1}{l|}{independent} & \multicolumn{1}{r|}{67.686} & \multicolumn{1}{r}{\textbf{69.130}} \\
    \multicolumn{1}{r||}{40\%} & \multicolumn{1}{l|}{dependent} & \multicolumn{1}{r|}{77.420} & \multicolumn{1}{r||}{\textbf{81.736}} & \multicolumn{1}{l|}{independent} & \multicolumn{1}{r|}{\textbf{73.993}} & \multicolumn{1}{r}{72.910} \\
    \multicolumn{1}{r||}{50\%} & \multicolumn{1}{l|}{dependent} & \multicolumn{1}{r|}{80.398} & \multicolumn{1}{r||}{\textbf{82.691}} & \multicolumn{1}{l|}{independent} & \multicolumn{1}{r|}{\textbf{75.059}} & \multicolumn{1}{r}{71.600} \\
    \end{tabular}
\end{center}
\end{table}

Table~\ref{table_results_vat} presents a comparison between the baseline method and VAT~\citep{miyato_maeda_koyama} for adapting to new scenarios on the controlled large-scale low-frequency dataset. In the independent classification task, VAT underperforms relative to the baseline, likely due to the scarcity of subjammers in the dataset, which makes it challenging for VAT to accurately guide the training process. However, in the dependent classification task, VAT proves beneficial in predicting the correct training direction. It is evident that increasing the amount of labeled data enhances the performance of the classification task.

\begin{table}[t!]
\begin{center}
    \caption{Results for our baseline ResNet18 model (top) and VAT~\citep{miyato_maeda_koyama} (bottom) on the real-world highway dataset 1 for adapting to new interferences. We evaluate different percentages of labeled data.}
    \label{table_results_vat2}
    \small \begin{tabular}{ p{1.3cm} | p{0.5cm} | p{0.5cm} | p{0.5cm} | p{0.5cm} | p{0.5cm} | p{0.5cm} | p{0.5cm} }
    \multicolumn{1}{c||}{\textbf{Percentage}} & \multicolumn{1}{c|}{\textbf{Multi-Accuracy [\%]}} & \multicolumn{1}{c|}{\textbf{Accuracy [\%]}} & \multicolumn{1}{c|}{\textbf{F1-Score}} & \multicolumn{1}{c|}{\textbf{F2-Score}} & \multicolumn{1}{c|}{\textbf{F5-Score}} & \multicolumn{1}{c|}{\textbf{Precision}} & \multicolumn{1}{c}{\textbf{Recall}} \\ \hline
    \multicolumn{1}{r||}{10\%} & \multicolumn{1}{r|}{99.144} & \multicolumn{1}{r|}{99.911} & \multicolumn{1}{r|}{0.55696} & \multicolumn{1}{r|}{0.45082} & \multicolumn{1}{r|}{0.40886} & \multicolumn{1}{r|}{0.91667} & \multicolumn{1}{r}{0.40000} \\
    \multicolumn{1}{r||}{30\%} & \multicolumn{1}{r|}{99.159} & \multicolumn{1}{r|}{99.921} & \multicolumn{1}{r|}{0.62651} & \multicolumn{1}{r|}{0.52419} & \multicolumn{1}{r|}{0.48182} & \multicolumn{1}{r|}{0.92857} & \multicolumn{1}{r}{0.47273} \\
    \multicolumn{1}{r||}{50\%} & \multicolumn{1}{r|}{99.136} & \multicolumn{1}{r|}{99.898} & \multicolumn{1}{r|}{0.64286} & \multicolumn{1}{r|}{0.64982} & \multicolumn{1}{r|}{0.65363} & \multicolumn{1}{r|}{0.63158} & \multicolumn{1}{r}{0.65455} \\
    \multicolumn{1}{r||}{70\%} & \multicolumn{1}{r|}{99.281} & \multicolumn{1}{r|}{99.942} & \multicolumn{1}{r|}{0.79279} & \multicolumn{1}{r|}{0.79710} & \multicolumn{1}{r|}{0.79944} & \multicolumn{1}{r|}{0.78571} & \multicolumn{1}{r}{0.80000} \\
    \multicolumn{1}{r||}{100\%} & \multicolumn{1}{r|}{\textbf{99.972}} & \multicolumn{1}{r|}{\textbf{99.980}} & \multicolumn{1}{r|}{\textbf{0.92453}} & \multicolumn{1}{r|}{\textbf{0.90406}} & \multicolumn{1}{r|}{\textbf{0.89341}} & \multicolumn{1}{r|}{\textbf{0.96078}} & \multicolumn{1}{r}{\textbf{0.89091}} \\ \hline
    \multicolumn{1}{r||}{10\%} & \multicolumn{1}{r|}{99.251} & \multicolumn{1}{r|}{99.914} & \multicolumn{1}{r|}{0.57500} & \multicolumn{1}{r|}{0.46939} & \multicolumn{1}{r|}{0.42714} & \multicolumn{1}{r|}{0.92000} & \multicolumn{1}{r}{0.41818} \\
    \multicolumn{1}{r||}{30\%} & \multicolumn{1}{r|}{99.253} & \multicolumn{1}{r|}{99.883} & \multicolumn{1}{r|}{0.61017} & \multicolumn{1}{r|}{0.63604} & \multicolumn{1}{r|}{0.65090} & \multicolumn{1}{r|}{0.57143} & \multicolumn{1}{r}{0.65455} \\
    \multicolumn{1}{r||}{50\%} & \multicolumn{1}{r|}{\textbf{99.383}} & \multicolumn{1}{r|}{\textbf{99.944}} & \multicolumn{1}{r|}{\textbf{0.76596}} & \multicolumn{1}{r|}{0.69498} & \multicolumn{1}{r|}{0.66195} & \multicolumn{1}{r|}{\textbf{0.92308}} & \multicolumn{1}{r}{0.65455} \\
    \multicolumn{1}{r||}{70\%} & \multicolumn{1}{r|}{99.286} & \multicolumn{1}{r|}{99.901} & \multicolumn{1}{r|}{0.69767} & \multicolumn{1}{r|}{\textbf{0.76531}} & \multicolumn{1}{r|}{\textbf{0.80745}} & \multicolumn{1}{r|}{0.60811} & \multicolumn{1}{r}{\textbf{0.81818}} \\
    \end{tabular}
\end{center}
\end{table}

Table~\ref{table_results_vat2} presents a comparative analysis between the baseline ResNet18 model and the VAT method~\citep{miyato_maeda_koyama} on the real-world highway dataset 1. The performance of VAT improves with an increase in the percentage of labeled data, as evidenced by the F2-score. The final accuracy achieved by VAT (shown in the bottom part) is notably superior to that of the baseline model (shown in the top part), as reflected in the F2-score. Additionally, VAT demonstrates more consistent classification of interferences, as indicated by the F5-score and recall score.

\begin{figure}[!t]
	\begin{minipage}[t]{0.325\linewidth}
        \centering
    	\includegraphics[trim=11 11 10 11, clip, width=1.0\linewidth]{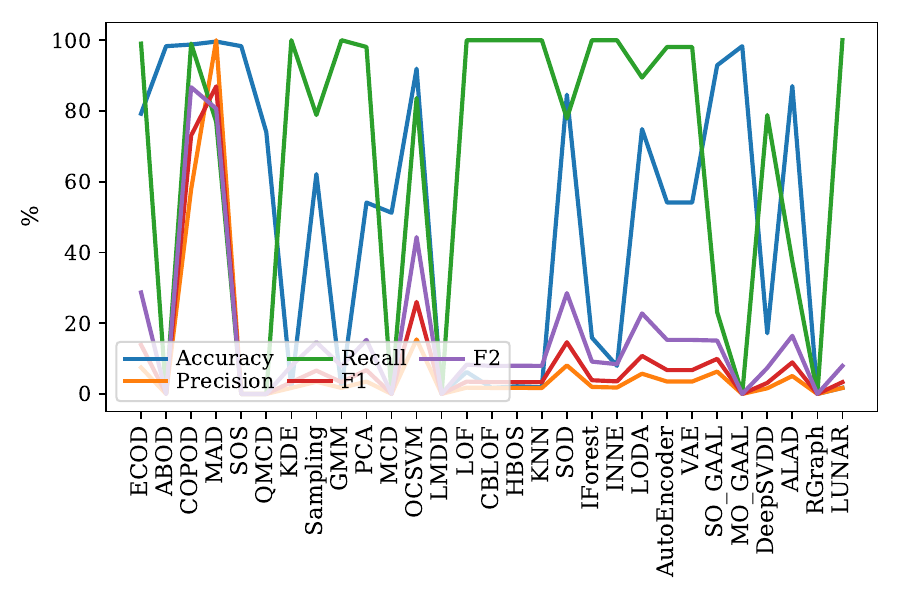}
        \subcaption{Train and test set: Real-world highway 1.}
        \label{figure_pyod1}
    \end{minipage}
    \hfill
	\begin{minipage}[t]{0.325\linewidth}
        \centering
    	\includegraphics[trim=11 11 10 11, clip, width=1.0\linewidth]{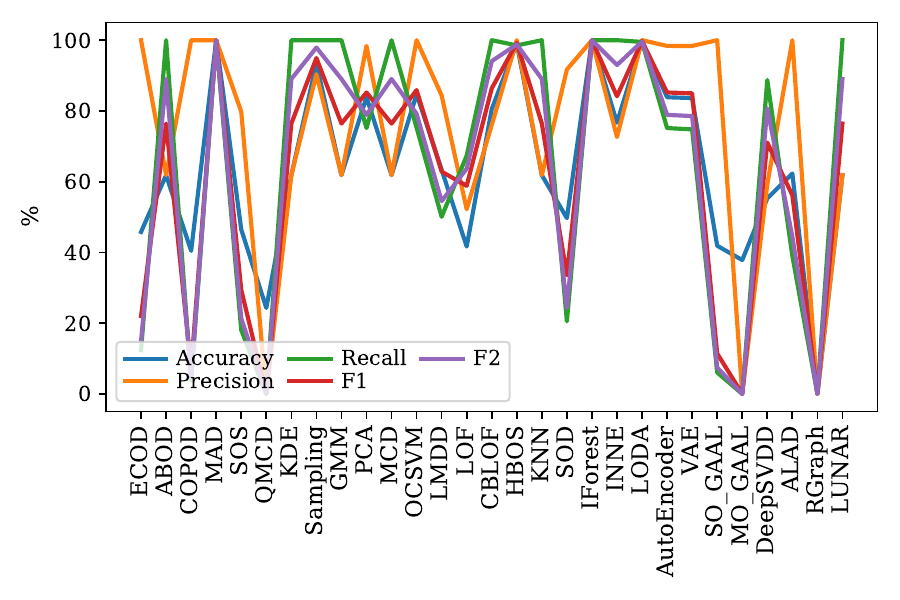}
        \subcaption{Train and test set: Controlled large-scale.}
        \label{figure_pyod2}
    \end{minipage}
    \hfill
	\begin{minipage}[t]{0.325\linewidth}
        \centering
    	\includegraphics[trim=11 11 10 11, clip, width=1.0\linewidth]{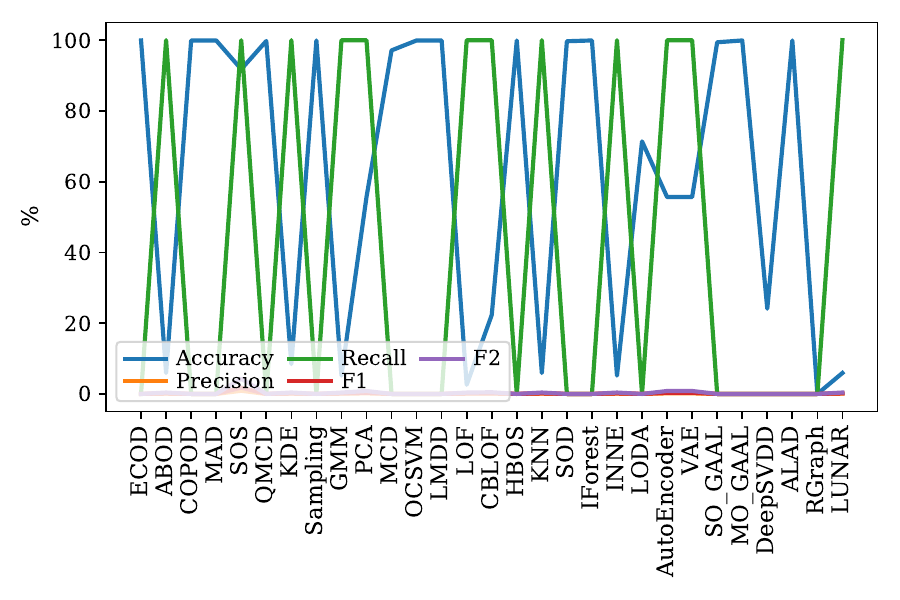}
        \subcaption{Train and test set: Controlled small-scale.}
        \label{figure_pyod3}
    \end{minipage}
	\begin{minipage}[t]{0.325\linewidth}
        \centering
    	\includegraphics[trim=11 11 10 11, clip, width=1.0\linewidth]{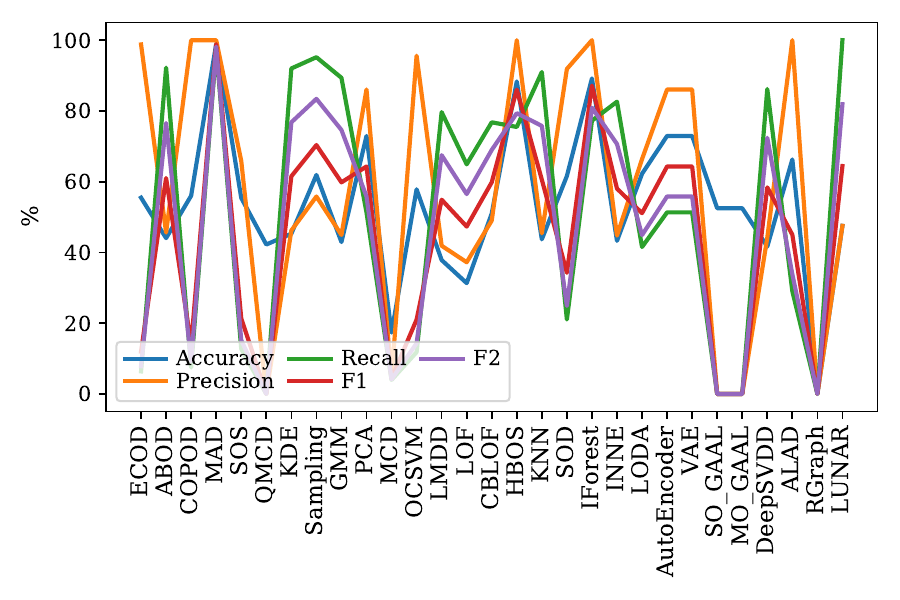}
        \subcaption{Train and test set: Seetal Alps sensor 1.}
        \label{figure_pyod4}
    \end{minipage}
    \hfill
	\begin{minipage}[t]{0.325\linewidth}
        \centering
    	\includegraphics[trim=11 11 10 11, clip, width=1.0\linewidth]{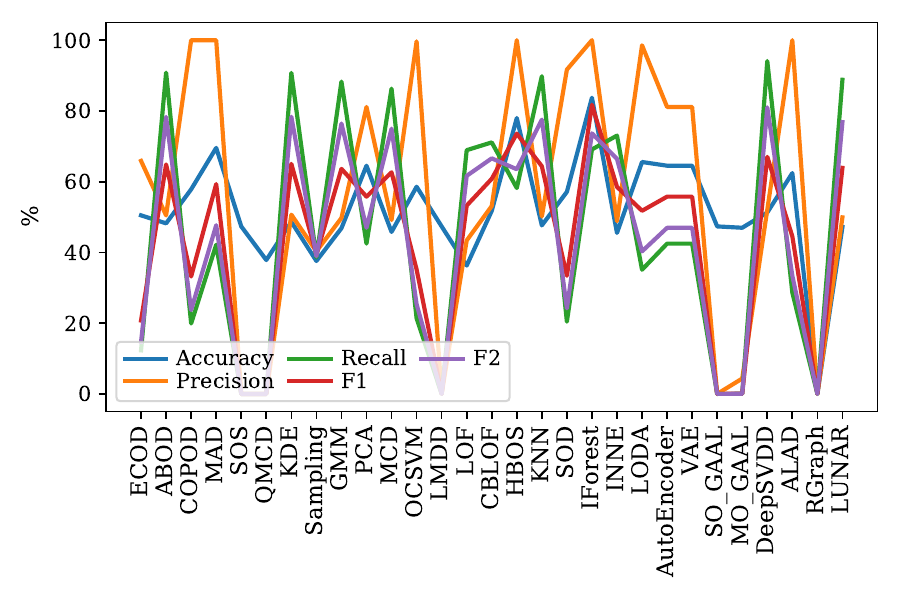}
        \subcaption{Train and test set: Seetal Alps sensor 2.}
        \label{figure_pyod5}
    \end{minipage}
    \hfill
	\begin{minipage}[t]{0.325\linewidth}
        \centering
    	\includegraphics[trim=11 11 10 11, clip, width=1.0\linewidth]{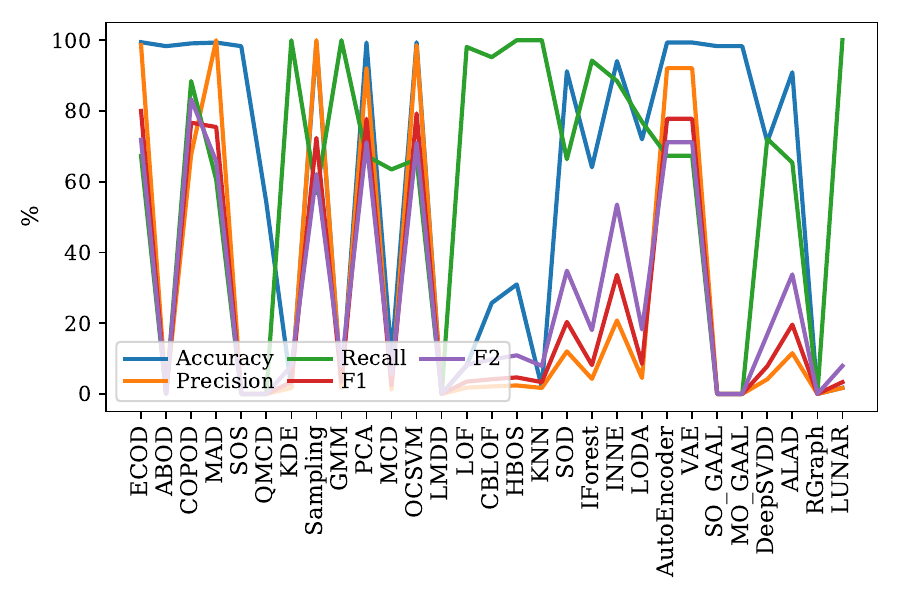}
        \subcaption{Train set: combined. Test set: Real-world highway.}
        \label{figure_pyod6}
    \end{minipage}
    \caption{Evaluation results of all outlier detection methods on several LC sensor datasets.}
    \label{figure_pyod}
\end{figure}

\subsection{Evaluation of Outlier Detection Methods on LC Data}
\label{label_eval_outlier_detection}

Subsequently, we evaluate the outlier detection methods described in Section~\ref{section_exp_outlier_detection} across all LC datasets. The results are summarized in Figure~\ref{figure_pyod}. On the real-world highway dataset 1 (Figure~\ref{figure_pyod1}), the methods ABOD, COPOD, MAD, SOS, and MO\_GAAL achieve the highest accuracies, reaching up to 100\%. In contrast, on the controlled large-scale dataset (Figure~\ref{figure_pyod2}), the outlier detection methods generally perform better, as the dataset is evenly balanced between classes. As a result, the F1-score and F2-score are higher compared to those on the highway dataset, with Isolation Forest achieving the highest accuracy. On the controlled small-scale dataset (Figure~\ref{figure_pyod3}), either the accuarcy or recall reaches 100\%, indicating that the models exclusively predict either positive classes (i.e., interferences) or negative classes. The Seetal Alps dataset (Figure~\ref{figure_pyod4} and Figure~\ref{figure_pyod5}) poses challenges due to real-world multipath effects, resulting in model accuracies between 50\% and 70\%. When trained on all datasets combined (Figure~\ref{figure_pyod6}) and tested on the real-world highway dataset, classification accuracy either improves (for models ECOD, ABOD, COPOD, MAD, SOS, AutoEncoder, VAE, and SO\_GAAL) or declines (for models SOD, LOF, CBLOF, and HBOS). Thus, despite discrepancies in the data, training on additional dataset can enhance model generalization, particularly when transitioning from controlled indoor data to real-world highway data.

\begin{figure}[!t]
	\begin{minipage}[t]{0.24\linewidth}
        \centering
    	\includegraphics[trim=11 11 10 11, clip, width=1.0\linewidth]{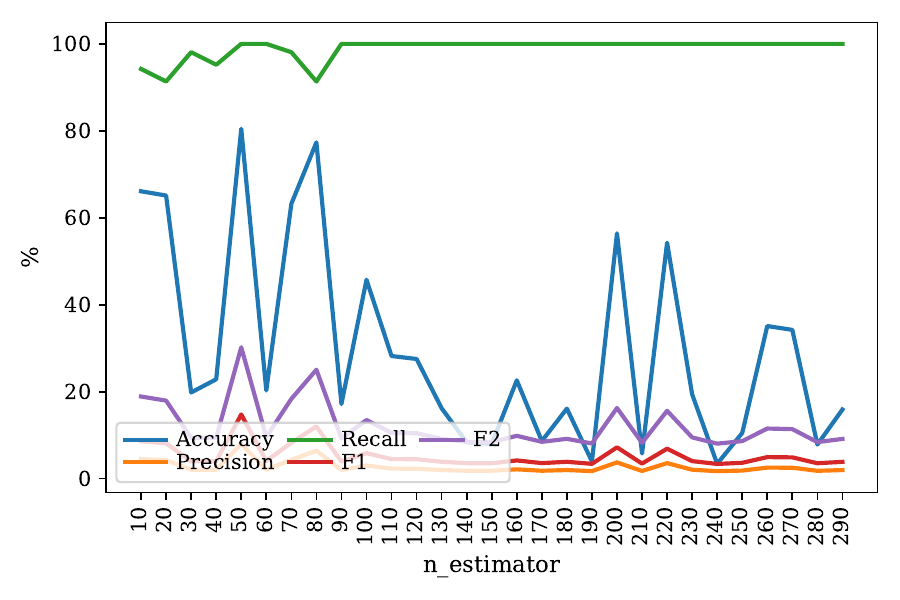}
        \subcaption{Train and test set: Real-world highway 1.}
        \label{figure_pyod_param1}
    \end{minipage}
    \hfill
	\begin{minipage}[t]{0.24\linewidth}
        \centering
    	\includegraphics[trim=11 11 10 11, clip, width=1.0\linewidth]{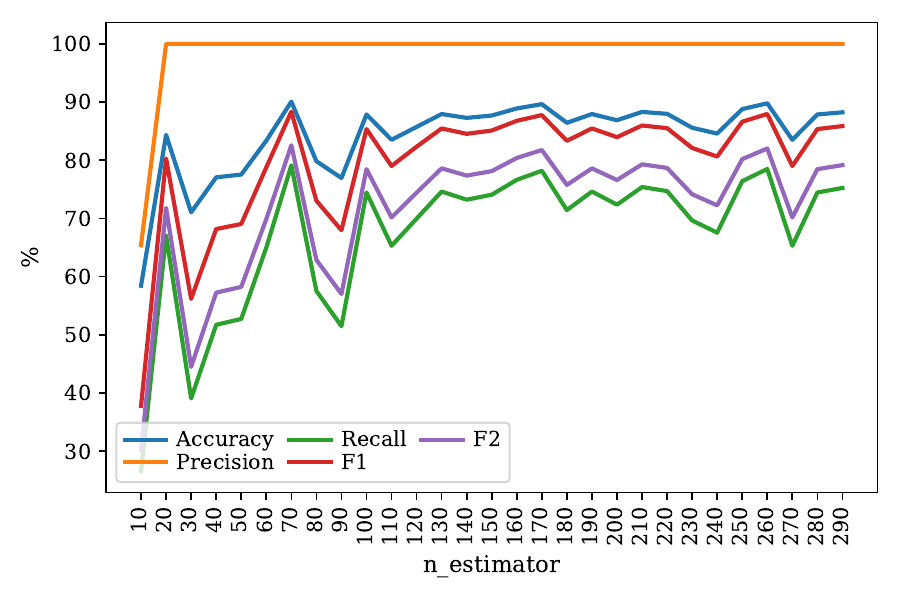}
        \subcaption{Train set: real-world highway 1. Test set: Seetal Alps Sensor 1.}
        \label{figure_pyod_param2}
    \end{minipage}
    \hfill
	\begin{minipage}[t]{0.24\linewidth}
        \centering
    	\includegraphics[trim=11 11 10 11, clip, width=1.0\linewidth]{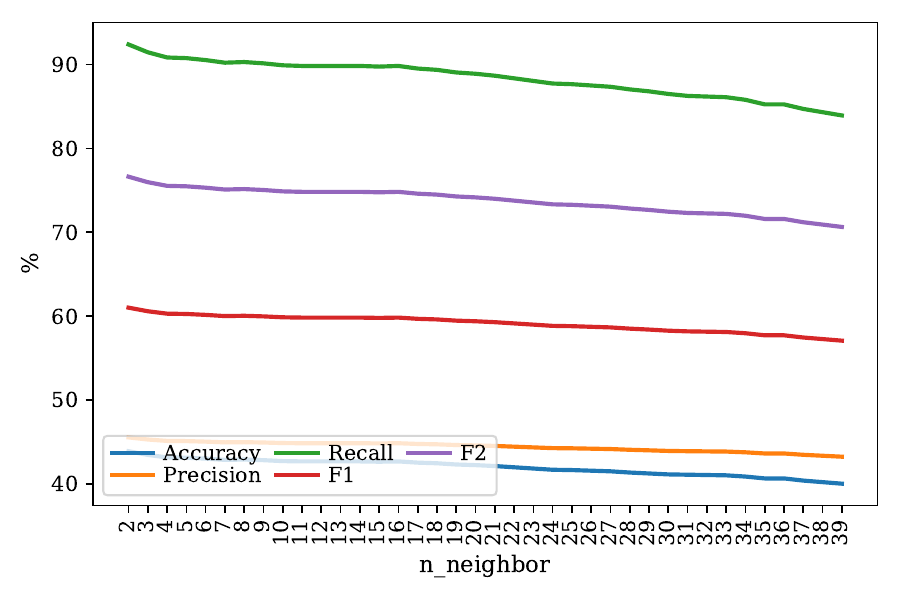}
        \subcaption{Train set: real-world highway 1. Test set: Seetal Alps Sensor 1.}
        \label{figure_pyod_param3}
    \end{minipage}
    \hfill
	\begin{minipage}[t]{0.24\linewidth}
        \centering
    	\includegraphics[trim=11 11 10 11, clip, width=1.0\linewidth]{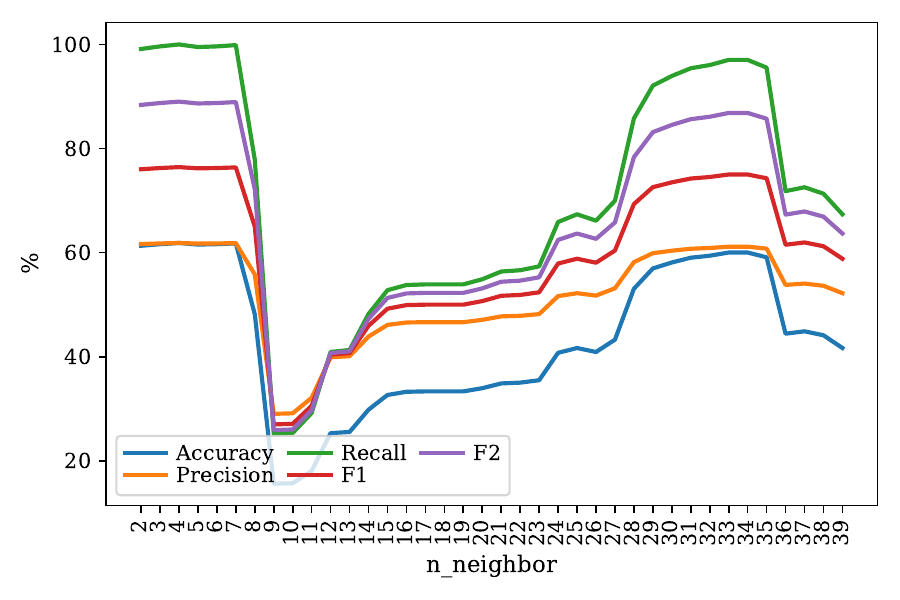}
        \subcaption{Train and test set: Controlled large-scale dataset.}
        \label{figure_pyod_param4}
    \end{minipage}
    \hfill
    \caption{Parameter searches for the Isolation Forest (a and b) and LOF (c and d) methods.}
    \label{figure_pyod_param}
\end{figure}

Furthermore, we conduct a detailed analysis of the hyperparameters for Isolation Forest and LOF (Figure~\ref{figure_pyod_param}). For Isolation Forest, we examine the number of base estimators, denoted as $n_{\text{estimator}}$, within the ensemble. Increasing the number of estimators generally enhances the model's performance but also leads to longer computation time. The model exhibits sensitivity to the number of estimators on the real-world highway dataset (Figure~\ref{figure_pyod_param1}), where accuracy decreases when $n_{\text{estimator}} > 100$. In contrast, for the Seetal Alps dataset (Figure~\ref{figure_pyod_param2}), accuracy improves as the number of estimators increases. This improvement is attributed to the large number of features and complex relationships in the Seetal Alps dataset, where additional estimators help the model capture the data's intricacies, thereby enhancing its robustness to various types of anomalies. On the highway dataset, a lower number of estimators may prevent the model from overfitting noise, leading to better generalization. For LOF (Figure~\ref{figure_pyod_param3}), we investigate the number of neighbors, $n_{\text{neighbor}}$, which represents the number of neighbors used in queries to determine the outlier status of a data point. As the number of neighbors increases, the LOF score becomes more influenced by a broader neighborhood, reducing its sensitivity to small, local variations. This approach is beneficial when anomalies are expected to deviate significantly from a large cluster of points, as seen in the highway dataset. Conversely, reducing the number of neighbors makes the LOF score more sensitive to the local density surrounding a point, which aids in detecting outliers in more subtle or densely clustered areas, as demonstrated in the controlled large-scale dataset (Figure~\ref{figure_pyod_param4}).

\begin{figure}[!t]
	\begin{minipage}[t]{0.495\linewidth}
        \centering
    	\includegraphics[trim=10 10 10 10, clip, width=1.0\linewidth]{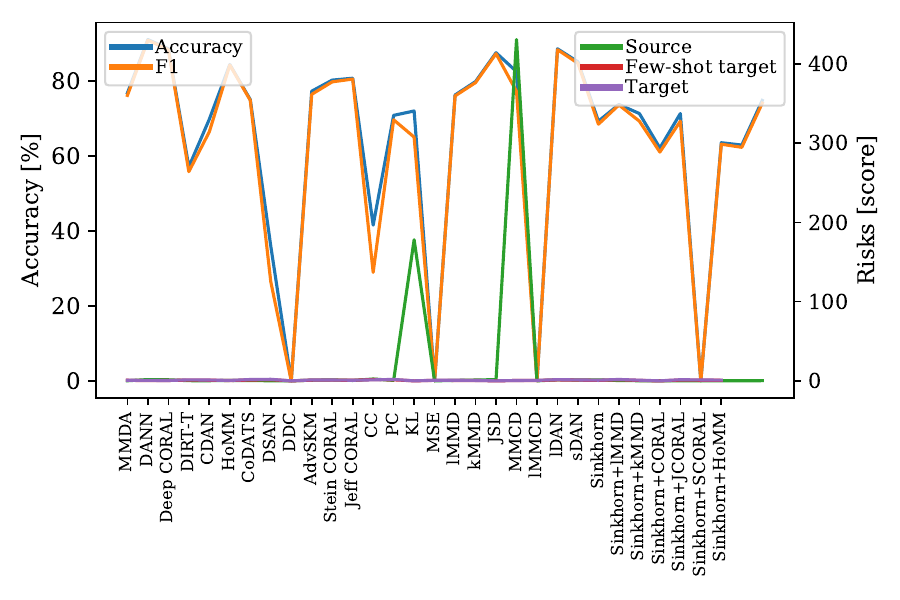}
        \subcaption{Source domain: Controlled large-scale dataset.}
        \label{figure_results_da1}
    \end{minipage}
    \hfill
	\begin{minipage}[t]{0.495\linewidth}
        \centering
    	\includegraphics[trim=0 0 0 0, clip, width=1.0\linewidth]{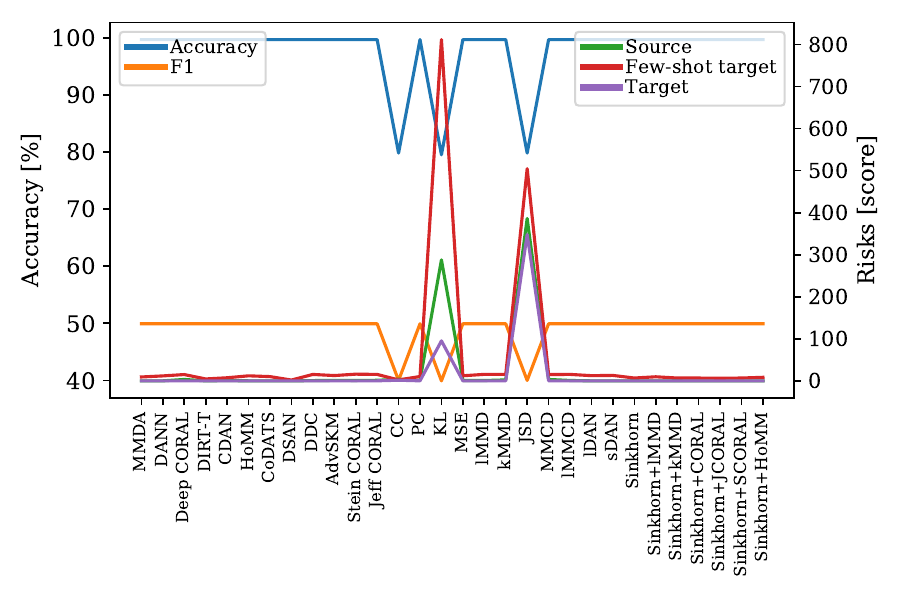}
        \subcaption{Source domain: Controlled small-scale dataset.}
        \label{figure_results_da2}
    \end{minipage}
	\begin{minipage}[t]{0.495\linewidth}
        \centering
    	\includegraphics[trim=0 0 0 0, clip, width=1.0\linewidth]{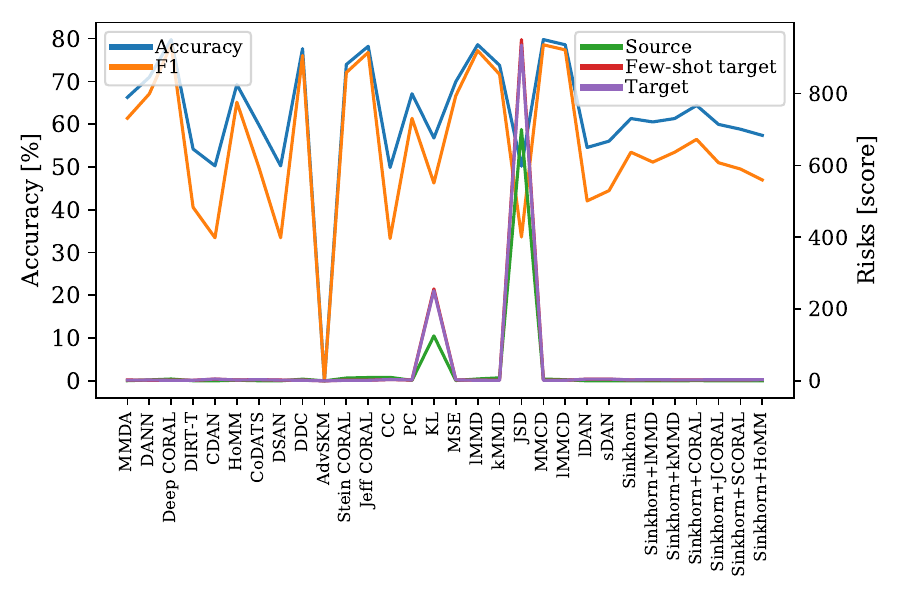}
        \subcaption{Source domain: Seetal Alps, sensor 1.}
        \label{figure_results_da3}
    \end{minipage}
    \hfill
	\begin{minipage}[t]{0.495\linewidth}
        \centering
    	\includegraphics[trim=0 0 0 0, clip, width=1.0\linewidth]{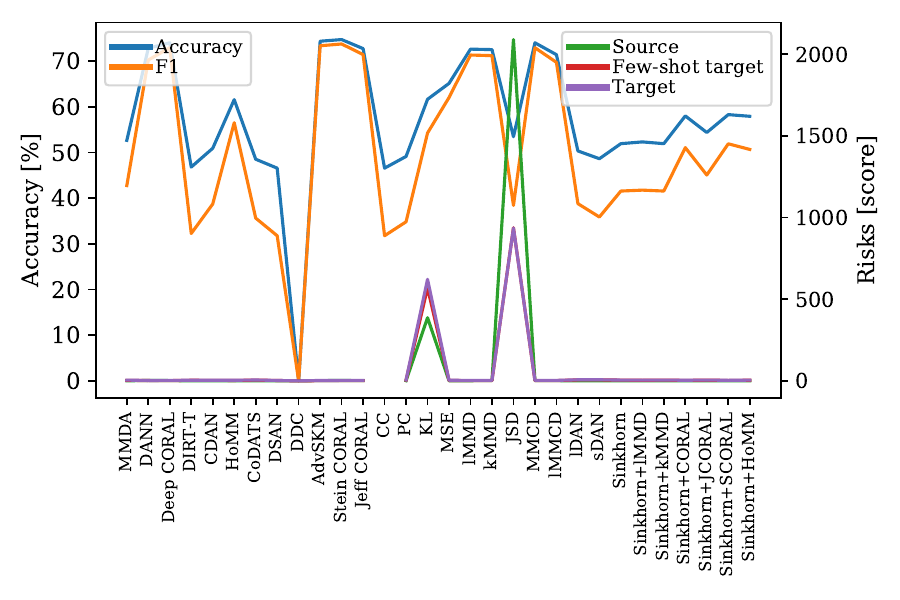}
        \subcaption{Source domain: Seetal Alps, sensor 2.}
        \label{figure_results_da4}
    \end{minipage}
    \caption{Evaluation results of all domain adaptation methods from various source domains to the target domain \textit{real-world highway dataset 1}.}
    \label{figure_results_da}
\end{figure}

\subsection{Evaluation of Domain Adaptation Methods on LC Data}
\label{label_eval_domain_adaptation}

Model selection and hyperparameter tuning present significant challenges in unsupervised DA due to the absence of target domain labels. To address these challenges, we draw on the source, target, and few-shot target risks proposed by \cite{ragab_eldele_tan} for evaluation and hyperparameter selection. The source risk involves selecting candidate models that minimize the cross-entropy loss on a test set from the source domain. This approach is straightforward and requires no additional labeling effort, as it utilizes existing source domain labels. However, the effectiveness of the source risk is heavily dependent on the sample size of the source data and the extent of the distribution shift. In cases of substantial distribution shift and limited source data, the source risk may be less effective than the target risk. Nonetheless, the source risk is advantageous in that it can be estimated using only source labels, whereas the target risk necessitates labeled data from the target domain~\citep{ragab_eldele_tan}. The target risk involves reserving a significant subset of target domain samples and their labels as a validation set, which is then used to select the optimal candidate model. This approach naturally yields the best-performing hyperparameters for the target domain and can be considered an upper bound for the performance of an unsupervised DA method~\citep{ragab_eldele_tan}. The concept of few-shot target risk is introduced to identify a more practical and realistic model selection method for unsupervised DA. Few-shot labeling, which involves labeling a small number of samples, is both practical and cost-effective. Despite being computed with only a few samples, the few-shot risk proves to be effective and performs comparably to the traditional target risk~\citep{ragab_eldele_tan}.

Figure~\ref{figure_results_da} provides a summary of the DA results for all methods discussed in Section~\ref{section_exp_da_models}. In this study, we adapted models from four source domain datasets to the target domain, \textit{real-world highway dataset 1}, to demonstrate our ability to transition from controlled indoor environments to real-world settings. For the controlled large-scale dataset (as shown in Figure~\ref{figure_results_da1}), DANN, Deep CORAL, HoMM, and DAN achieved the highest accuracies, reaching up to 90\%, which approaches the upper bound depicted in Figure~\ref{figure_pyod2}. As a result, higher-order MMD-based methods outperformed other techniques. Additionally, the source, target, and few-shot target risks were all close to zero, indicating that the few-shot target risk is an appropriate metric for hyperparameter selection. For the Seetal Alps datasets (refer to Figure~\ref{figure_results_da3} and Figure~\ref{figure_results_da4}), all DA methods demonstrated robustness, achieving accuracies close to the upper bound shown in Figure~\ref{figure_pyod4} and Figure~\ref{figure_pyod5}. In particular, Deep CORAL, DDC, linear and kernelized MMD, and the combination of MMD and CORAL achieved high accuracies of up to 80\%. These experiments underscore the effectiveness of DA methods on the LC GNSS datasets, confirming the feasibility of adapting to novel environments. The adaptation to new interference types has been previously demonstrated in \cite{ott_heublein_icl}.

\begin{table}[t!]
\begin{center}
    \caption{Results (accuracy in \%) for predicting the driving direction from LC sensor data for the Seetal Alps dataset from both sensors for various ML model input lengths.}
    \label{table_results_direction}
    \small \begin{tabular}{ p{1.3cm} | p{0.5cm} | p{0.5cm} | p{0.5cm} | p{0.5cm} | p{0.5cm} | p{0.5cm} | p{0.5cm} | p{0.5cm} | p{0.5cm} | p{0.5cm} | p{0.5cm} }
    \multicolumn{1}{c||}{} & \multicolumn{2}{c|}{\textbf{Length: 10}} & \multicolumn{2}{c|}{\textbf{Length: 15}} & \multicolumn{2}{c|}{\textbf{Length: 20}} & \multicolumn{2}{c|}{\textbf{Length: 25}} & \multicolumn{2}{c}{\textbf{Length: 30}} \\
    \multicolumn{1}{c||}{\textbf{Dataset}} & \multicolumn{1}{c}{\textbf{Intf.}} & \multicolumn{1}{c|}{\textbf{Direc.}} & \multicolumn{1}{c}{\textbf{Intf.}} & \multicolumn{1}{c|}{\textbf{Direc.}} & \multicolumn{1}{c}{\textbf{Intf.}} & \multicolumn{1}{c|}{\textbf{Direc.}} & \multicolumn{1}{c}{\textbf{Intf.}} & \multicolumn{1}{c|}{\textbf{Direc.}} & \multicolumn{1}{c}{\textbf{Intf.}} & \multicolumn{1}{c}{\textbf{Direc.}} \\ \hline
    \multicolumn{1}{l||}{Seetal Alps Sensor 1} & \multicolumn{1}{r}{\textbf{98.259}} & \multicolumn{1}{r|}{\textbf{72.148}} & \multicolumn{1}{r}{93.937} & \multicolumn{1}{r|}{70.733} & \multicolumn{1}{r}{86.006} & \multicolumn{1}{r|}{64.391} & \multicolumn{1}{r}{89.051} & \multicolumn{1}{r|}{68.195} & \multicolumn{1}{r}{52.805} & \multicolumn{1}{r}{52.805} \\
    \multicolumn{1}{l||}{Seetal Alps Sensor 2} & \multicolumn{1}{r}{\textbf{94.296}} & \multicolumn{1}{r|}{\textbf{79.851}} & \multicolumn{1}{r}{70.995} & \multicolumn{1}{r|}{56.998} & \multicolumn{1}{r}{92.984} & \multicolumn{1}{r|}{80.091} & \multicolumn{1}{r}{90.015} & \multicolumn{1}{r|}{76.792} & \multicolumn{1}{r}{89.760} & \multicolumn{1}{r}{77.206} \\
    \multicolumn{1}{l||}{Seetal Alps Combined} & \multicolumn{1}{r}{89.481} & \multicolumn{1}{r|}{73.629} & \multicolumn{1}{r}{84.206} & \multicolumn{1}{r|}{69.311} & \multicolumn{1}{r}{89.287} & \multicolumn{1}{r|}{75.028} & \multicolumn{1}{r}{94.198} & \multicolumn{1}{r|}{81.572} & \multicolumn{1}{r}{\textbf{97.273}} & \multicolumn{1}{r}{\textbf{86.994}} \\
    \end{tabular}
\end{center}
\end{table}

\subsection{Predicting Driving Direction from LC Data}
\label{label_eval_driving_direction}

As introduced in Section~\ref{label_data_low_cost}, our objective is to predict vehicle driving behavior using LC sensor data. This is demonstrated using the Seetal Alps datasets, as illustrated in Figure~\ref{figure_recording_setup5} and Figure~\ref{figure_lc_energy6}. We trained models for two tasks: interference classification and driving direction classification, evaluating various LC input lengths (10, 15, 20, 25, and 30 timesteps). The classification results are summarized in Table~\ref{table_results_direction}. With an input length of 10 timesteps, we achieved classification accuracies of 98.3\% for sensor 1 and 94.3\% for sensor 2. However, the accuracy decreased when combining data from both sensors. Specifically, driving direction was classified with accuracies of 72.1\% and 79.9\% for sensors 1 and 2, respectively. By increasing the input length to 30 timesteps and combining both sensors, the accuracy of direction prediction improved to 86.99\%. In conclusion, even with a small amount of LC data, we can successfully classify both interference and driving direction.

\begin{figure}[!t]
	\begin{minipage}[t]{1.0\linewidth}
        \centering
    	\includegraphics[trim=0 0 0 0, clip, width=1.0\linewidth]{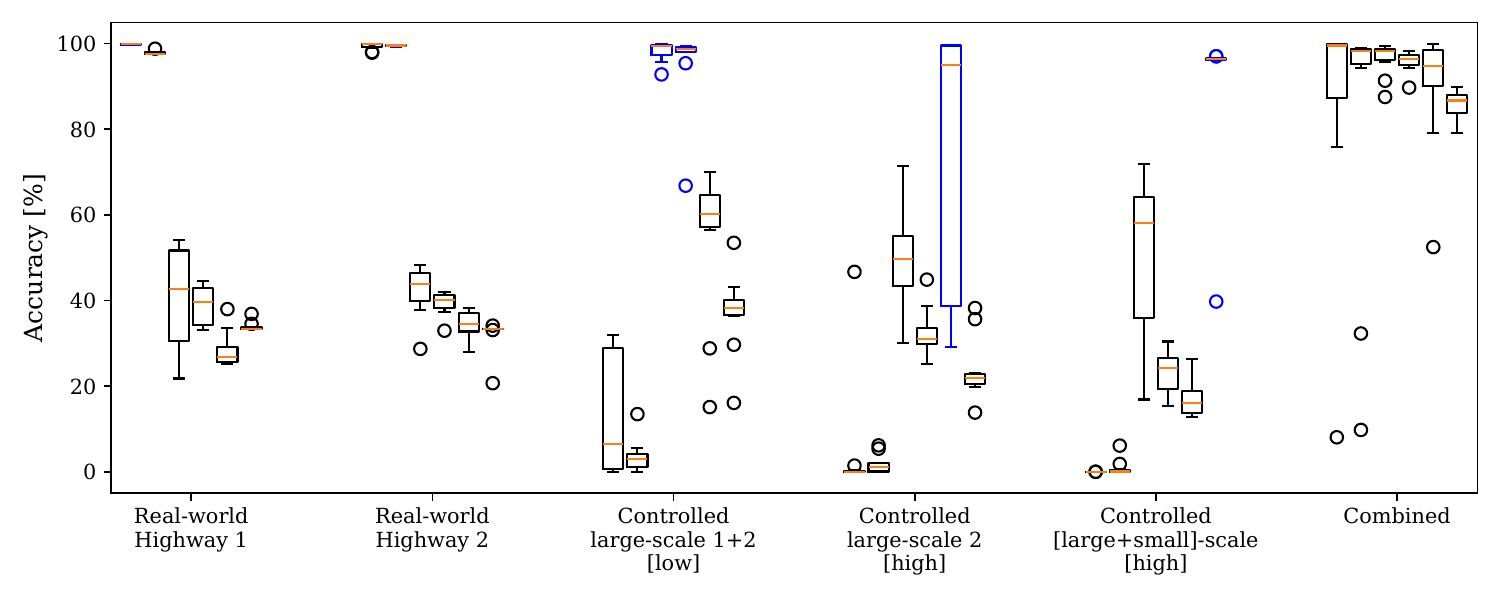}
        \subcaption{Accuracy [\%].}
        \label{figure_data_augmentation1}
    \end{minipage}
    \hfill
	\begin{minipage}[t]{1.0\linewidth}
        \centering
    	\includegraphics[trim=0 0 0 0, clip, width=1.0\linewidth]{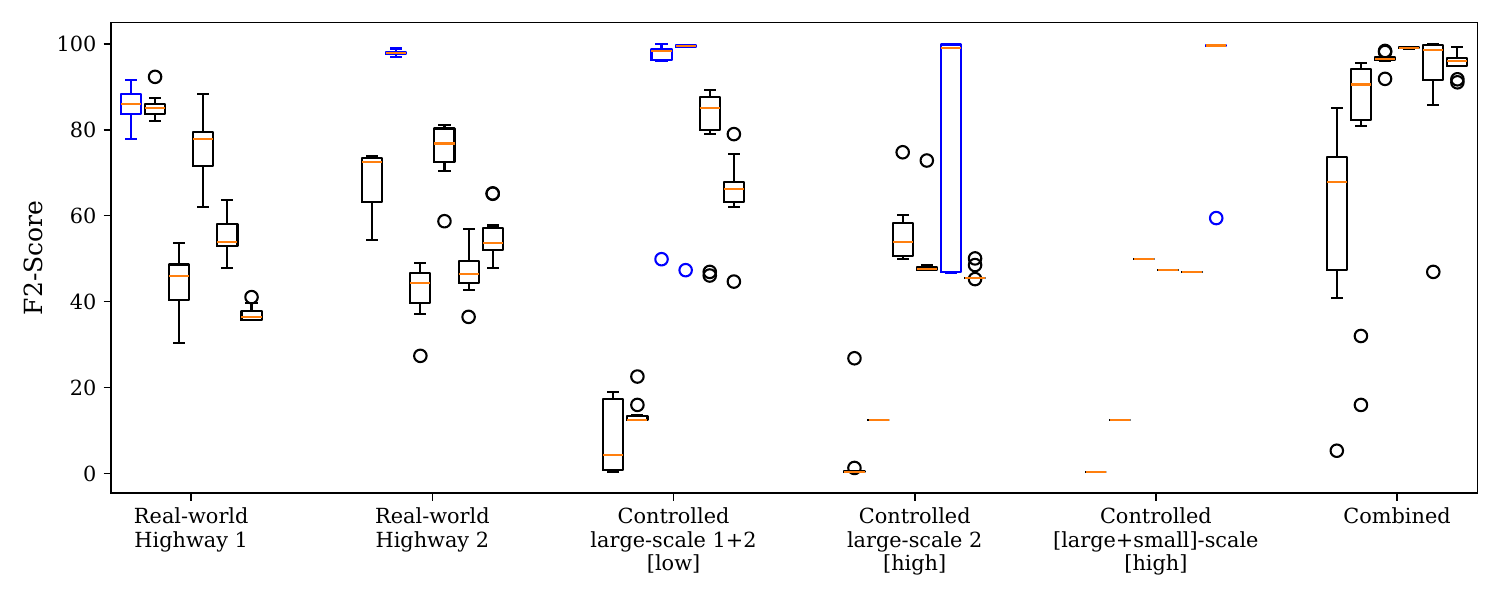}
        \subcaption{F2-score.}
        \label{figure_data_augmentation2}
    \end{minipage}
    \caption{Cross-validation of all snapshot datasets with data augmentation. Blue marks an equal train and test dataset.}
    \label{figure_data_augmentation}
\end{figure}

\subsection{Data Augmentation Results}
\label{label_eval_data_augmentation}

In Figure~\ref{figure_data_augmentation}, we present the results of cross-validation conducted on snapshot datasets with data augmentation, which should be compared to Figure~\ref{figure_cross_validation}, where no data augmentation was applied. On the same training and test dataset, the model's accuracy remains unchanged, indicating that the model is already well-fitted to this specific dataset. However, when cross-validating across different datasets, the model trained with data augmentation demonstrates greater robustness and shows significant improvement, particularly on the real-world highway dataset 1. Notably, when all datasets are combined for training, the accuracy does not improve further, as the model already exhibits high generalization.
\section{Conclusion}
\label{label_conclusion}

In the context of GNSS interference monitoring, our primary objective was to assess the effectiveness of supervised and unsupervised machine learning (ML) methods for classifying interferences and to address the discrepancies observed between datasets obtained in controlled environments and those collected in real-world scenarios. Based on our analysis, we have reached the following conclusions.
\begin{enumerate}
    \item We established a sensor station equipped with two antennas to capture GNSS snapshots encompassing a wide range of interference classes and characteristics. Data collection efforts resulted in two datasets recorded along German highways to detect jamming devices in vehicles, three datasets gathered in controlled indoor environments -- both small-scale and large-scale -- using low-frequency and high-frequency antennas, and one dataset obtained in the Seetal Alps in Austria.
    \item A significant disparity in data arises from substantial variations in jammer characteristics, differences in recording parameters such as frequency and antenna properties, satellite constellations, and environmental factors like multipath effects. Initially, we evaluate the feature embeddings of sensor channels that exhibit a considerable distinction between interference and non-interference classes, as well as notable differences across the proposed datasets.
    \item Additionally, our cross-validation across all datasets highlighted the challenges associated with training and testing on different datasets. Nevertheless, our ResNet18-based model successfully captured essential features across all data when the datasets were combined, enabling the model to be effectively applied across various scenarios.
    \item We analyzed the impact of interference on visible satellites, which is also influenced by the bandwidth of the interference.
    \item We conducted an analysis of the impact of interference on the position, velocity, and time (PVT) accuracy of a smartphone within a moving vehicle, revealing a significant effect.
    \item We presented an extensive benchmark of 20 supervised ML methods applied to snapshot data, demonstrating high interference classification accuracy on real-world highway datasets. Based on our findings, we recommend employing a large architecture with a high parameter count. Addressing the imbalance between interference and non-interference classes remains a challenging task.
    \item The proposed pseudo-labeling method demonstrated effectiveness in the interference classification task, enabling a reduction in the required amount of labeled data by approximately 30\%.
    \item Our benchmark of 20 outlier detection methods on low-cost (LC) GNSS data demonstrated high accuracy on the real-world highway dataset 1, with Isolation Forest achieving the highest accuracy. The Seetal Alps dataset posed significant challenges due to real-world multipath effects, leading to model accuracies ranging between 50\% and 70\%. Despite data discrepancies, incorporating additional datasets can improve model generalization, especially when transitioning from controlled indoor data to real-world highway data. However, it is important to note that outlier detection methods are sensitive to hyperparameter settings.
    \item We applied 24 domain adaptation (DA) models from four source domain datasets to the target domain, specifically the real-world highway dataset 1, to demonstrate our capability to transition from controlled indoor environments to real-world settings. Among the DA methods, DANN, Deep CORAL, HoMM, and DAN achieved the highest accuracies, approaching the upper bound.
    \item Additionally, our objective is to predict vehicle driving behavior using LC sensor data. Our model demonstrates the capability to classify interference with an accuracy of up to 98.3\%, while also predicting vehicle direction with an accuracy of up to 86.99\%.
    \item Data augmentation significantly enhances accuracy and F2-score across various training and test datasets by increasing model generalization. However, when applied to the same train/test dataset, data augmentation has no impact. Additionally, in the combined training of all datasets, data augmentation does not lead to improved results due to the already high variance present in the data.
\end{enumerate}
In conclusion, we have introduced a diverse range of datasets and conducted a detailed analysis of data discrepancies. We recommend employing pseudo-labeling, outlier detection, domain adaptation, and data augmentation methods to address domain shifts within the datasets. By integrating these proposed techniques, we are able to effectively adapt to novel environments and interference classes, thereby establishing an efficient GNSS interference monitoring system.

In future work, we will introduce an uncertainty-based voting mechanism designed to select more accurate pseudo-labels across multiple models. Additionally, we will assess disentanglement techniques for extracting GNSS-specific features, facilitating the generation of arbitrary amounts of data, and thereby enhancing our datasets. Finally, we will explore the use of diffusion models for data augmentation and interference mitigation, focusing on the unsupervised segmentation of interferences and the generation of values for missing information.

\section*{Acknowledgements}

We express our gratitude to Jonathan Hansen, Simon Kocher, Fabian Benschuh, Jan Hofmann, and Santiago Urquijo for their assistance with data collection. This work has been carried out within the DARCII project, funding code 50NA2401, supported by the German Federal Ministry for Economic Affairs and Climate Action (BMWK), managed by the German Space Agency at DLR and assisted by the Bundesnetzagentur (BNetzA) and the Federal Agency for Cartography and Geodesy (BKG). This work was also supported by the Bavarian Ministry for Economic Affairs, Infrastructure, Transport and Technology through the Center for Analytics Data Applications (ADA-Center) within the framework of ``BAYERN DIGITAL II''.

\bibliographystyle{apalike}
\bibliography{ION_GNSS}

\end{document}